\PassOptionsToPackage{table,xcdraw,dvipsnames}{xcolor}
\PassOptionsToPackage{colorlinks}{hyperref}
\documentclass[AMS,Times2COL]{WileyNJDv5} 

\articletype{Original Research Paper}%

\received{Date Month Year}
\revised{Date Month Year}
\accepted{Date Month Year}
\journal{Journal}
\volume{00}
\copyyear{2023}
\startpage{1}

\usepackage{amsmath,amssymb,amsfonts}
\usepackage{graphicx}
\usepackage{textcomp}

\usepackage{enumitem}
\useunder{\uline}{\ul}{}
\usepackage[utf8]{inputenc}
\usepackage{nicefrac}
\usepackage{xspace}
\usepackage{float}
\usepackage{lineno}
\usepackage{caption}
\usepackage{subcaption}
\usepackage{rotating, lscape, longtable, tabu, booktabs, siunitx, multirow}
\usepackage{verbatim}
\usepackage{mathtools}

\newcommand{\bkgblue}{{\color{blue}\textit{bkg}}}
\newcommand{\ipsgreen}{{\color{ForestGreen}\textit{ips}}}
\newcommand{\difred}{{\color{red}\textit{dif}}}
\newcommand{\cellyellow}{{\color{YellowOrange}\textit{cell}}}

\newenvironment{nohyphens}{%
	\par
	\hyphenpenalty=10000
	\exhyphenpenalty=10000
	\sloppy 
}{\par}

\begin{document}

\title{Tokenizing Semantic Segmentation with Run Length Encoding}

\author[1]{Abhineet Singh}

\author[1]{Justin Rozeboom}

\author[1]{Nilanjan Ray}

\authormark{SINGH \textsc{et al.}}
\titlemark{Tokenizing Semantic Segmentation with Run Length Encoding}

\address[1]{\orgdiv{Department of Computing Science}, \orgname{University of Alberta}, \orgaddress{\state{Edmonton}, \country{Canada}}}

\corres{Corresponding author Abhineet Singh, UCOMM-4057, University of Alberta. \email{asingh1@ualberta.ca}}


\abstract[Abstract]{
\begin{nohyphens}
This paper presents a new unified approach to semantic segmentation in both images and videos by using language modeling to output the masks as sequences of discrete tokens.
We use run length encoding (RLE) to discretize the segmentation masks, and adapt the Pix2Seq framework to learn autoregressive models to output these tokens.
We propose novel tokenization strategies to compress the lengths of the token sequences to make it practicable to extend this approach to videos.
We also show how instance information can be incorporated into the tokenization process to perform panoptic segmentation.
We evaluate our models on two domain-specific datasets to demonstrate their competitiveness with the state of the art
in certain scenarios,
in spite of being severely bottlenecked by our limited computational resources.
We supplement these analyses by proposing several promising approaches to foster future competitiveness in general-purpose applications, and facilitate this by making our code and models publicly available.
\end{nohyphens}
}


\jnlcitation{\cname{%
\author{Singh A.},
\author{Rozeboom J.},
\author{Ray N.}}.
\ctitle{Tokenizing Semantic Segmentation with Run Length Encoding} \cjournal{\it IET Computer Vision.} \cvol{2026;00(00):1--18}.}

\maketitle

\renewcommand\thefootnote{}
\footnotetext{\textbf{Abbreviations:} RLE, Run Length Encoding; LAC, Lengths-As-Class; BAC, Background-As-Class; TAC, Time-As-Class; LTAC, Lengths-and-Time-As-Class, IW, Instance-Wise; CW, Class-Wise; IPSC, Induced Pluripotent Stem Cell dataset; ARIS, Alberta River Ice Segmentation dataset.
}

\renewcommand\thefootnote{\fnsymbol{footnote}}
\setcounter{footnote}{1}

\section{Introduction}
\label{sec:intro}
Computer vision models typically produce outputs that are continuous-valued and fixed-sized, that is, they consist of real numbers and their size is independent of the contents of the input images.
This is particularly unsuitable for tasks like object detection and multi-object tracking where the output is inherently sparse and discrete in nature.
This is less of an issue for dense recognition tasks like semantic segmentation, though it has been shown \cite{p2s_generalist} that such tasks can also be modeled with tokenization.
This paper provides additional evidence for this by adapting the Pix2Seq language modeling framework \cite{p2s} for autoregressive semantic segmentation in both images and videos.
It serves as a companion to our concurrent work on tokenizing video object detection \cite{singh25_p2s_vid_det_access} to demonstrate the wide applicability of tokenization to both sparse and dense vision tasks.
It also proposes a simple way to add instance information to the token sequence, thereby achieving panoptic segmentation.

A brief review of tokenization in computer vision is included in our companion paper \cite{singh25_p2s_vid_det_access}, so we are not repeating it here for the sake of brevity.
To the best of our knowledge, \cite{p2s_generalist} is the only other existing work that applies tokenization to semantic segmentation.
Our method differs from \cite{p2s_generalist} in two key aspects.
Firstly, \cite{p2s_generalist} uses diffusion to output the raw segmentation masks without any form of compression. 
We instead use autoregression to output a lossless compressed representation of masks, thus significantly reducing output redundancy.
We also posit that our approach is a purer form of language modeling since \cite{p2s_generalist} requires the mask pixels to be converted from discrete integers into continuous floating-point values in order to be compatible with the diffusion model, thereby partially undoing the benefits of tokenization.
Secondly, \cite{p2s_generalist} gives restricted coverage to video segmentation, which is limited to only pairs of consecutive frames. 
We provide a more comprehensive treatment for video masks, with experiments covering up to $8$ frames and a unified representation that can theoretically handle any number of frames and is limited only by the available GPU memory.  

\begin{figure*}[t]
	\centering
	\includegraphics[width=\textwidth]{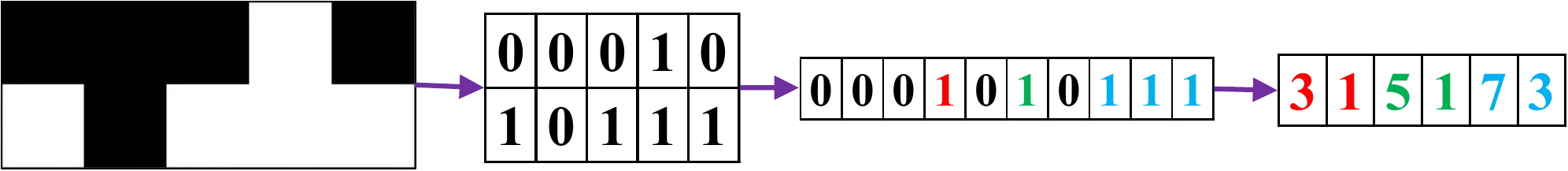}
	\caption{
		Generating RLE sequence for a $2\times 5$ binary mask using row-major flattening.
		Note that the start indices use 0-based indexing instead of the 1-based indexing commonly used in RLE.
	}
	\label{fig:seg_rle}	
\end{figure*}

To summarize, following are our main contributions:
\begin{itemize}
	\item We present a new approach to semantic segmentation by tokenizing the masks with run length encoding (RLE).
	\item We extend the RLE representation to work with video masks.
	\item We propose several tricks to compress the RLE sequence, e.g. Lengths-As-Class and Time-As-Class for static and video masks respectively.
	\item We propose a means to incorporate instance information in the RLE sequence to perform panoptic segmentation.
\end{itemize}

We have implemented the methods described here as extensions of the Pix2Seq framework \cite{p2s_git} and make our code and trained
models publicly available \cite{p2sv_git}.
Note that, as in \cite{singh25_p2s_vid_det_access}, many of the tokenization concepts in this paper can be illustrated better with animations than still images.
Since we cannot include animations here, we have created a website for this work \cite{p2sv_web} that contains animated versions of many of these images. We include links to these in the respective captions.

Also similarly to \cite{singh25_p2s_vid_det_access}, a significant portion of this paper has previously been included in a doctoral dissertation  \cite{p2sv_thesis}.
We have experimented with several more datasets and tokenization strategies since the dissertation in an attempt to achieve competitive performance under a wider range of scenarios.
We have improved and expanded our code base significantly in the process,
including
support for:
\begin{itemize}
    \item COCO \cite{coco_dataset} and Cityscapes \cite{cityscapes_dataset} datasets.
	\item Online RLE computation in tensorflow to allow for better data augmentation, in contrast to the original offline computation in numpy which necessitated all the augmented images to be generated before starting training.
	\item Class-token weight-equalization \cite[Sec. V.A ]{singh25_p2s_vid_det_access} for semantic segmentation to improve handling of class imbalance in the training set.
	\item Online randomization of the order of runs before generating the RLE sequence, similar to the randomized order of objects in \cite{p2s}, to improve generalization performance.
	\item Multi-headed decoder (Sec. \ref{future_work_architecture}) to output each RLE component by a separate head.
    \item Background-As-Class (BAC) and differential masks RLE compression strategies (Sec. \ref{future_work_compression}) to help reduce memory consumption.
\end{itemize}
However, we have thus far been unable to sufficiently overcome the hardware bottleneck (Sec. \ref{future_work}), so we decided to publish our existing work to allow others in the community
with fewer hardware constraints and a fresher perspective
to help take it forward.

The remainder of this paper is organized as follows.
Sec. \ref{rle} briefly explains the RLE encoding we have used to represent segmentation masks with sequences of discrete tokens.
Sec. \ref{static_seg} follows with details of how we tokenized RLE for the case of static images including the sliding window patches (Sec. \ref{sliding_win}) and Lengths-As-Class (LAC) encoding (Sec. \ref{lac}) we employed to deal with high resolution and multi-class masks.
Finally, Sec. \ref{vid_seg} concludes with an extension of this representation for videos, including the Time-As-Class (TAC) and Lengths-and-Time-As-Class (LTAC) schemes (Sec. \ref{tac}) we used to compress the RLE further to make it feasible to incorporate multiple masks without exceeding the token sequence length beyond practicable limits.
We have used the same network architectures for video segmentation as video detection \cite{singh25_p2s_vid_det_access}, so we do not cover these here.

\section{Run Length Encoding (RLE)}
\label{rle}
We have chosen to tokenize semantic segmentation masks using the run-length encoding (RLE) representation \cite{Robinson67_rle}.
This is a lossless data compression technique that flattens the segmentation mask into a 1D vector and represents this as a sequence of \textit{runs}.
A \textit{run} is a continuous sequence of non-zero pixel values that can be represented by a pair of integers - \textit{start}, \textit{length} - where \textit{start} is the index of the first pixel in the sequence and \textit{length} is the number of pixels in the sequence.
An example\footnote{This example has been borrowed from \href{https://ccshenyltw.medium.com/run-length-encode-and-decode-a33383142e6b}{this} online article} is shown in Fig. \ref{fig:seg_rle}.
A binary segmentation mask can thus be represented by a sequence of these pairs: 
\begin{itemize}[left=5pt,topsep=0pt,noitemsep,label={}]
	\item $
	{\color{red}\textit{start}_1, \textit{length}_1,}
	{\color{ForestGreen}\textit{start}_2, \textit{length}_2,}
	{\color{blue}\textit{start}_3, \textit{length}_3,}
	...
	$
\end{itemize}
while a multi-class mask would need a sequence of triplets since each run would also have the class ID: 
\begin{itemize}[left=5pt,topsep=0pt,noitemsep,label={}]
	\item $
	{\color{red}\textit{start}_1, \textit{length}_1, \textit{class}_1,}
	{\color{ForestGreen}\textit{start}_2, \textit{length}_2, \textit{class}_2,}
	...
	$
\end{itemize}
The mask can be flattened in either row-major and column-major order. 
Both of these lead to similar sequence lengths on average, although one might be more suitable than the other for specific cases.
For example, row-major would provide shorter sequences if most of the objects are short and wide while column-major would work better for tall and narrow objects.

We also considered two alternative mask representations including polygons \cite{Castrejn17_prnn,Acuna18_prnn,liang2020_polytransform,lazarow2022_BoundaryFormer,p2s_multi} and quadtrees \cite{Chitta19_quadtree}.
However, statistical tests showed that all of these representations require roughly the same number of tokens as RLE so we chose the latter since it provides two advantages over the others.
Firstly, it is much easier to implement, especially when generalizing to multi-class and video segmentation masks.
Secondly, it is likely to be more robust to noisy tokens during mask reconstruction at inference since a single run, which forms the geometrical unit of RLE tokenization, has a much smaller impact on the overall mask quality than a polygon or quadtree. 
For example, if a few RLE tokens go missing at inference, it is likely to have minimal impact on the overall mask quality since each run usually affects only a few nearby pixels in the same row (or column). 
Also, the gaps or artifacts in the mask thus created might be remedied relatively easily by image processing techniques like morphological operations \cite{Soille2003_Morphological}. 
However, a single missing polygon token could severely degrade the mask in a way that cannot be easily fixed by post-processing.

On the other hand, RLE does have an important disadvantage in that it makes object-level information difficult to learn since it represents each object by a large number of independent tokens.
It would be an interesting area of future work to figure out how to add object-level information to the mask tokens without losing the robustness benefit provided by the small quantum of RLE.

\section{Static Image Segmentation}
\label{static_seg}
\subsection{Tokenization}
\label{static_tokenization}
As detailed in the design of our video architectures \cite[Sec. III.B]{singh25_p2s_vid_det_access}, Pix2Seq is very difficult to train from scratch, so it is important that we are able to use as much of the pretrained weights as possible.
This in turn requires that the baseline architecture be modified as little as possible.
This imposes two architectural constraints on RLE tokenization – the size of the vocabulary $V$ and the maximum sequence length $L$.

\subsubsection{Architectural Constraints}
\label{Architectural_Constraints}
Out of the box, Pix2Seq has $L=512$ and $V=3K$ in the original object-detection-only variant \cite{p2s}, though the multi-task version \cite{p2s_multi} implemented in the same code-base supports $V$ upto $32K$.
The choice of how to tokenize RLE involves a trade-off between these two constraints,
wherein a mask can be encoded by fewer tokens
(thereby decreasing $L$)
by adding more unique tokens to the vocabulary
(thereby increasing $V$)
and vice versa.
Also, increasing either $L$ or $V$ causes both training time and GPU memory consumption to rise too, though this increase is significantly more pronounced for $L$ than $V$.
Hence, our overall objective was to keep the RLE sequence as short as possible while allowing the number of required tokens to increase up to $32K$.
We have been able to get the model working well with $L$ up to $4K$ and $V$ up to $28K$ on our limited hardware resources.
It is even possible to get training started with $L$ up to $8K$ but anything much above $4K$ results in the training run crashing soon afterwards, irrespective of $V$.
We trained all our models on RTX 3090 GPUs so it appears that the 24 GB RAM available on these is simply not enough for $L >> 4K$.
Also, the batch sizes we can use near these upper limits of $L$ and $V$ are far too small to
train models of such complexity to generalize well.

\subsubsection{Mask Flattening}
\label{Mask_Flattening}
\begin{figure}[t]
	\centering
	\includegraphics[width=0.49\textwidth]{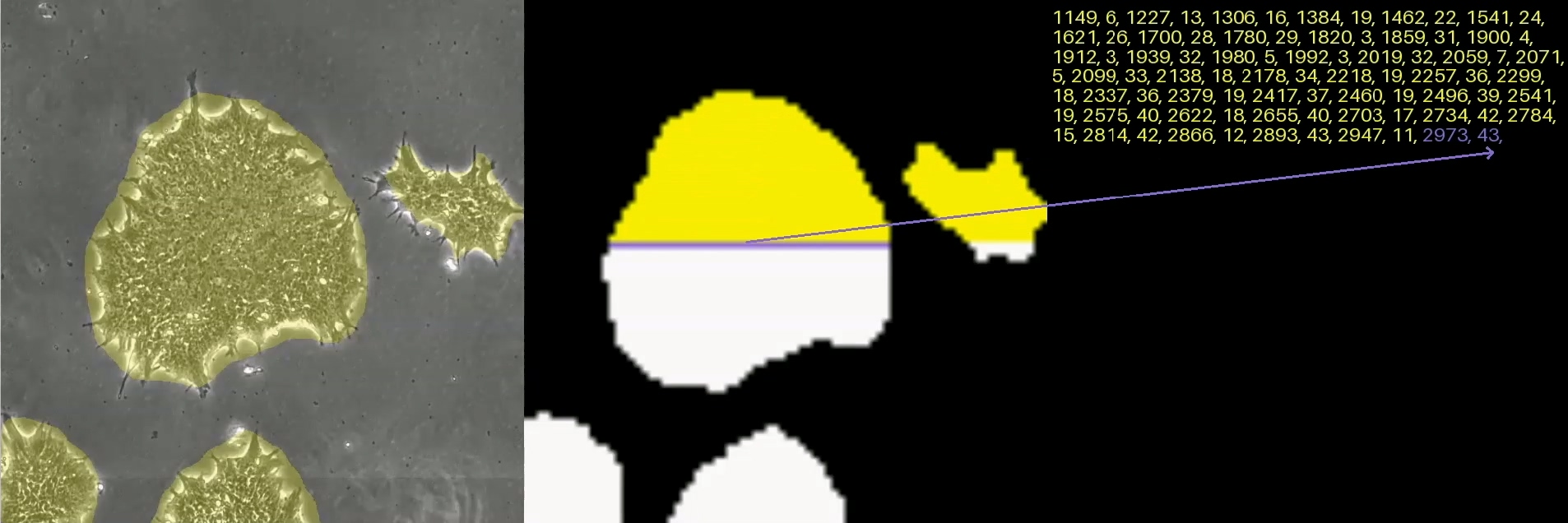}
	\includegraphics[width=0.49\textwidth]{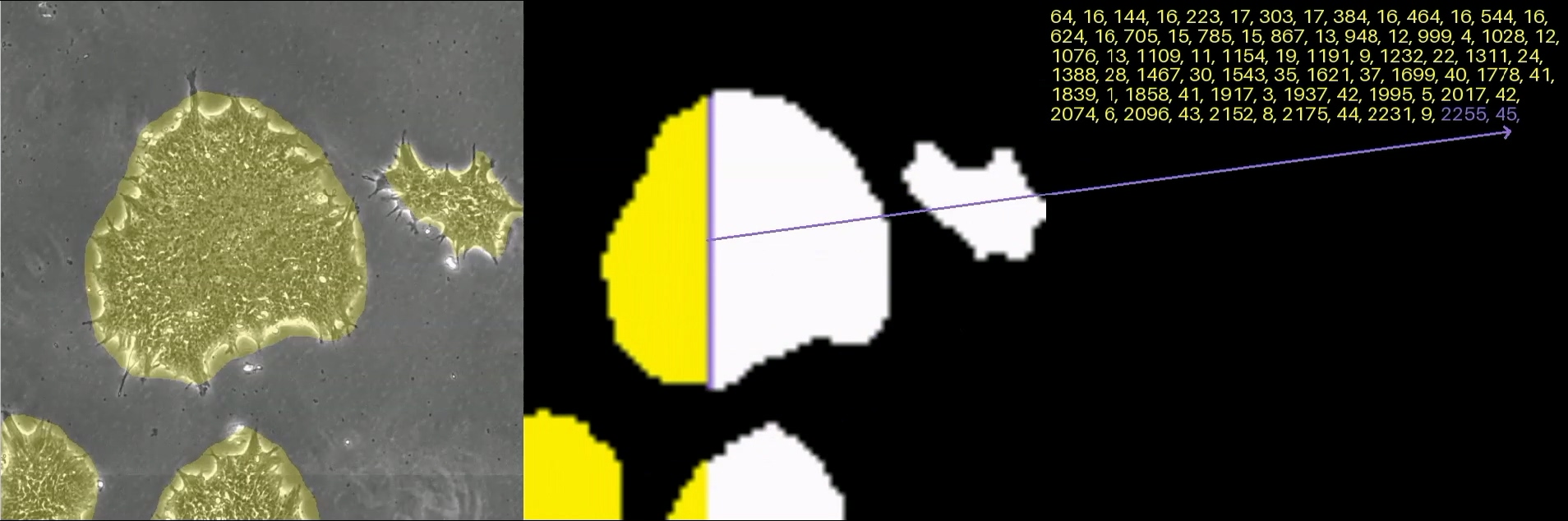}
	\caption{
		Visualization of RLE tokenization of binary segmentation masks with row-major (top) and column-major (bottom) flattening of the masks.
		Each figure shows (from left to right) the source image patch with the foreground mask drawn on it in yellow, binary version of this mask with the already tokenized segment in yellow, and the corresponding tokens.
		The run that is currently being tokenized is shown in purple.
		Animated versions of these figures are available
		\href{https://webdocs.cs.ualberta.ca/~asingh1/p2s\#seg_binary_row_major}{here}
		and
		\href{https://webdocs.cs.ualberta.ca/~asingh1/p2s\#seg_binary_column_major}{here}.
		Best viewed under high magnification.		
	}
	\label{fig:seg_binary}	
\end{figure}

Let us assume that we have an $S\times S$ binary segmentation mask.
If we employ the conventional practice of flattening the mask into a 1D vector, we can use a single token to represent the \textit{starts} but we would need $S^2$ different tokens in the vocabulary for the start indices.
Alternately, we can skip the flattening and just use the 2D coordinates directly to represent the start of each run.
In this case, we need only $S$ start tokens but now we need $3$ tokens to represent each run, which increases $L$ by $50\%$.
As mentioned above, reducing $L$ is more important than reducing $V$ so we have mostly been working with the flattened version, though we did train a few models with 2D start tokens as well.
Fig \ref{fig:seg_binary} shows examples of RLE tokenization for binary segmentation masks with both row-major (top) and column-major (bottom) mask flattening. 

\subsubsection{Shared Tokens}
\label{Shared_Tokens}
Further, we have the option to share the same set of tokens for both \textit{starts} and \textit{lengths} since the \textit{lengths} can theoretically be as large as the \textit{starts} when a single run covers the entire mask.
However, in practice, very few runs extend across multiple rows (or columns) so we decided to use separate tokens for \textit{lengths} and imposed a limit of $S$ on the maximum length that a run can have.
We handle runs that exceed $S$ by splitting them into multiple consecutive runs.
For example, with $S=80$, an overlong run $(300, 125)$ can be split into two runs $(300, 80)$ and $(380,45)$.
In theory, this can increase the total number of runs (and therefore $L$) dramatically but, as mentioned above, it is extremely rare for runs to extend across multiple rows (or columns) so it does not matter much in practice.

Having separate tokens for \textit{starts} and \textit{lengths} also provides a significant advantage during inference when we need to resolve the $L\times V$ probability distribution into actual tokens.
If \textit{starts} and \textit{lengths} are using the same set of $S^2$ tokens (assuming the flattened mask case), we would need to take the \textit{argmax} over all the $S^2$ probability values to generate each length token.
Using a separate set of $S$ tokens for \textit{lengths} allows us to take the \textit{argmax} over only these $S$ probabilities which helps to correct the run in cases where the network confounds the positioning of \textit{starts} and \textit{lengths} tokens.

\subsection{Sliding Windows}
\label{sliding_win}
\begin{figure}[t]
	\centering
	\includegraphics[width=0.49\textwidth]{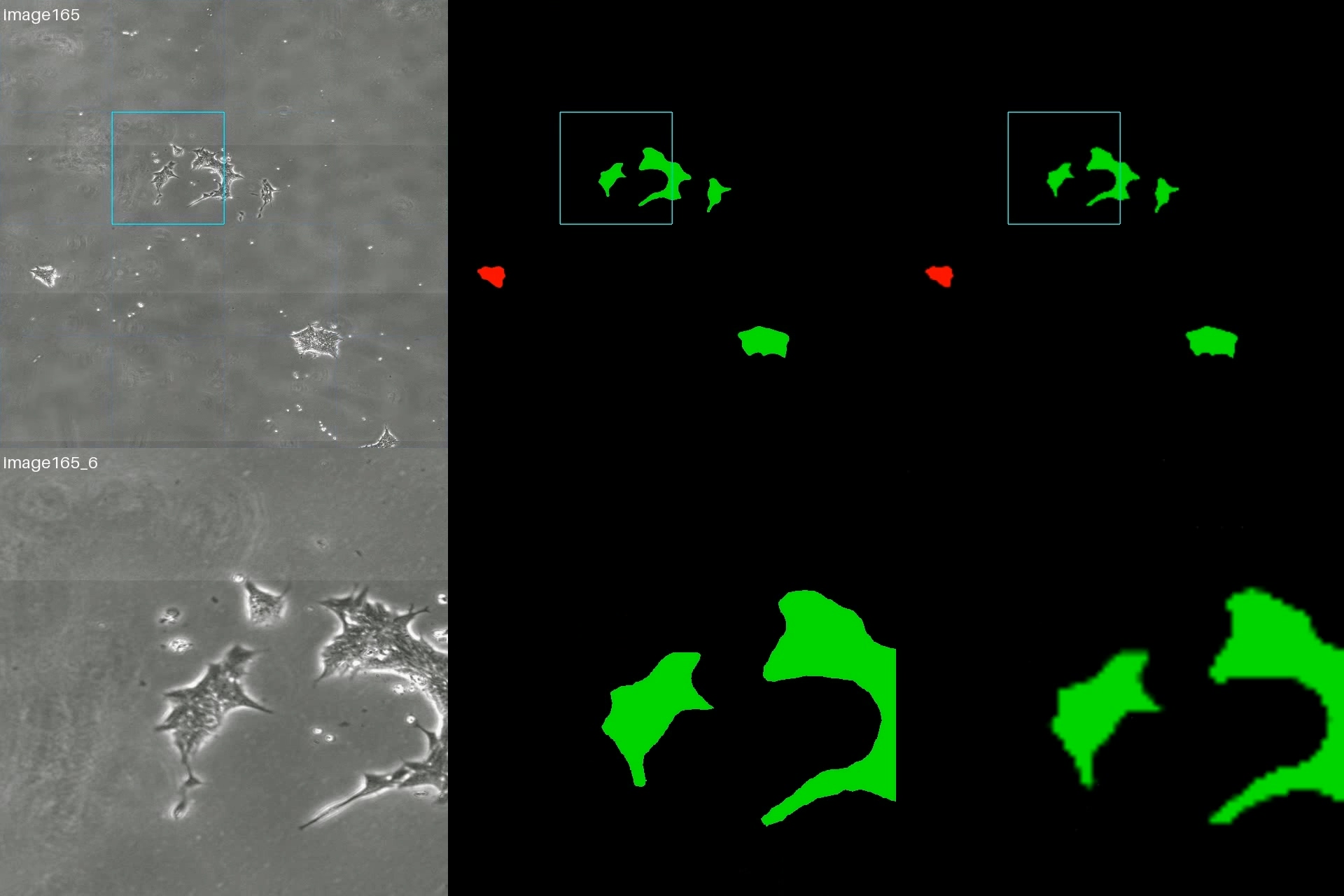}
	\caption{
		An example of the sliding window patch extraction and mask subsampling process on an image from the IPSC dataset.
		The top row shows (from left right) source image resized to $2560\times 2560$ with patch location shown by the blue box, corresponding mask at its full resolution of $2560\times 2560$, and this mask subsampled by a factor of 8 to $320\times 320$. 
		The bottom row shows (from left right) $640\times 640$ patch corresponding to the blue box, corresponding patch mask at its full resolution $640\times 640$, and this mask subsampled by a factor of 8 to $80\times 80$. 
		This subsampled $80\times 80$ mask is the one that is used for generating the RLE sequence.
		An animated version of this figure is available \href{https://webdocs.cs.ualberta.ca/~asingh1/p2s\#sliding_window_patches}{here}.		
	}
	\label{fig:sliding_win}	
\end{figure}
We collected statistics on the lengths of the RLE sequences required to represent segmentation masks for complete images over the entire IPSC and ARIS datasets (Sec. \ref{datasets}).
This turned out to be well over $512$ for many images even when they were resized down to $I=320$, which itself reduced the resolution of many IPSC sequences by a factor of more than $10$ \cite{Singh23_ipsc}.
In addition, even $S=320$ is not feasible with 1D start tokens because of the impracticably large vocabulary size $V=S^2=320^2=102.4K$ that this requires. 
Although such resolutions can be managed using 2D start tokens, this increases $L$ by $50\%$ which in turn significantly reduces the training batch sizes that can be used.

We resolved
these issues
by first extracting smaller patches of size $P < I$ from those images in a sliding window manner \cite{Singh2020_River_ice} and then training on these patches instead of the complete images. 
An example is shown in Fig. \ref{fig:sliding_win}.
We also applied all the patch dataset augmentation techniques from \cite{Singh2020_River_ice}.
These include employing random strides smaller than $P$ to generate overlapping patches, and applying random geometric transforms like rotation and horizontal and vertical flipping to the patches thus generated.

\subsubsection{Redundancy}
\label{redundancy}
Similar to overlapping temporal windows in video processing \cite{singh25_p2s_vid_det_access}, overlapping patches can be used to increase redundancy during inference.
In fact, when doing video segmentation (Sec. \ref{vid_seg}), we can get redundancy from both temporal and spatial windows.
However, unlike object detection, there is no straightforward method to combine information from multiple masks of the same region in the image to create a composite mask that is better than any of its constituents.
The best method we found was to employ a pixel-level voting strategy where we collect the class label for every  pixel from all the patches that contain that pixel and then use the most frequently occurring class  as the final label for that pixel.
However, even this resulted in a slight overall degradation of mask quality as compared to simply using the class labels from the last patch that contains each pixel (Sec. \ref{exp_vid_len}).
We also experimented with other strategies like pixel-wise \textit{or}, \textit{and}, \textit{min} and \textit{max} but none outperformed the voting scheme.

\subsubsection{Subsampling}
\label{subsampling}
We empirically found that patch sizes ranging from $P=I/4$ to $P=I/8$
are small enough to produce RLE sequences of suitable sizes (i.e. with $512\leq L\leq 4096$) on both IPSC and ARIS datasets, provided that the masks themselves are subsampled down to between $S=80$ and $S=160$ (Tables \ref{tab:rle_len_full_image} - \ref{tab:rle_len_full_and_patch}).
We are using the smallest version of the Pix2Seq architecture which takes $640 \times 640$ images as input and this gives the target size for the patches.

IPSC sequences have a large range of image sizes from $900\times 1700$ to $ 3833\times 4333$ \cite{Singh23_ipsc}.
Therefore, in addition to $L$ exceeding the required limit for the larger images, extracting patches directly from the source images leads to widely varying magnification levels between the different sequences.
Therefore, for most of our experiments on this dataset,  we first resized the images to $I=2560$ before extracting the patches.
This allows patches that are a quarter of the full image to reach the target size $P=I/4=2560/4=640$ and is also large enough that we do not lose much resolution in most of the images.
Unlike IPSC, the ARIS dataset has much more uniform sizes across its images  \cite{Singh2020_River_ice}, so that extracting patches directly from the source images with $P=640$ works fine on this dataset.
Also, ARIS images are small enough that directly resizing them to $I=640$ and skipping patch generation altogether also works without losing too much resolution.

After extracting the patches, we finally subsampled the $P=640$ patch masks down to either $S=80$ or $S=160$ and generated the RLE training data using these subsampled masks.
We collected statistics on the amount of degradation in the mask quality as a result of this subsampling, in terms of segmentation metrics (Sec. \ref{metrics}) obtained by comparing the subsampled masks to the original ones.
This turned out to be $<10\%$ in nearly every case (Table \ref{tab:segm_metrics_cmb}), which is sufficient for most practical purposes.

\subsection{Lengths-As-Class (LAC)}
\label{lac}
\begin{figure}[t]
	\centering
	\includegraphics[width=0.49\textwidth]{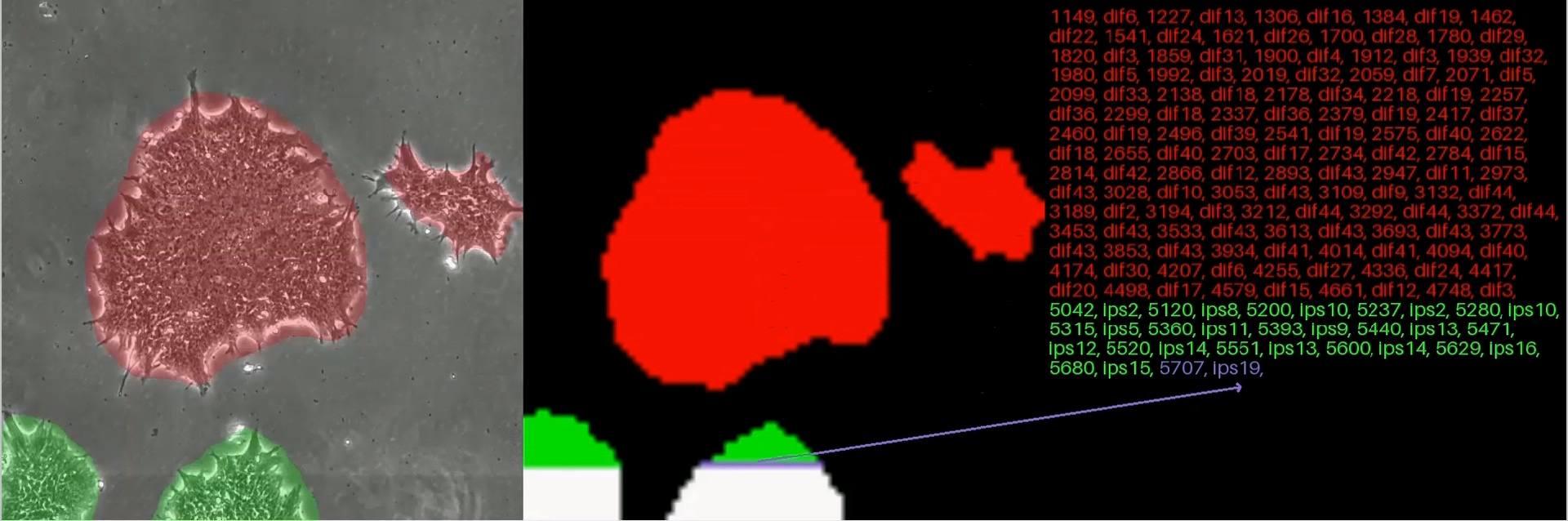}
	\caption{
		Visualization of LAC tokenization for multi-class segmentation.
		The figure shows (from left to right) the source image patch with the two classes in red and green, binary version of this mask with the already tokenized segment in red or green depending on the class, and the corresponding tokens.
		The run that is currently being tokenized is shown in purple.
		The LAC tokens are shown here as concatenations of class name and length but each such combination represents a single unique token.
		Animated version of this figure is available
		\href{https://webdocs.cs.ualberta.ca/~asingh1/p2s\#seg_lac}{here}.	
		Best viewed under high magnification.	
	}
	\label{fig:seg_multi_class}	
\end{figure}
As mentioned before, naïve encoding of multi-class segmentation masks requires $3$ tokens per run since we need an extra token for the class.
This can unnecessarily increase the sequence length by $50\%$ so we can go back to using $2$ tokens per run by combining the length and class tokens into a single composite token that represents both.
This can be done by considering the lengths as classes too, such that each unique combination of length and class is represented by a separate LAC token.
The total number of LAC tokens is then the product of the maximum length of a run (i.e., $S$) and the number of classes $C$.
The first $S$ LAC tokens correspond to runs of class $1$, next $S$ tokens correspond to class $2$ and so on.

This tokenization is particularly efficient when $C$ is small. 
For example, both IPSC and ARIS datasets have $C=2$ so that, with $S=80$, we get the number of \textit{starts} tokens = $80\times 80 = 6400$, number of LAC tokens = $80\times 2 = 160$ and $V = 6400 + 160 = 6560$. 
Without LAC tokenization, number of \textit{lengths} tokens = $80$, number of class tokens = $2$ and $V = 6400 + 80 + 2 = 6482$.
Therefore, we are able to reduce the RLE sequence length substantially without any significant increase in $V$.
However, for very large $C$, e.g., $C=80$ as in the COCO dataset \cite{coco_dataset}, $V$ can nearly double ($12.8K$ versus $6.56K$), though still remaining very much within the feasible range (i.e. $<32K$).
Fig. \ref{fig:seg_multi_class} shows examples of RLE tokenization for multi-class segmentation masks with LAC tokenization. 

\section{Video Segmentation}
\label{vid_seg}

\begin{figure}[t]
	\centering
	\includegraphics[width=0.49\textwidth]{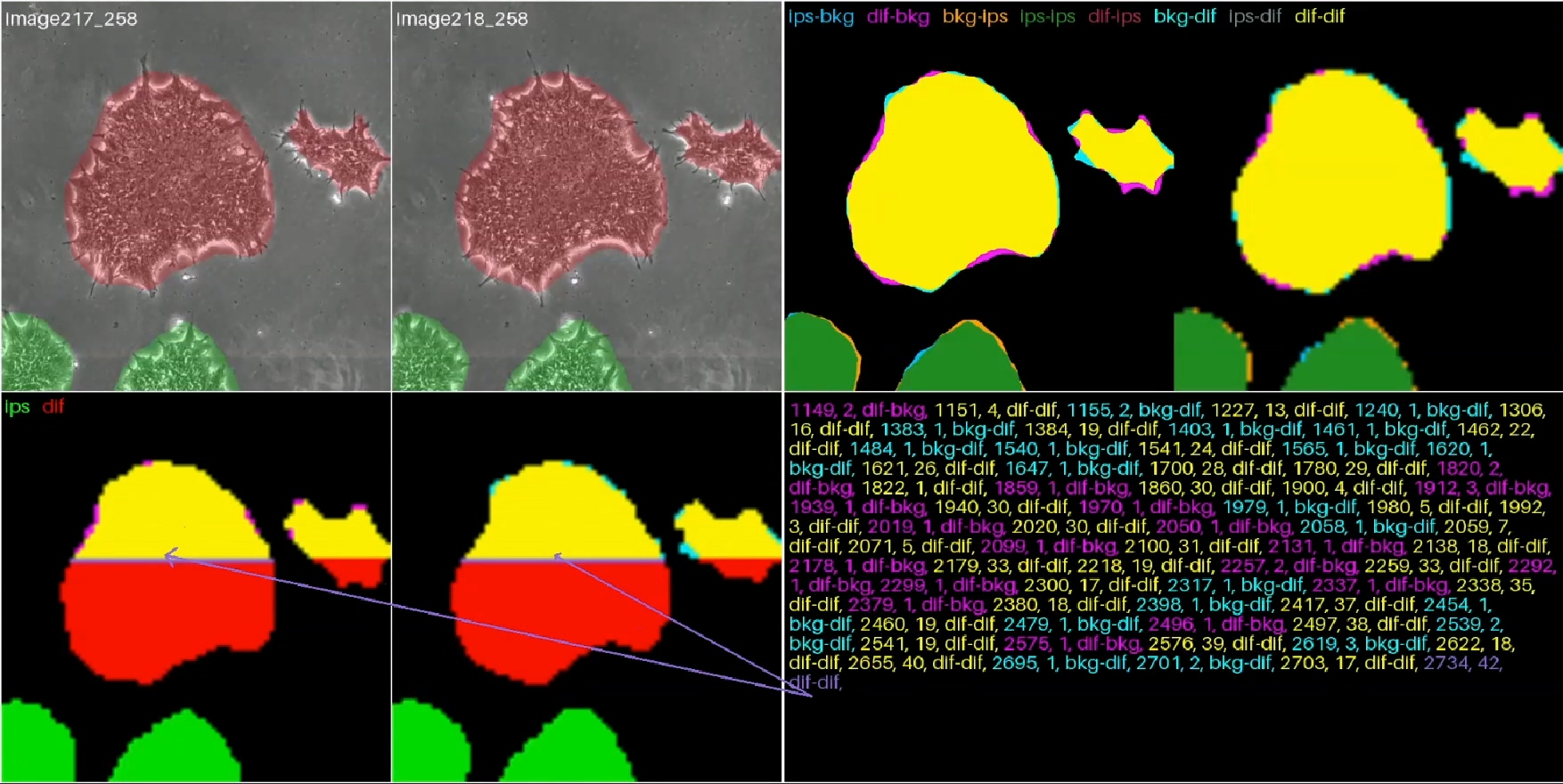}
	\caption{
		Visualization of TAC tokenization for multi-class video segmentation with $N=2$.
		The top row shows (from left to right) $F_1$, $F_2$, full resolution TAC mask, and subsampled TAC mask.
		The TAC masks show 8 TAC classes whose colors are shown at the top.
		The bottom row shows subsampled $F_1$ and $F_2$ masks, partially colored with TAC colors for runs whose tokens are shown on the right.
		Tokens are colored according to the TAC class in each run except the current one that is shown in purple.
		Animated version of this figure is available
		\href{https://webdocs.cs.ualberta.ca/~asingh1/p2s\#vid_seg_multi_class_tac}{here}.
		Best viewed under high magnification.
	}
	\label{fig:vid_seg_multi_class_tac}	
\end{figure}

A straightforward extension of RLE tokenization to perform semantic segmentation on a video with $N$ frames would involve flattening the $N\times S\times S$ 3D mask into a 1D vector of size $N\times S^2$ using either row-major or column-major ordering.

As shown in Fig. \ref{fig:vid_seg_binary_row_major}, row-major or C ordering (3D-C) results in the runs for individual video frames simply getting concatenated together, i.e., all the runs for $F_1$ come together in a sequence, followed by the runs for $F_2$ and so on (where  $F_i$, $1\leq i\leq N$, denotes the $i^{th}$ video frame).
This does not account for spatiotemporal consistencies in the masks since the runs corresponding to the same object from different video frames are completely unrelated.
Column-major or Fortran ordering (3D-F) partially accounts for spatiotemporal mask consistency but has large numbers of very short runs due to small changes in the object position or shape between consecutive frames.
This not only significantly increases the sequence length but the very short, often unit-sized, runs are also likely to be difficult to train on. 
An example is shown in Fig. \ref{fig:vid_seg_binary_column_major}. 
Another problem with the straightforward 3D flattening of video masks (either 3D-C or 3D-F) is that the number of \textit{starts} tokens increases linearly with $N$ by a factor of $S^2$.
This becomes infeasible for $N > 5$ with $S=80$ and $N > 1$ with $S=160$.

\subsection{Time-As-Class (TAC)}
\label{tac}

These issues can be largely resolved by extending the LAC encoding into Time-As-Class (TAC) tokenization to combine the temporal dimension with class IDs so that every possible combination of class IDs across the video frames is represented by a separate TAC token.
For example, Table \ref{tab:tac_tokens} shows TAC tokens for $N=2$ and $N=3$ for both binary and multi-class cases.
Due to the combinatorial nature of TAC tokenization, the total number of TAC tokens becomes $(C+1)^N-1$.
Even though this increases exponentially with $N$, $V$ remains practicable for up to $N=14$ and $N=9$ respectively for binary and multi-class cases with $S=80$ (Table \ref{tab:vocab_80}) and upto $N=12$ and $N=6$  with $S=160$ (Table \ref{tab:vocab_160}).
This is aided by the fact that the number of \textit{starts} tokens now becomes independent of $N$ since the 3D video mask with $C$ classes has effectively been collapsed into a 2D mask with $(C+1)^N-1$ classes. 
Fig. \ref{fig:vid_seg_multi_class_tac}, \ref{fig:vid_seg_binary_tac} and \ref{fig:vid_seg_tac_len_3} show examples of TAC tokenization for both binary and multi-class cases with $N=2$ and $N=3$.

\begin{table}[t]
	\centering
	\caption{
		TAC tokens for IPSC dataset with binary and multi-class masks for $N=2$ and $N=3$.
		Here, \bkgblue{} refers to the background while \ipsgreen{} and \difred{} are the two classes in the IPSC dataset.
		Both classes are represented by \cellyellow{} in the binary case.
		Note that there is no TAC token corresponding to the class combinations with \bkgblue{} in every frame (i.e., \bkgblue-\bkgblue{} for $N=2$ and \bkgblue-\bkgblue-\bkgblue{} for $N=3$) since RLE only represents foreground pixels, which requires that atleast one class be non-\bkgblue{}.
		Also, $N=3$ multi-class case would have $(C+1)^N-1 = 3^3 - 1 = 26$ TAC tokens, out of which only $15$ are shown for brevity.
	}
	\begin{tabular}{c}
		\centering
		\includegraphics[width=0.47\textwidth]{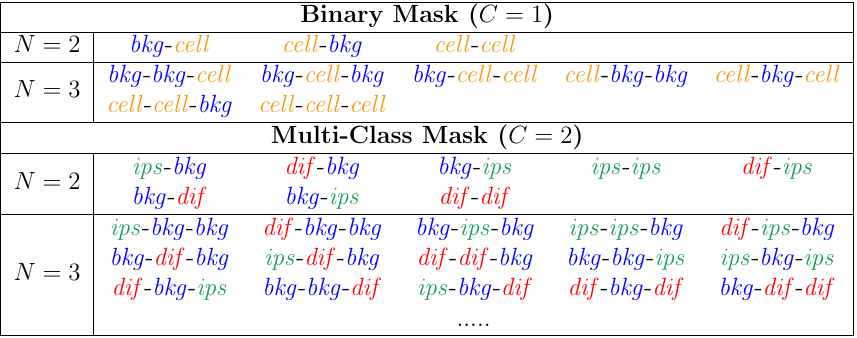}
	\end{tabular}	
	\label{tab:tac_tokens}
\end{table}

\subsection{Length-and-Time-As-Class (LTAC)}
\label{ltac}
TAC and LAC techniques can be combined to represent each run with only 2 tokens for multi-class video masks too.
However, the number of LTAC tokens = $S\times (C+1)^N-1$ becomes impractical for $N > 8$ and $N > 5$ respectively for binary and multi-class cases.
Fig. \ref{fig:vid_seg_ltac} shows examples of LTAC tokenization for both cases.

\subsection{Class-wise (CW) Tokenization}
\label{cw_iw}
As mentioned above, TAC and LTAC (and to a lesser extent LAC) tokenization schemes suffer from the problem of exponential increase in $V$ when $C$ becomes high (e.g., $C=80$ in COCO dataset \cite{coco_dataset}).
This can be ameliorated by decoupling the class ID from the RLE sequence so that the latter is composed only of \textit{starts} and \textit{lengths}.
A simple way to achieve this is to first generate RLE tokens for the binary mask of each class and then concatenate these RLE sequences, separated by the respective class tokens to mark the end of each sequence.
The class tokens would therefore serve the dual purpose of separating the RLE sequences for the different classes and specifying the classes themselves.
This allows the number of classes for the purpose of computing $V$ for these tokenization schemes to remain constant at $C=2$, irrespective of the actual number of classes in the dataset.

On the flip side, this representation at least partially nullifies the advantage of small quantum that RLE provides as far as classification accuracy is concerned.
When each run is classified separately, even if a few of them are misclassified, it does not affect the overall classification accuracy greatly.
However, when all the runs corresponding to a class are classified by a single token, an error in the latter causes all of those runs to become misclassified.
Related to this is the fact that the class tokens are greatly outweighed by the coordinate tokens since there are anywhere from a few tens to a few hundreds of coordinate tokens for every class token.
Based on our experiments with video detection models \cite[Sec. V.A ]{singh25_p2s_vid_det_access}, a promising way to handle this last problem is by increasing the weight of the class token \cite[Sec. V.A ]{singh25_p2s_vid_det_access} so that all the coordinate tokens combined have about the same weightage as the single class token.
However, our early experiments with class weight equalization for segmentation have shown only minor improvement so we have excluded these here and leave more comprehensive experiments for future work.

\subsection{Instance-wise (IW)  Tokenization}
\label{iw}
If object instance information is available, as in the IPSC dataset, RLE sequences can be generated for the binary masks corresponding to each instance instead of each class.
These sequences can again be concatenated, separated by the respective class tokens, similar to CW tokenization.
This representation allows us to perform instance segmentation in addition to semantic segmentation and therefore achieve full video panoptic segmentation.
This also partially solves the problem of weight imbalance between coordinate and class tokens because
the number of runs required to represent each object instance is usually much smaller than those needed to represent all the objects belonging to each class.
Fig. \ref{fig:seg_iw} shows an example of IW tokenization for static segmentation masks.

Early experiments with this scheme on static masks have shown it to be comparable to the other tokenization schemes in terms of semantic segmentation performance.
However, its instance segmentation performance does not compare favorably with object detection models, mainly because of the low resolution of the mask.
Training with sufficiently high batch sizes is currently only possible with $S=80$ and $S=128$ using respective image sizes of $I=640$ and $I=1024$.
Given the high resolution of the IPSC images, this requires the original masks to be downsampled by a factor of anywhere from $10$ to $50$ which is far too large to be able to extract spatially accurate bounding boxes.
We have trained a couple of models with $S=512$ and $S=640$ too but we had to use batch sizes that are too small to train these successfully.
As a result, these models underperform in terms of both semantic and instance segmentation.
We leave further exploration of both CW and IW tokenization schemes as part of future work when more GPU memory is available.

\section{Results}
\label{results}
This section presents results comparing our proposed token-based
semantic segmentation models with the conventual deep learning models from \cite{Singh2020_River_ice} and \cite{Singh23_ipsc}.
Unless otherwise specified, all of our models have $640\times 640$ ResNet-50 as their backbone and the video models use the middle-fusion video architecture \cite[Sec. III.B.2]{singh25_p2s_vid_det_access} with $N=2$.
For the sake of brevity, the static and video semantic segmentation models are abbreviated as \textbf{P2S-SEG} and \textbf{P2S-VIDSEG} respectively for the remainder of this paper.
We have used the same training setup and validation protocols as detailed in \cite{singh25_p2s_vid_det_access} so we will not be repeating these here and instead refer the interested reader to Sec. IV.C of that paper for these details.

\subsection{Datasets}
\label{datasets}
We have evaluated our models on the following two datasets:
\begin{itemize}[left=0pt,label=\textendash]
	\item Alberta River Ice Segmentation (ARIS) dataset \cite{Singh2020_River_ice}: We have tested on the standard configuration with 32 training and 18 test images.
	We have also performed image-level ablation tests with 4, 8, 16 and 24 training images, all of these being tested on the same 18 test images. 	
	This dataset does not contain video labels so we have only used it for testing the static segmentation models.
	\item Induced Pluripotent Stem Cell (IPSC) dataset \cite{Singh23_ipsc}: We have tested on both early and late-stage training configurations. Unlike ARIS, this dataset does contain video labels so we have used it for testing both static and video segmentation models.
\end{itemize}
We have also done some experiments on larger datasets like COCO \cite{coco_dataset} and Cityscapes \cite{cityscapes_dataset} but found that the results on these are not yet good enough to include here (Sec. \ref{future_work_datasets})
and are thus left for a future work.

\subsection{Metrics}
\label{metrics}
We have used three main metrics for measuring semantic segmentation performance.
The first two are recall and precision as defined in \cite{Singh2020_River_ice} and redefined below.
These are used for both ARIS and IPSC datasets.
\begin{itemize}
	\item Recall:
	\begin{align}
		\label{chap_3:eq_mean_acc}
		\texttt{rec}
		=
		\frac{1}{C}
		\sum_{i} \frac{n_{ii}}{t_i}
	\end{align}

	\item Precision:
	\begin{align}
		\label{chap_3:eq_mean_iou}
		\texttt{prec}
		=
		\frac{1}{C}
		\sum_{i} \frac{n_{ii}}{t_i + \sum_{j}n_{ji} - n_{ii}}
	\end{align}
	\item Frequency Weighted Recall:
	\begin{align}		
		\label{chap_3:eq_pix_acc}
		\texttt{fw\_rec}
		=
		\frac{\sum_i n_{ii}}{\sum_i t_{i}}
	\end{align}
	\item Frequency Weighted Precision:
	\begin{align}
		\label{chap_3:eq_fw_iou}
		\texttt{fw\_prec}
		=
		\sum_{k} (t_k)^{-1}	
		\sum_{i} \frac{t_i n_{ii}}{t_i + \sum_{j}n_{ji} - n_{ii}}
	\end{align}
\end{itemize}
The third one
is the Dice score,
also known as the Dice-Sørensen coefficient
\cite{Dice1945_Dice_score,Srensen1948_dice_score},
which is used only for the IPSC dataset.
Dice score is widely used
as a single metric to represent the overall segmentation quality
by incorporating both recall and precision.
It is defined as:
\begin{align}
	\label{eq_dice_score}
	\texttt{dice}_i
	=
	\frac{2\times n_{ii}}{t_i + \sum_{j}n_{ji}}
\end{align}
where $ 1\leq i\leq C $ is the class index, $ n_{ij} $ is the number of pixels of class $ i $ predicted to belong to class $ j $ and  $ t_i $ is the total number of pixels of class $ i $ in the ground truth.
We have also used the domain-specific ice concentration mean absolute error (MAE) \cite{Singh2020_River_ice} for the ARIS dataset.

\subsection{Performance Overview}
\label{res_overview}

\begin{table*}[t]
	\centering
	\caption{Segmentation recall, precision and ice concentration median MAE on ARIS dataset for P2S-SEG and conventional deep learning models along with SVM.
		The \textit{fw} in \textit{ice+water (fw)} stands for frequency weighted.
		Relative increase over SVM in recall and precision is computed as $ (\texttt{model\_value} - \texttt{svm\_value})/\texttt{svm\_value}\times 100 $ while the relative decrease in median MAE is computed as $ (\texttt{svm\_mae} - \texttt{model\_mae})/\texttt{svm\_mae}\times 100 $.
		The best and the second best models in each case are shown in bold and highlighted in green and yellow respectively.
		This data is shown as a bar plot in Fig. \ref{617_summary_bar}.
	}
	\begin{tabular}{c}
		\includegraphics[width=\textwidth]{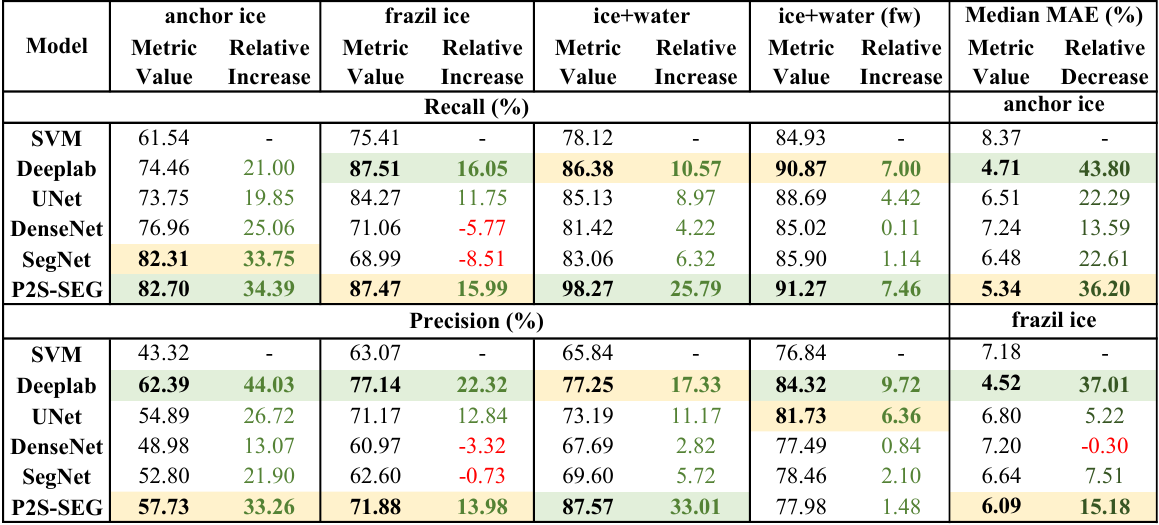}
	\end{tabular}	
	\label{tab:617_summary_single_p2s}
\end{table*}

We experimented with many different configurations or variants of P2S-SEG and P2S-VIDSEG (Sec. \ref{exp}) but this section only summarizes the best results we found.
The specific model configurations that we have included here for each dataset are detailed in Table \ref{tab:model_configs}.
Note that we were only able to train a small fraction of all the models we would have liked to have trained due to limited time and computational resources, so these results very likely do not indicate the best performance that these models are capable of, especially in the case of the video models with larger values of $N$.
\subsubsection{Summary}
\label{res_summary}
Following are the key takeaways from the results presented in the remainder of this paper:
\begin{itemize}[left=0pt,label=\textendash]
	\item Both static and video language models perform about the same as similarly sized conventional models.
	As with deep learning in general, the overall performance depends more on the size of the backbone than any specific output modeling.	
	\item P2S-SEG and P2S-VIDSEG models compare more favourably against conventional segmentation models on ARIS dataset than on IPSC dataset.
	\item Language models are better at localizing objects than classifying them correctly so that their class-agnostic performance tends to compare more favourably with conventional models than their overall performance.
	\item Related to the last point is that the language models are relatively less robust to class imbalance and tend to overfit to the more numerous class.
	This causes P2S-SEG and P2S-VIDSEG to compare more favourably against conventional models in terms of segmentation recall rather than precision. 
	\item P2S-VIDSEG does not show any consistent improvement over P2S-SEG, probably because the stride-based redundancy advantage does not apply to semantic segmentation.
	\item There is no consistent improvement performance with increase in $N$  (Sec. \ref{exp_vid_len}).
	\item Static models trained to predict video outputs by processing only the first frame in each video temporal window  (Sec. \ref{exp_arch_static}) are able to keep up with the video models surprisingly well, even for large values of $N$, indicating that the latter are not able to make sufficient use of the video information.
\end{itemize}

\subsubsection{ARIS}
\label{seg_aris}
\begin{figure}[t]
	\centering
	\includegraphics[width=0.235\textwidth]{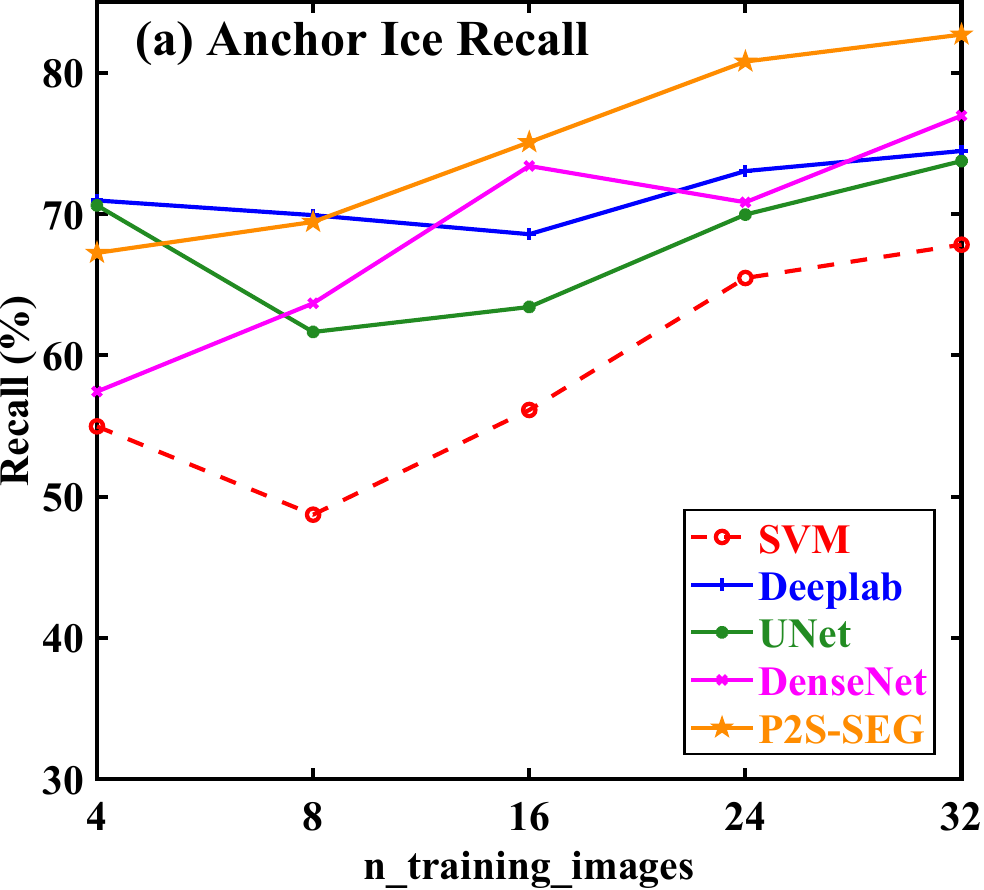}
	\includegraphics[width=0.235\textwidth]{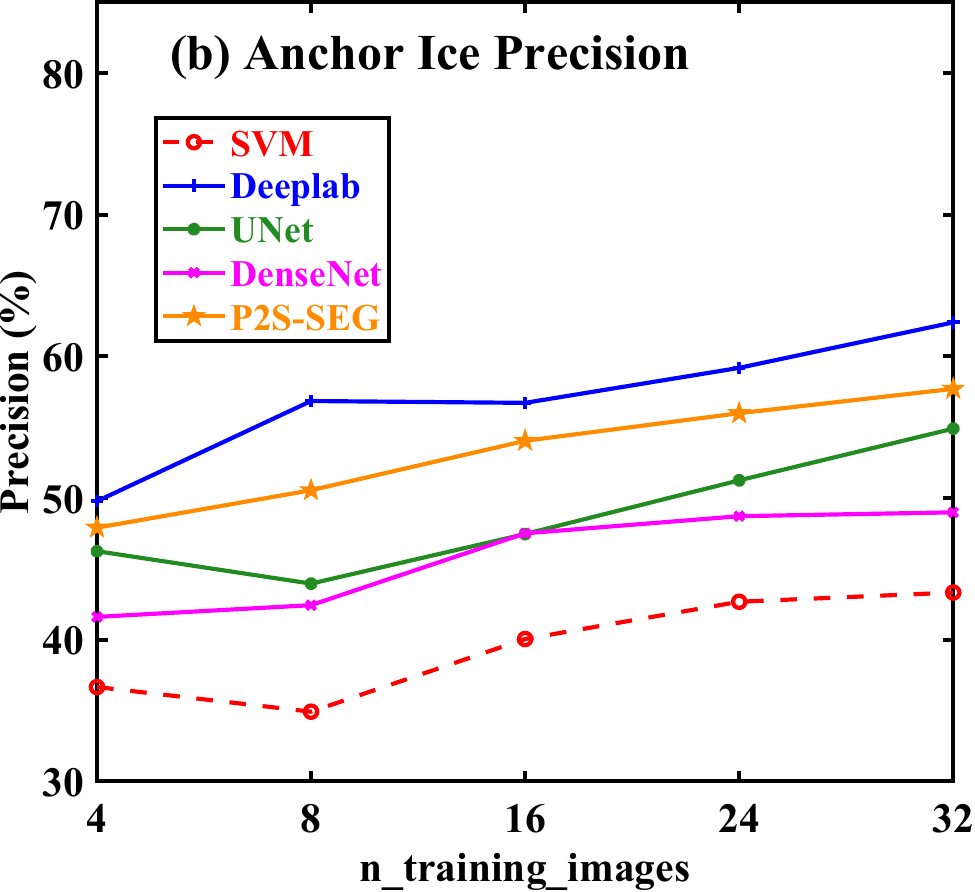}
	\includegraphics[width=0.235\textwidth]{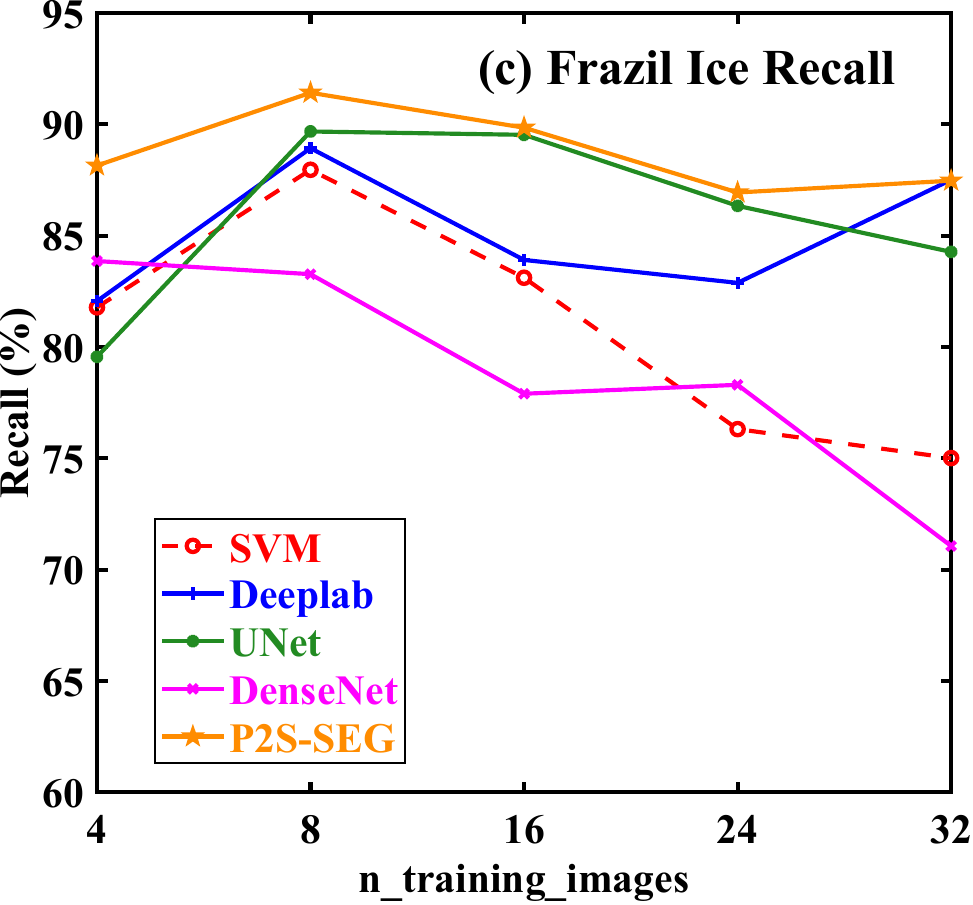}
	\includegraphics[width=0.235\textwidth]{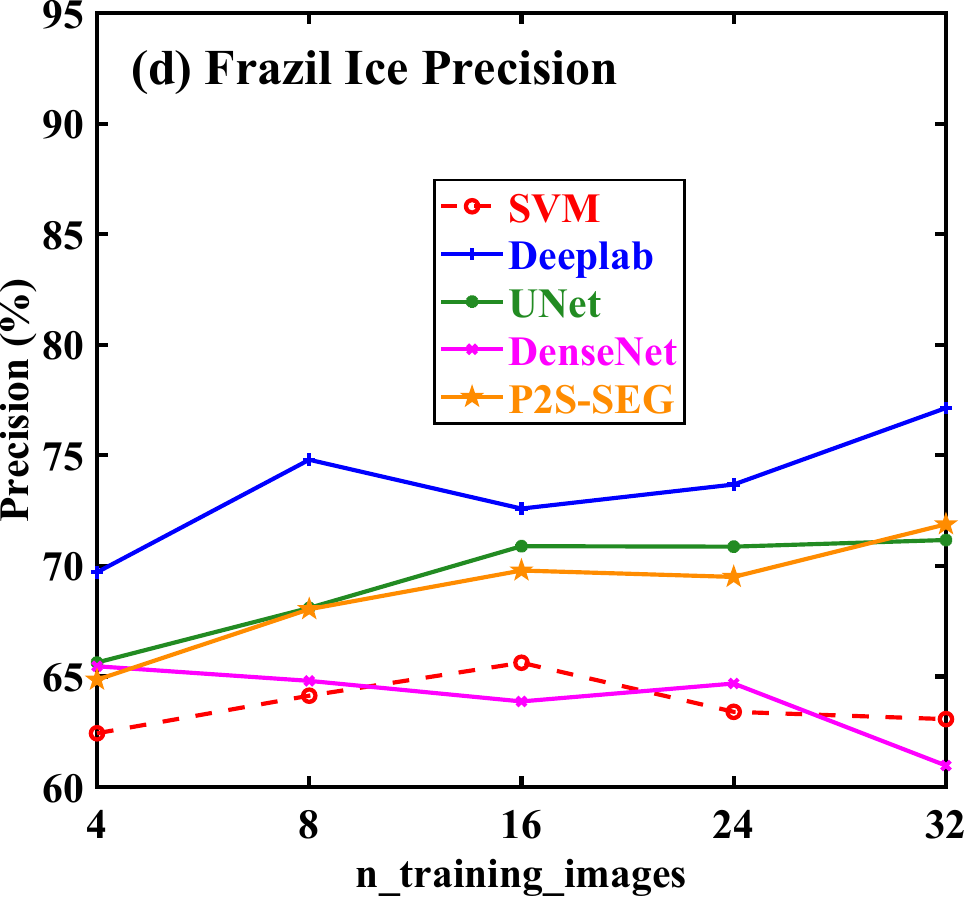}
	\caption{
		Results of ablation tests with training images on ARIS dataset for (a-b) anchor ice and (c-d) frazil ice.
		Note the variable Y-axis limits between the top and bottom plots.
	}
	\label{617_ablation}
\end{figure}

Table \ref{tab:617_summary_single_p2s} presents a summary of results on the ARIS dataset.
P2S-SEG performs remarkably well here and turns out to be either the first or the second best model in nearly every case.
Language modeling is particularly effective at class agnostic tasks, as represented by the ice+water metrics, where it outperforms all the other models by a large margin.
Somewhat paradoxically, its performance drops significantly at the frequency weighted version of this metric, especially in terms of precision.
All other models find this metric easier since it gives greater weightage to water which constitutes a majority of these images \cite[Table II]{Singh2020_River_ice} and is relatively easy to separate from ice.
Fig. \ref{617_ablation} shows the results of image-level ablation testing on this dataset.
P2S-SEG remains the best model in terms of recall and among the best two models in terms of precision.
It predictably  finds frazil ice precision most difficult to handle since there is significantly more frazil ice than anchor ice in the training images.
This causes the model to overfit to this class and misclassify anchor ice as frazil ice which in turns lowers the corresponding precision.

\subsubsection{IPSC}
\label{seg_ipsc}
\begin{figure*}[!t]
	\centering
	\includegraphics[width=0.32\textwidth]{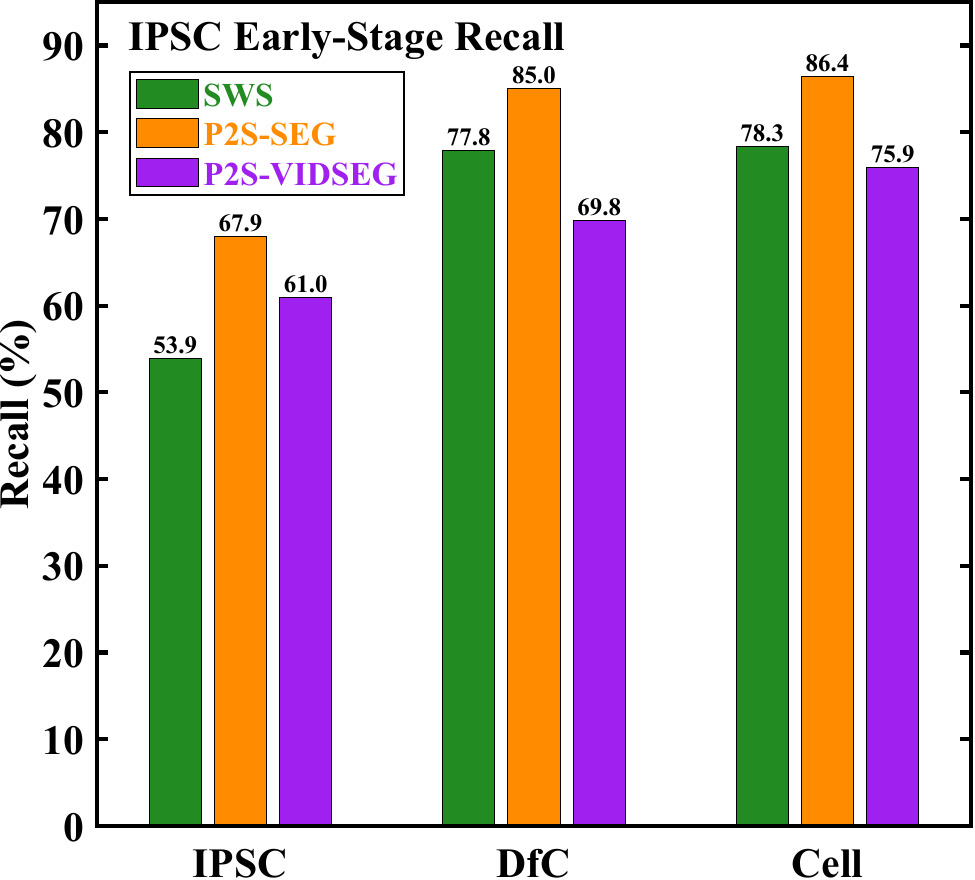}
	\includegraphics[width=0.32\textwidth]{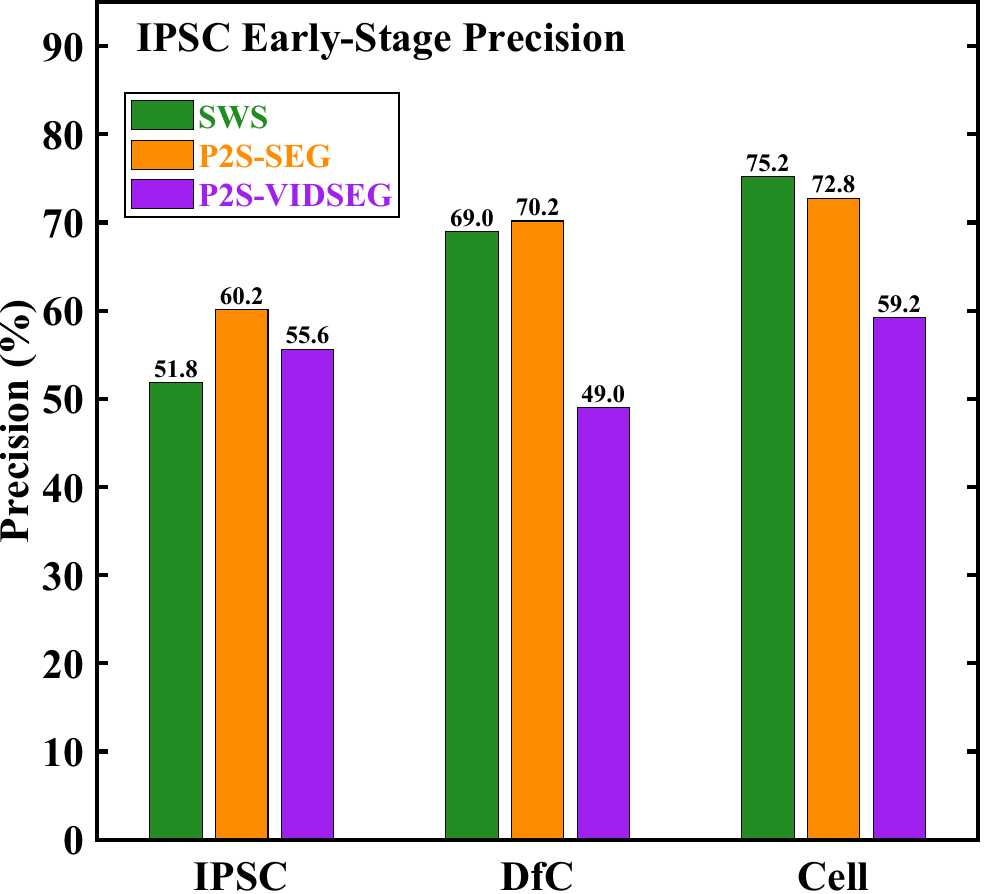}
	\includegraphics[width=0.32\textwidth]{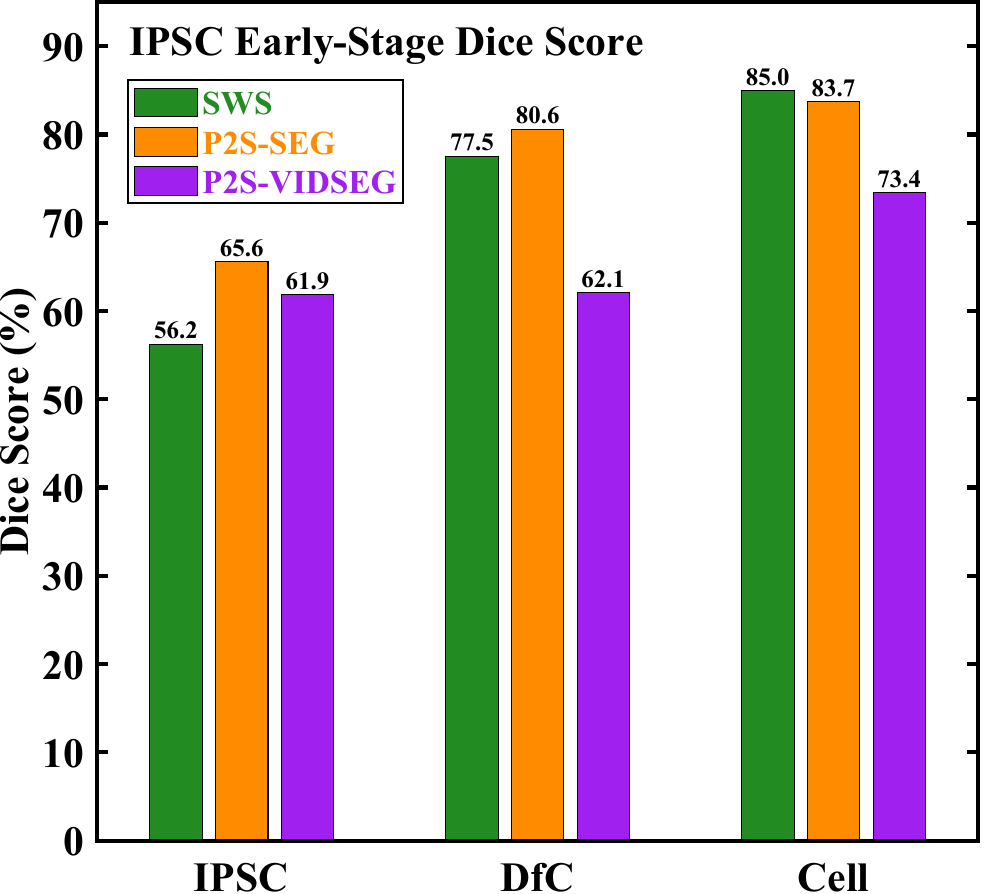}
	\includegraphics[width=0.32\textwidth]{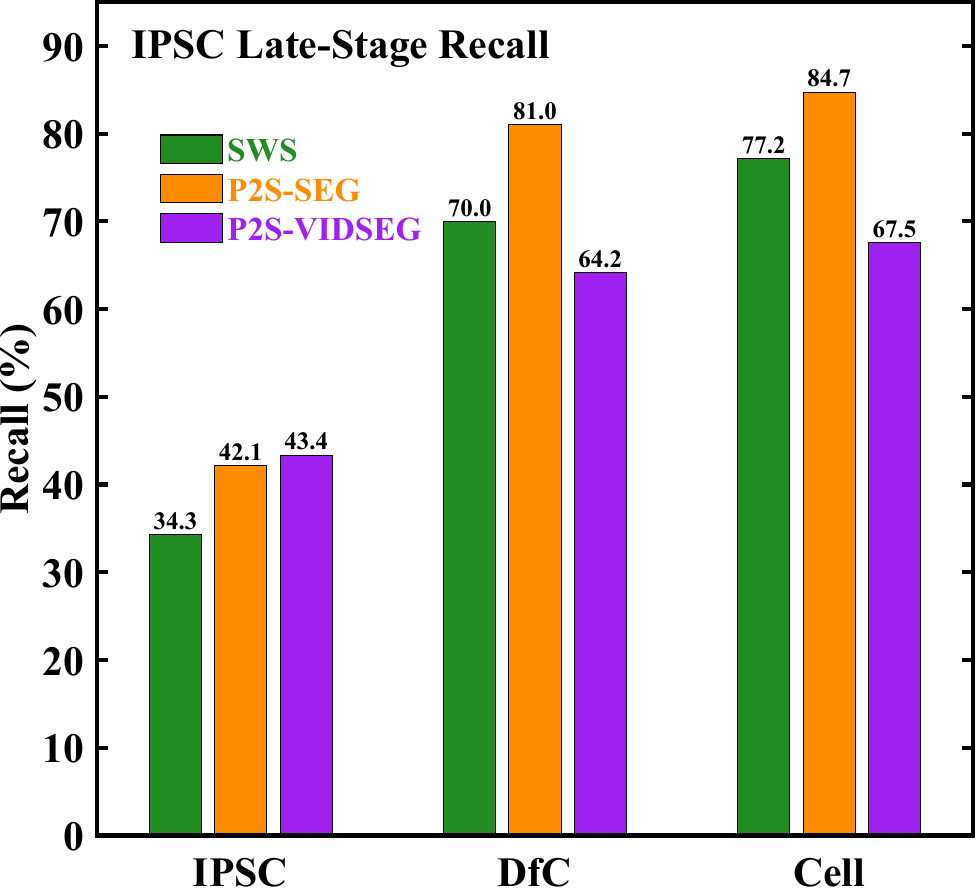}
	\includegraphics[width=0.32\textwidth]{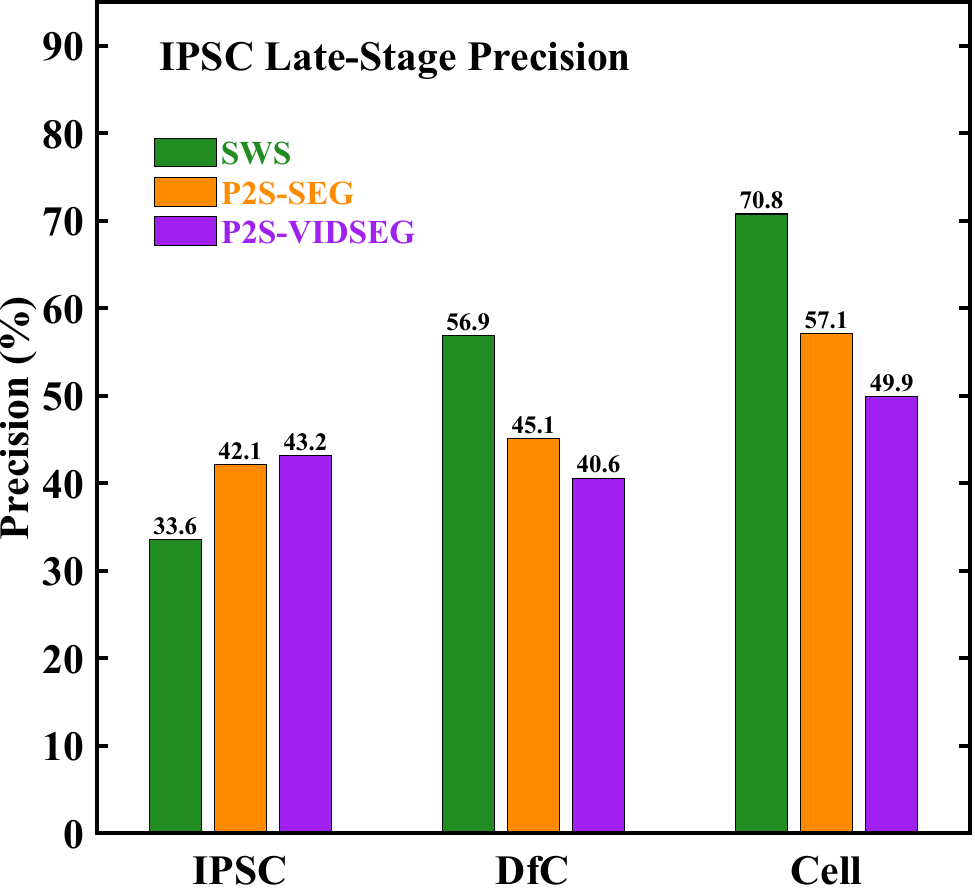}
	\includegraphics[width=0.32\textwidth]{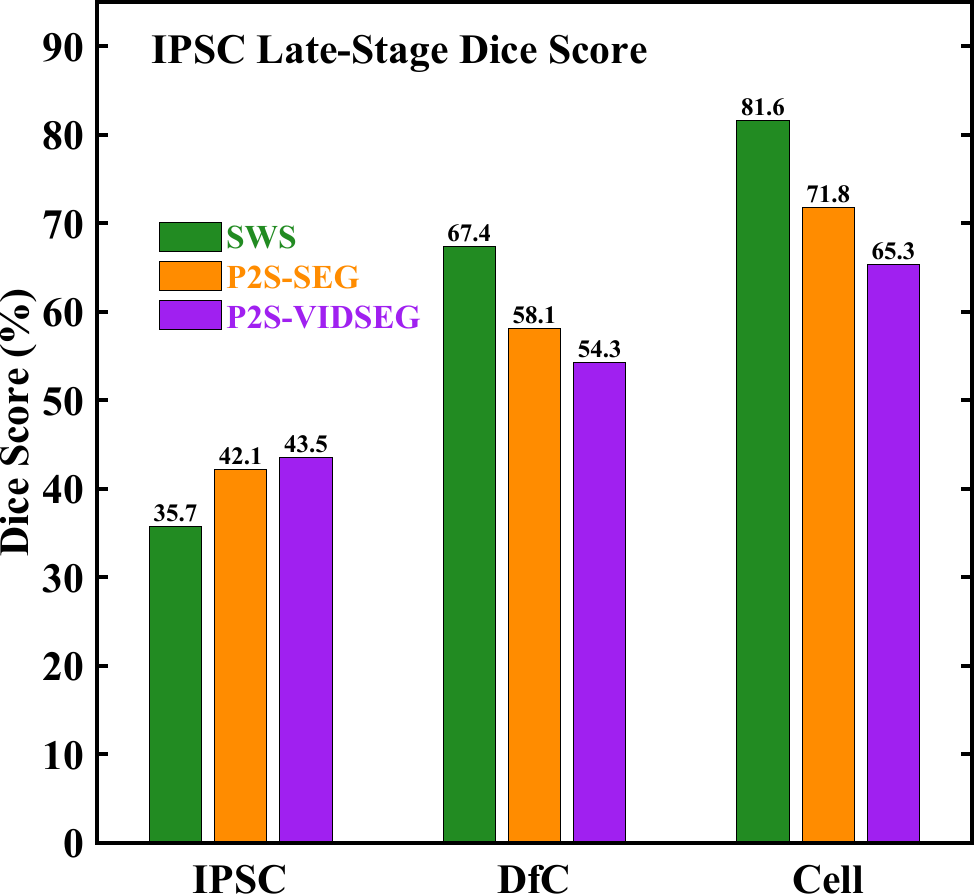}
	\caption{
		Semantic segmentation results on both (top) early  and (bottom) late-stage IPSC datasets in terms of (left to right) recall, precision and dice score.
		\textit{IPSC} and \textit{DfC} on the x-axis refers to the two classes of cells while \textit{Cell} represents the class-agnostic case where all the cells are considered as belonging to the same class.
	}
	\label{ipsc_seg}
\end{figure*}
\begin{figure}[t]
	\centering
	\includegraphics[width=0.47\textwidth]{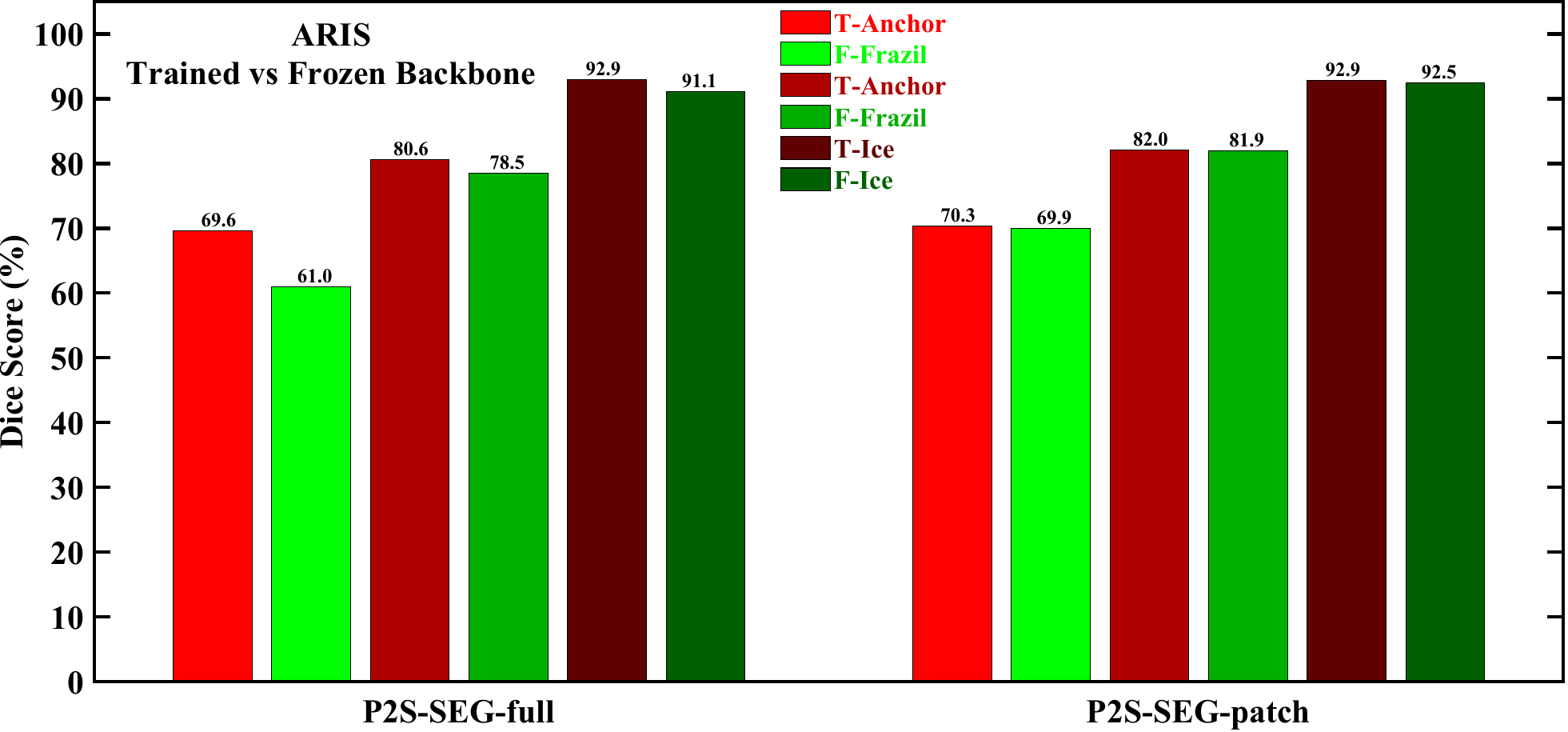}
	\includegraphics[width=0.47\textwidth]{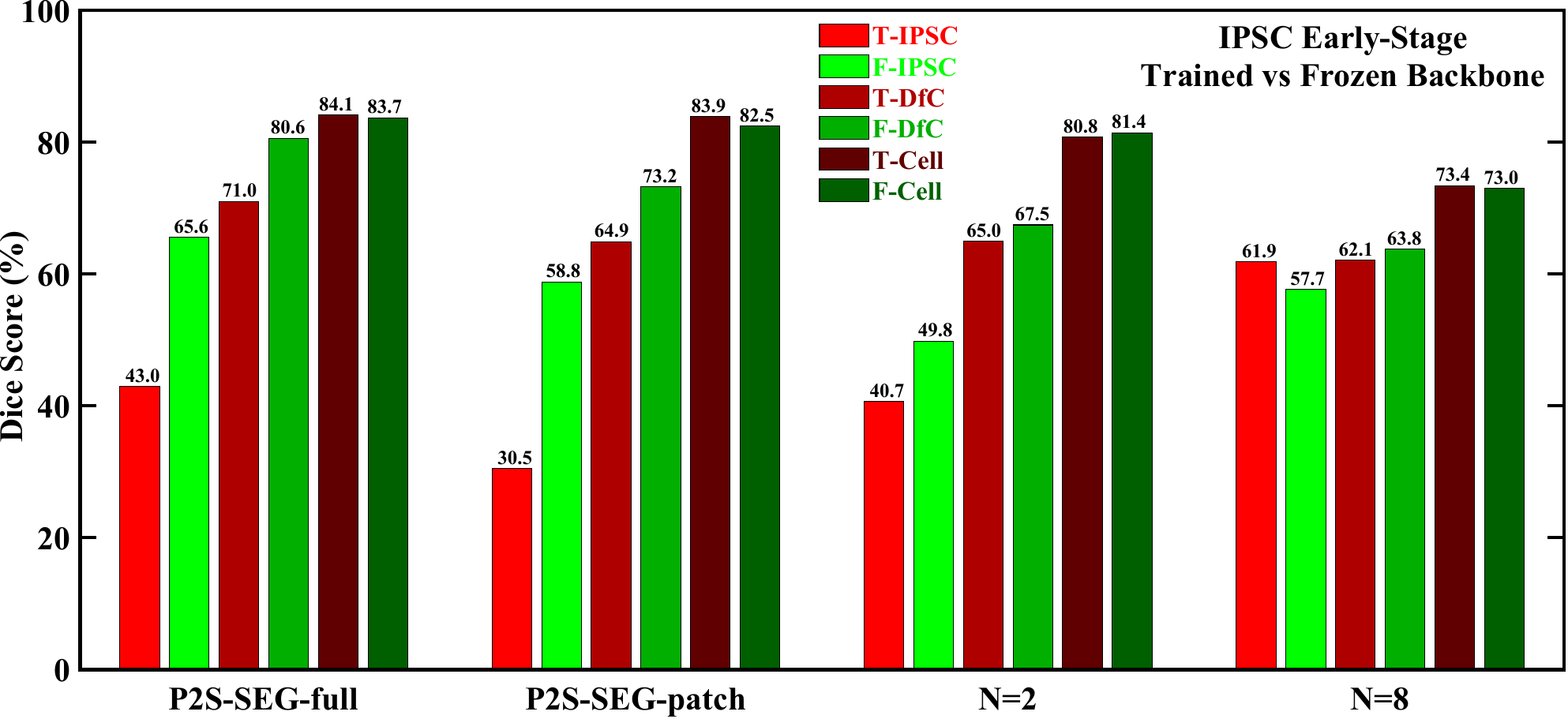}
	\caption{
		Comparing the segmentation performance of models trained with and without frozen backbones on the (top) ARIS and (bottom) IPSC early-stage datasets.
		Models trained with and without frozen backbones are respectively shown in shades of green and red and denoted in the legend with prefixes \textit{F} and \textit{T}.
		The P2S-SEG suffixes \textit{patch} and \textit{full} respectively refer to models trained with and without sliding window patches.
		In the former case, $P=640$, $S=80$ for both datasets, $I=1280$ for ARIS and $I=2560$ for IPSC.
		In the latter case, $I=P=640$ for both datasets, $S=160$ for ARIS and $S=320$ for IPSC.
	}
	\label{fbb_seg}
\end{figure}
\begin{figure}[t]
	\centering
	\includegraphics[width=0.23\textwidth]{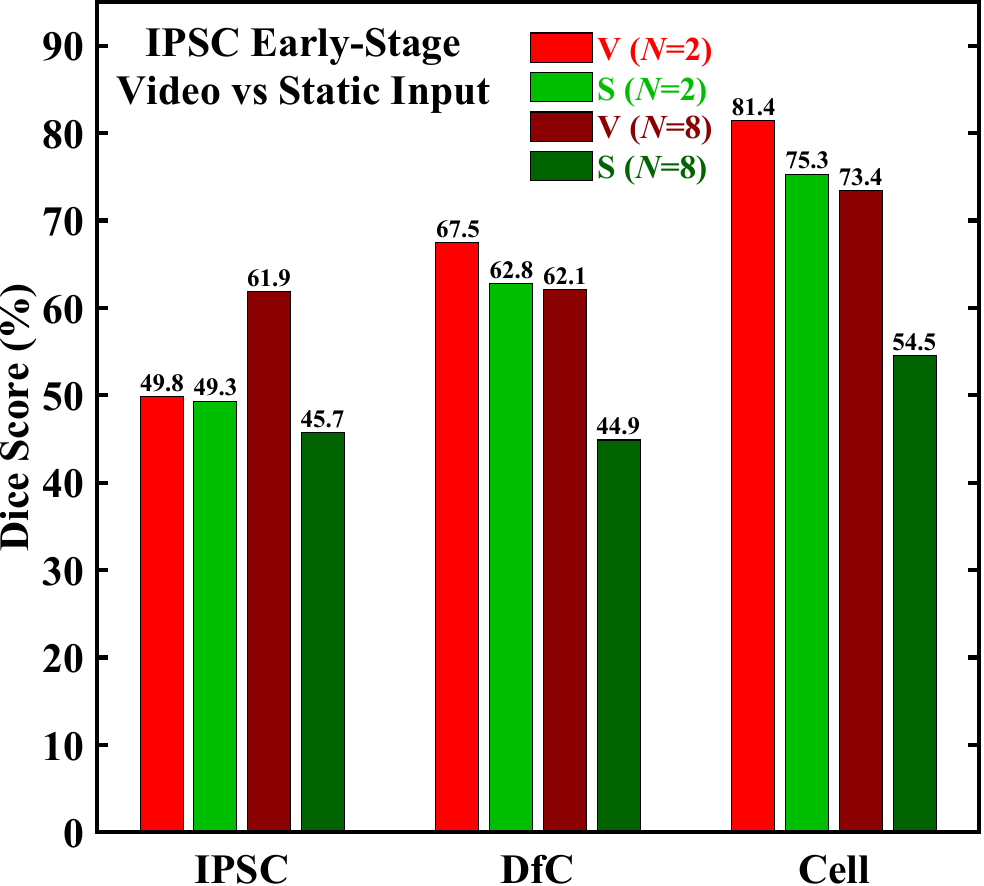}
	\includegraphics[width=0.23\textwidth]{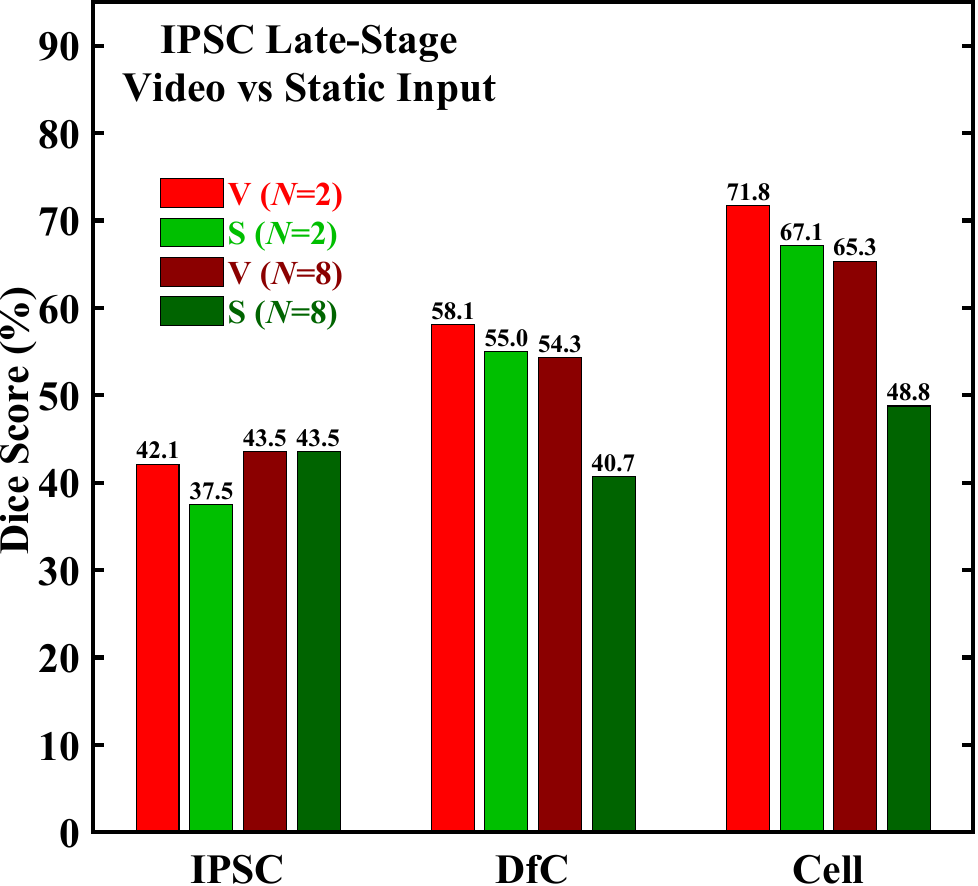}
	\caption{
		Performance impact of replacing $N$ video frames with only the first frame in the sequence as input to P2S-VIDSEG models.
		The two cases are denoted with \textit{V} and \textit{S} in the legend and shown in shades of red and green respectively.
		Results are shown for both (left) early and (right) late-stage IPSC datasets in terms of Dice score
		Recall and precision exhibited the same trends as Dice score and have therefore be relegated to Fig. \ref{app:static_vid_len_seg}.
		Best viewed under high magnification.
	}
	\label{static_vid_seg}
\end{figure}
\begin{figure}[t]
	\centering
	\includegraphics[width=0.235\textwidth]{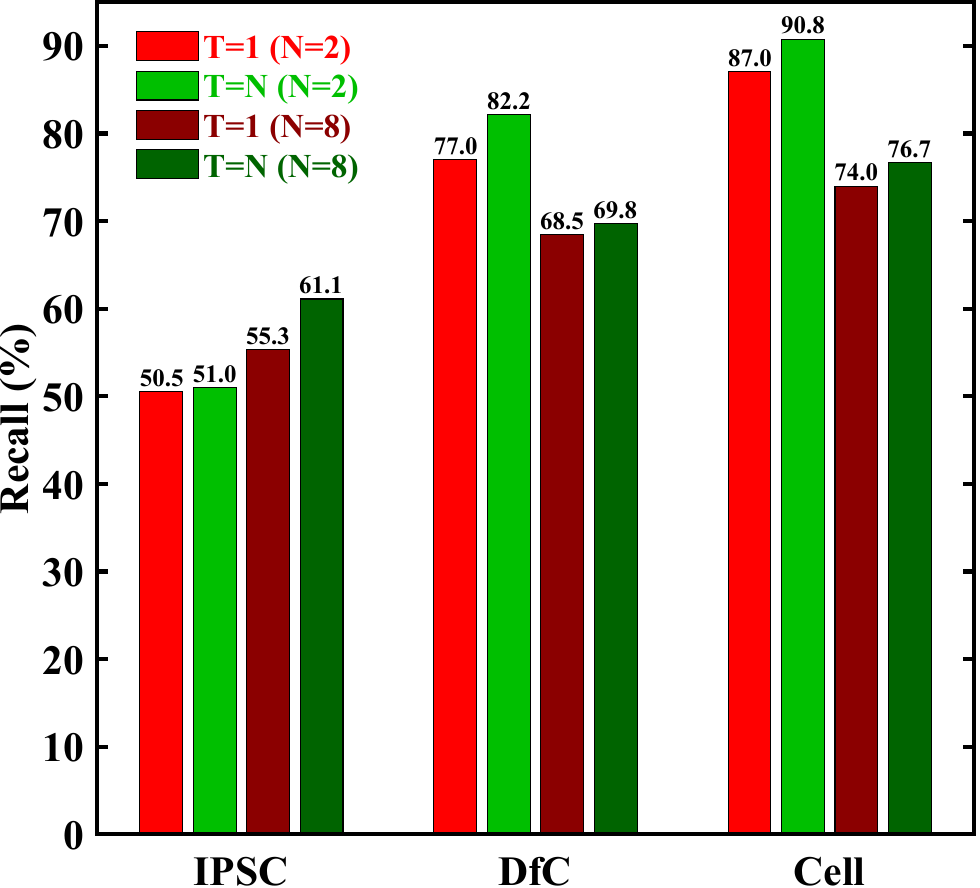}
	\includegraphics[width=0.235\textwidth]{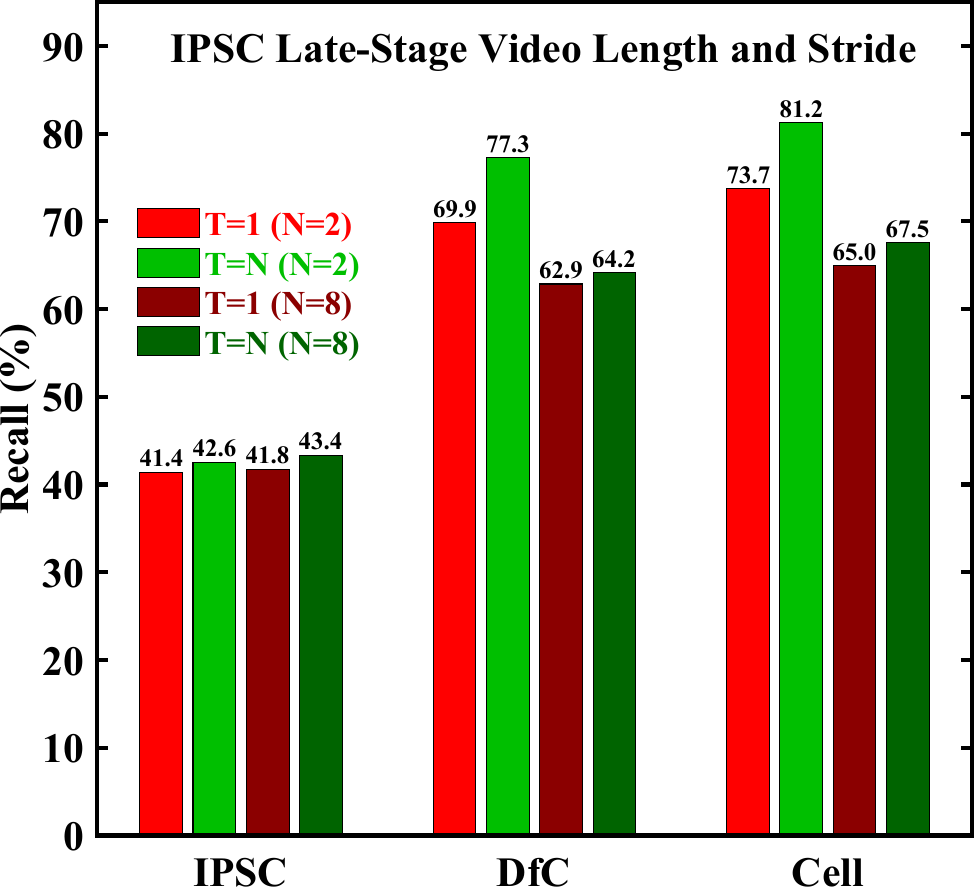}
	\includegraphics[width=0.235\textwidth]{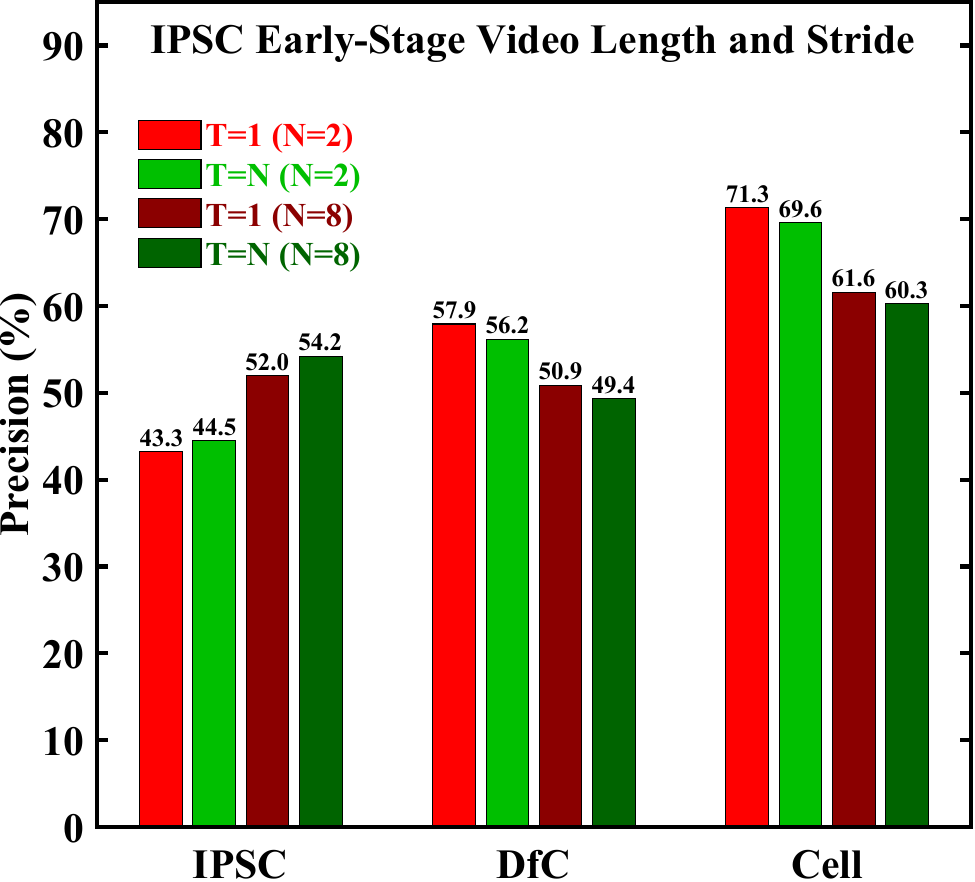}
	\includegraphics[width=0.235\textwidth]{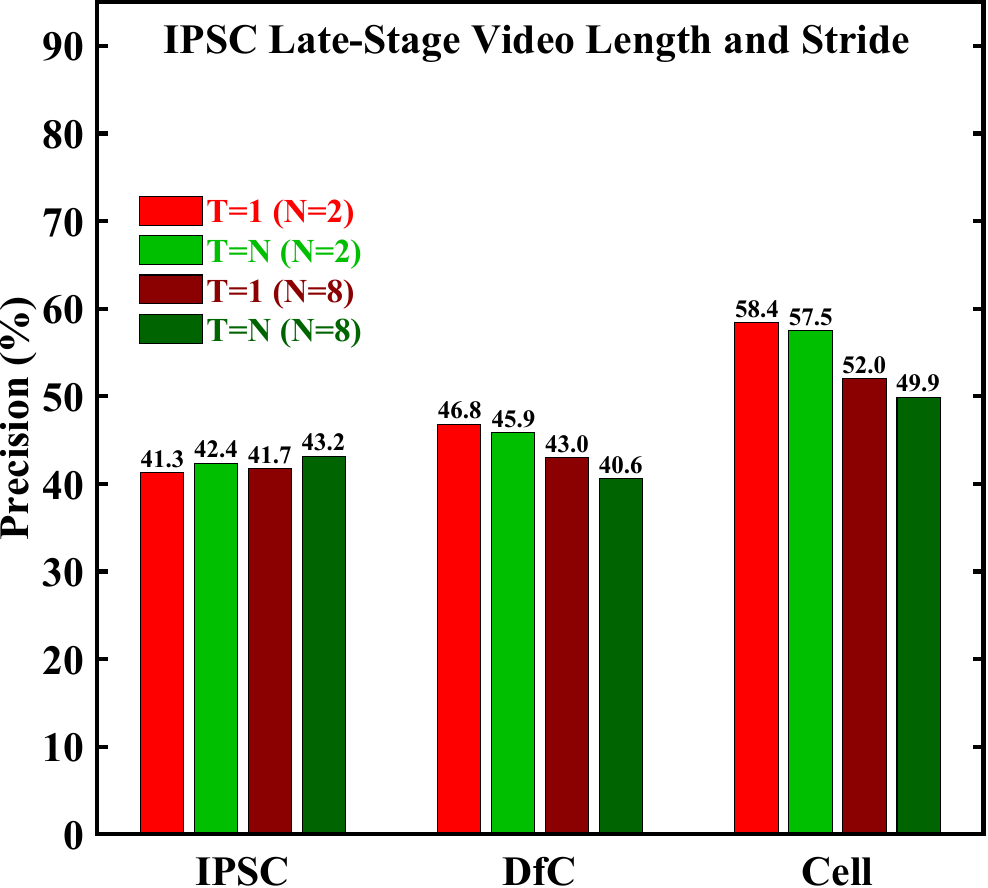}
	\caption{
		Impact of video length $N$ and stride $T$ on P2S-VIDSEG performance over (left) early and (right) late-stage configurations of the IPSC dataset.
		$T=1$ and $T=N$ are respectively represented with shades of red and green.
		Lighter shades of each colour represent $N=2$ while the darker shades represent $N=8$.
		Dice score showed similar patterns as precision and has thus been relegated to Fig. \ref{app:static_vid_len_seg}.
		Best viewed under high magnification.
	}
	\label{vid_len_seg}
\end{figure}
Fig. \ref{ipsc_seg} shows the segmentation results on both early and late-stage training configurations of the IPSC dataset.
Conventional modeling is represented here by the single Swin Transformer model  \cite{Liu2021_Swin_Transformer} denoted as SWS which can be taken to represent the current state of the art in semantic segmentation.
Language modeling compares less favourably against conventional modeling on this dataset than on ARIS,
possibly because SWS is a newer and bigger transformer-based model as opposed to the older CNN-based models used on ARIS.
Even so, P2S-SEG performs about the same as SWS overall, being slightly better on the early-stage dataset and slightly worse on the late-stage variant.
Similarly, P2S-SEG is notably better on \textit{IPSC} class while SWS is better on \textit{DfC} and \textit{Cell}.
Finally, P2S-SEG fares better in terms of recall while SWS has an edge on precision.
P2S-SEG also outperforms P2S-VIDSEG in nearly every case except the most challenging case of \textit{IPSC} recognition on the late-stage dataset, where the latter outperforms both P2S-SEG and SWS on all three metrics.
\subsubsection{Caveat}
\label{seg_caveat}
Semantic segmentation is very difficult to evaluate quantitatively and the performance numbers exhibit complex trade-offs between the various metrics as well as between the various classes.
A couple of example are provided in Tables \ref{tab:seg_details_ipsc_early_static} and \ref{tab:seg_details_ipsc_late_vid} that list the values of these three metrics (along with some others from \cite{Singh2020_River_ice}) on the validation set during the training runs for P2S-SEG on early-stage and P2S-VIDSEG on late-stage dataset respectively.
We chose the specific checkpoints for inclusion here by looking for a good balance between all these conflicting considerations among candidate checkpoints once the validation performance had plateaued over all the metrics.

When looking for the best compromise, we did have a bias in favour of \textit{IPSC} among classes (\textit{Anchor Ice} for ARIS) and Dice score among metrics, since these seem to us to best represent the overall segmentation quality.
Further, there are 2-3 times as many \textit{DfC}s as \textit{IPSC}s in the ground truth so models tend to perform better on \textit{DfC} (and \textit{Cell}) later in their training run when they start overfitting to the more numerous class, something we wanted to avoid.
A different choice of trade-offs can generate quite different comparative plots so these results should be interpreted guardedly.
We have not implemented a live validation pipeline \cite{singh25_p2s_vid_det_access} for SWS so the results shown here were generated using the latest checkpoint once the training curve had plateaued.
This is why SWS shows a consistent tendency to outperform P2S-SEG in terms of \textit{DfC} and \textit{Cell} but is outperformed by the latter in terms of \textit{IPSC}.
The best we can say with reasonable certainty is that the overall performance of language and conventional models is in the same ballpark, the former being slightly better at recall and the latter at precision.

\subsection{Ablation Study}
\label{exp}
This section presents the results of our experimentation with some of the important model parameters that were useful in finding the optimal models that are reported in Sec. \ref{res_overview}.
We used the IPSC late-stage dataset for most of these experiments since it is small enough to allow training a large number of models, while at the same time also being challenging enough to be able to discriminate between these models.

\subsubsection{Frozen Backbone}
\label{exp_fbb}
Fig. \ref{fbb_seg} shows that the benefit of keeping the backbone frozen is much less obvious in case of semantic segmentation compared to object detection \cite{singh25_p2s_vid_det_access}.
This might be explained by the fact that RLE tokenization differs from static detection tokenization a lot more than does video detection.
In fact, given this difference, it is remarkable that frozen backbone is mostly able to keep up with full network training even in case of P2S-VIDSEG with $N=8$.
This might at least partly be due to the bottleneck imposed by insufficient $B$, without which the full network training might well lead to much better performance.

\subsubsection{Video Output with Static Input}
\label{exp_arch_static}
We wanted to find out how much useful information the network is able to learn from video frames so we trained static models to predict video output using only the first frame $F_1$ in each video temporal window as input.
This means that the model is trained to produce the same output as P2S-VIDSEG model but it only has access to the first frame in each temporal window,  rather than all $N$ frames.
It therefore needs to use the first frame to predict the contents of the future $N-1$ frames in the sequence.
We trained models with $N=2$, $N=4$, $N=6$ and $N=8$, all with the backbone frozen, and all on the IPSC late-stage dataset.
Note that these static-video models can be trained with the same (and much larger) batch size as P2S-SEG, irrespective of $N$.
This gives them an advantage over the true video models whose batch size decreases linearly with $N$.
The video stride $T$ is important in evaluating these models since $T=1$ ensures that each frame would be the first frame in some temporal window so that the static input models can output valid boxes only for the first frame and still not be penalized during inference.
However, this is unlikely to happen in practice since the model is trained to output boxes for all $N$ frames and therefore will be penalized during training for learning a simple strategy like this.

P2S-VIDSEG results for both configurations of IPSC dataset are shown in Fig. \ref{static_vid_seg}.
Video input has more consistent and strongly marked performance advantage over static input here than with video detection \cite{singh25_p2s_vid_det_access}, especially for $N=8$.
In part, this is due to the absence of temporal redundancy here, since $N=T$ is used in all cases.
It might also be due to greater integration of the outputs corresponding to different frames within the same shared tokens in video segmentation through TAC (Sec. \ref{tac}), as opposed to frame-specific tokens employed by video detection.

\subsubsection{Video Length and Stride}
\label{exp_vid_len}
Fig. \ref{vid_len_seg} shows the impact of $N$ and $T$ on video segmentation.
As mentioned in Sec. \ref{redundancy}, combining redundant outputs from overlapping temporal windows is far less straightforward for segmentation than it is for detection.
These plots show the results of the voting strategy that we found to work best, though its overall performance impact is still minimal.
Non-redundant output ($T=N$) fares slightly better than redundant output ($T=1$) in terms of recall over all three classes.
$T=N$ is also better in terms of precision for \textit{IPSC} but $T=1$ has a slight edge over \textit{DfC} and \textit{Cell}.
Like all Pix2Seq models, P2S-VIDSEG models have a tendency to overfit to the more numerous class, which in this case corresponds to the background, followed by \textit{DfC}.
Combining pixel labels from multiple temporal windows therefore results in some of the pixels, mostly near the cell boundaries, getting misclassified as either background or \textit{DfC}.
Misclassification as background explains the drop in recall across the board while misclassification as \textit{DfC} explains the drop in precision for \textit{IPSC} along with concurrent increase for \textit{DfC} and \textit{Cell}.

In terms of video length, $N=2$ significantly outperforms $N=8$ for \textit{DfC} and \textit{Cell} but the latter is appreciably better for \textit{IPSC}, significantly more so over the early-stage datset and with $T=N$.
This can be partly attributed to the much greater batch size bottleneck on $N=8$ and partly to our bias in favoring \textit{IPSC} performance over \textit{DfC} (Sec. \ref{seg_caveat}) when selecting the checkpoints for inclusion here.

\section{Challenges and Future Work}
\label{future_work}
This is still
a work in progress and
we are continuing to explore
many
avenues for future improvement.
\subsection{Larger Datasets}
\label{future_work_datasets}
A major limitation of our existing encoding schemes is that they offer competitive performance only on relatively small datasets with few classes.
Experiments on larger datasets like COCO \cite{coco_dataset} and Cityscapes \cite{cityscapes_dataset} have shown a tendency for performance to fall sharply with increase in the size of the training dataset and the number of classes.
For example, we achieved $> 50\%$ dice score on Cityscapes (3K images, 19 classes) but this dropped to $< 25\%$ on COCO (118K images, 133 classes).
We have found that most of this performance drop can be attributed to two factors.
Firstly, the low resolution masks we are currently able to output ($S < 160$) to keep the sequence length manageable ($L < 4096$) causes loss of fine structures in the mask.
This is not so noticeable on IPSC and ARIS images due to the somewhat spheroid shapes of most cells and ice blocks, but it has a significant impact on real-world scenes with complex-shaped objects like trees, chairs, traffic lights and bicycles.

Secondly, the previously-mentioned propensity of the Pix2Seq framework towards misclassification (Sec. \ref{res_summary}) gets exacerbated by the additional classes.
Table \ref{tab:ctscp} demonstrates this through the large discrepancy between image-level metrics
and their class-level counterparts.
The image-level results correspond closely with the training token accuracy of $89\%$ at which this model reached convergence, and this has been true for other training runs too, including on the COCO dataset.
This indicates that, while our models are able to generalize to the validation set fairly well, the loss function is optimizing for global mask correspondence at the expense of the less frequently occuring classes.
This conclusion is also borne out by the official cityscapes metrics \cite{cityscapes_dataset}.
The category score groups the 19 classes into 7 categories (Table \ref{tab:ctscp_cats}), thereby rendering it intermediate between a global and class-level metric.
This is reflected in the $72.2\%$ category score that is much higher than the $53.6\%$ class score but still significant lower than the $79.4 - 87.4\%$ image-level scores.
We are currently looking into class-wise token losses to address this disparity.
The idea is to divide the overall token sequence into a set of class-wise subsequences (e.g. one subsequence for each of the 19 classes in Cityscapes), compute the loss for each subsequence separately and then take the mean of these 19 losses as the overall loss.

\subsection{Hardware and Model Capacity}
\label{future_work_capacity}
One of the primary ways we have tried to improve performance on these datasets is through online RLE computation in tensorflow, as opposed to offline computation in numpy, which was used for all the experiments reported here.
This allows us to use a much wider range of data augmentation techniques, since all augmented images and corresponding RLEs no longer need to be generated before the training starts.
The latter is severely limited by the available storage, especially for larger datasets like COCO where we could only manage $<10$ augmented images for each real image.
Online RLE computation also makes it feasible to deploy additional online tricks like randomizing the order of runs before generating RLE (similar to the order of objects in \cite{p2s}), and more sophisticated forms of class-weight equalization, where token weights are computed dynamically.
However, we have encountered two main issues with this approach.
Firstly, computing RLEs in tensorflow turns out to be extremely computationally expensive and massively slows down training by an order of magnitude, so that we can only do 100K iterations of the online RLE training in the time it takes to do 1 million iterations with the offline variant.
While the better data augmentation does improve performance on a per-iteration basis, the offline variant still ends up having a slight edge overall.
For example, an online RLE model trained on COCO for 100K iterations only manages to match the performance of the corresponding offline model trained for 400K iterations.
Secondly, it seems that applying additional tricks like run-order randomization, and even some more complex forms of image augmentation transforms, makes the RLE sequence simply too difficult to learn for our relatively small ResNet50-based models, so that the token accuracy during training remains stuck at $< 50\%$ and even $< 30\%$ in some cases.
This might be fixable with the larger ViT-based model that Pix2Seq also supports, but our existing hardware resources do not allow us to use it.

\subsection{RLE Compression}
\label{future_work_compression}
In addition to better hardware, another way to improve training is to decrease $L$ and thus memory consumption.
We are experimenting with several encoding schemes to achieve this.
One such scheme is background-as-class (BAC) where we can represent each run with only length and class tokens by considering the background as an additional class.
This makes the runs contiguous so that the start of each run can now be obtained by summing the lengths of all previous runs.
This gives the same benefit of two tokens per run as LAC without the combinatorial increase in $V$, while also allowing us to encode much higher resolution masks without resorting to 2D starts (Sec. \ref{Mask_Flattening}).
In fact, combining BAC and LAC allows each run to be represented by a single token and makes training feasible with mask resolutions up to $S = 256$ while still maintaining $L < 4K$.
Another approach is to compute the RLE from a differential mask generated by taking the difference between consecutive pixels of the original mask.
This reduces each run to unit length and therefore allows us to exclude the length tokens.
The main disadvantage of both of these schemes is a reduction in the robustness to inference noise since the start or length of each run now depends on all previous runs, so that an error in a single run propagates to all subsequent ones. 
Additionally, these too seem to produce more difficult-to-learn RLE sequences, similar to the data-augmented online RLE training.

\subsection{Network Architecture}
\label{future_work_architecture}
An alternative means to reduce the effective $L$ is to change the network architecture itself. 
Fig. \ref{fig:image_decoder_multi_embedding} shows an example where we split the decoder head into multiple parallel heads that each output only one component of each run.
The overall RLE sequence is thus effectively split into $k$ shorter sequences, each of length $L/k$, where $k$ is the number of tokens needed for each run.
Since $L$ has a multiplicative impact on memory consumption, these $k$ heads combined have a significantly smaller memory footprint than a single head that outputs the full sequence.
However, our experiments so far seem to indicate that separating out the RLE components in this way makes the training biased towards easier-to-learn components even when using an appropriately weighted loss.
This massively slows down training for the remaining components and makes it virtually unfeasible.
For example, we found that the setup in Fig. \ref{fig:image_decoder_multi_embedding} resulted in the token accuracy for SX and CLS reaching $80 - 90\%$ while SY and LEN tokens were still at $< 40\%$ after several days of training.
However, we are hopeful of resolving this imbalance with a better loss function.
We are also working to extend this idea to create architectures with class-specific and instance-specific decoder heads to complement the CW and IW schemes.

\subsection{Domain-Specific Tokenization}
\label{future_work_tokenization}
Finally, the tokenization paradigm offers immense scope for designing 
better and
more domain-adaptable models through cleverer and more efficient encoding schemes.
The aim of such schemes can be either to
inject additional information into
the token sequence
or to remove superfluous information from it,
both as a means to improve performance and to provide a better fit for niche applications.
The latter is exemplified by medical imaging whose domain-specific requirements are often ill-suited for general-purpose segmentation, but can be excellent candidates for tokenization.

As mentioned in Sec. \ref{rle}, another such goal is to add object-level information to the RLE sequence while retaining the benefits of the small quantum of RLE and the resultant robustness to inference noise.
Sec. \ref{iw} proposed the IW tokenization as a potential solution that encodes instance information in the sequence by simply reordering the tokens and thus rendering panoptic segmentation feasible.
Early results on static masks are promising, although exhibiting strong signs of batch-size bottleneck, especially when coupled with higher resolution masks that are necessary to extract good quality instance information. 

\section{Conclusions}
\label{conclusions}
This paper introduced a new way to perform semantic segmentation in images and videos by modeling the outputs of these tasks as sequences of discrete tokens.
We have proposed these new methods as another step in the direction of the more general tokenization of visual recognition tasks that has been happening over the last few years through the paradigm of language modeling.
We have tested these models on two datasets with in-depth experiments to demonstrate their competitiveness with the state of the art in conventional modeling.
We have also proposed several promising avenues to improve our models in the future to make them competitive on a wider range of scenarios including larger-scale datasets.

\bmsection*{Financial disclosure}
None reported.

\bmsection*{Conflict of interest}
The authors declare no potential conflict of interests.

\bmsection*{Supporting information}
Additional supporting information may be found in the
online version of the article at the publisher’s website.

\bibliography{wileyNJD-AMA}

\appendix

\bmsection{RLE length Statistics}
\label{app_rle_stats}
This appendix provides statistics about the RLE length ($L$) required to represent masks from the complete IPSC dataset for various combinations of image size $I$, patch size $P$ and mask size $S$ for both static (Tables \ref{tab:rle_len_full_image} - \ref{tab:rle_len_full_and_patch}) and video  (Tables \ref{tab:rle_len_vid_len_2_2560_640} - \ref{tab:rle_len_vid_len_3_to_9_2560_640}) segmentation.
The first two rows in each case show the mean and maximum RLE lengths while the remaining rows show the percentage of images for which the lengths exceed the threshold in the first column.
The smallest threshold that can be used as $L$  in each case is highlighted in yellow.
We have empirically found that a small fraction of lengths ($< 2\%$) exceeding $L$ does not have a significant impact on the segmentation results, so that $L$ does not necessarily need to be $\geq$ maximum length.
Binary and multi-class LAC both use $2$ tokens per run, so ended up having identical lengths for all images over this dataset.
Note that this does not always have to be true in principle, since two runs with different classes that are next to each other in the same row are represented by a single run in the binary case but two different runs in the multi-class case.
However, such scenarios do not occur in the IPSC dataset.
Multi-class masks with separate class tokens use 3 tokens per run so have 1.5 times the lengths of the other two cases.
\begin{table}[t]
	\centering
	\caption{Statistics of static mask RLE lengths over the IPSC dataset with images resized from $I=1280$ to $I=80$, without patches and mask subsampling (so that $S = P = I$).
	}
	\begin{tabular}{c}
		\includegraphics[width=0.45\textwidth]{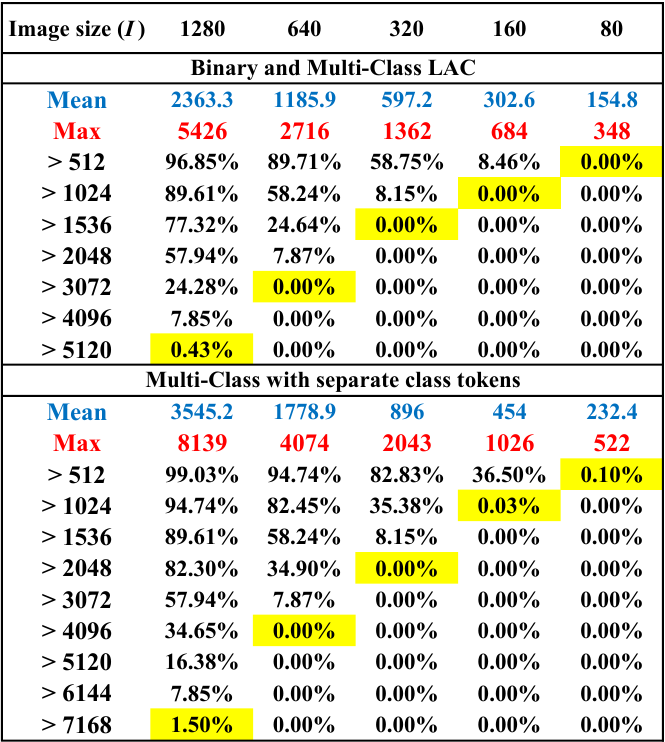}
	\end{tabular}	
	\label{tab:rle_len_full_image}
\end{table}

\begin{table}[t]
	\centering
	\caption{Statistics of static mask RLE lengths over the entire IPSC dataset with (top left) $I=P=640$, (top right) $I=P=1024$, (bottom left) $I=2560, P=640$, and (bottom right) $I=2560, P=1024$.
	}
	\begin{tabular}{c}
		\includegraphics[width=0.47\textwidth]{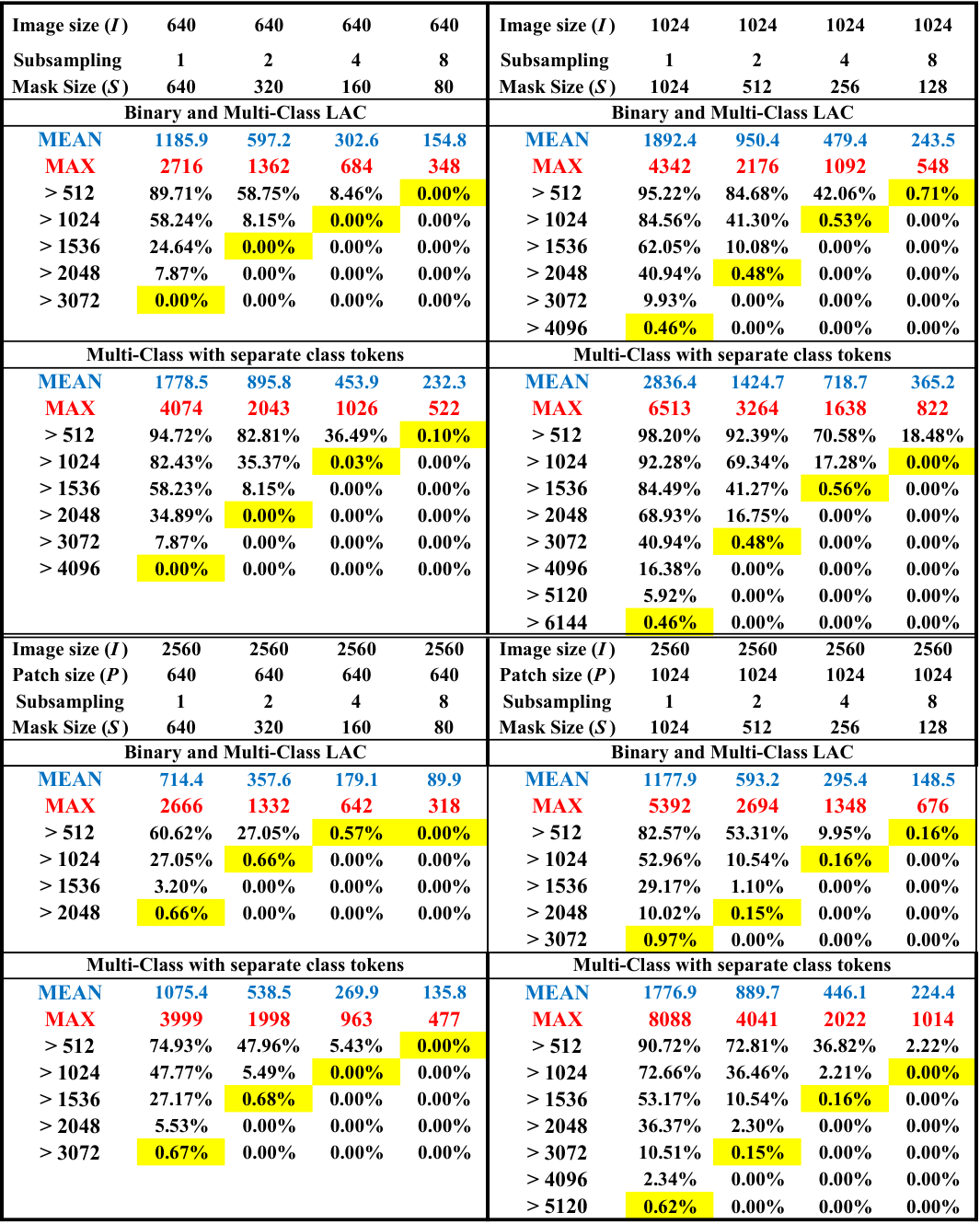}
	\end{tabular}	
	\label{tab:rle_len_full_and_patch}
\end{table}

\begin{table}[t]
	\centering
	\caption{Statistics of video mask RLE lengths over the IPSC dataset with $N=2$, $I=2560$, $P=640$ and $S=80, 160$.
	}
	\begin{tabular}{c}
		\includegraphics[width=0.47\textwidth]{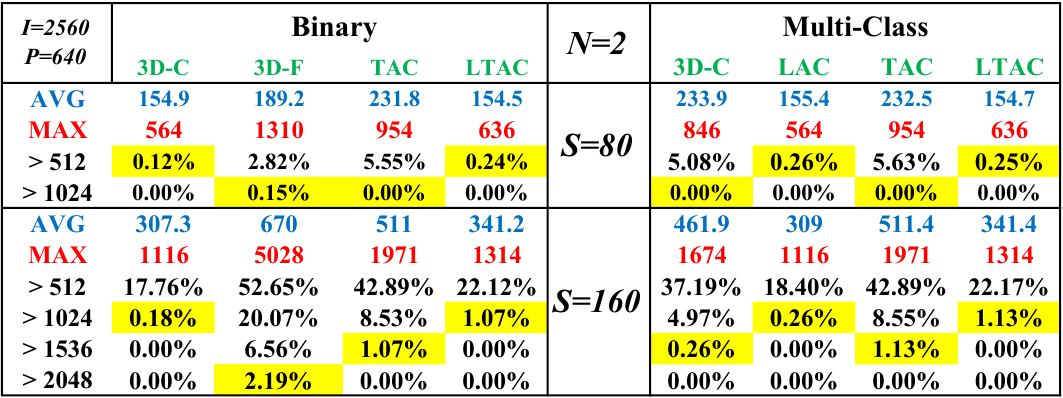}
	\end{tabular}	
	\label{tab:rle_len_vid_len_2_2560_640}
\end{table}

\begin{table}[t]
	\centering
	\caption{Statistics of video mask RLE lengths over the IPSC dataset with $I=2560$, $P=640$, $S=80$ and $N=3, 4, 6, 8, 9$.
	}
	\begin{tabular}{c}
		\includegraphics[width=0.47\textwidth]{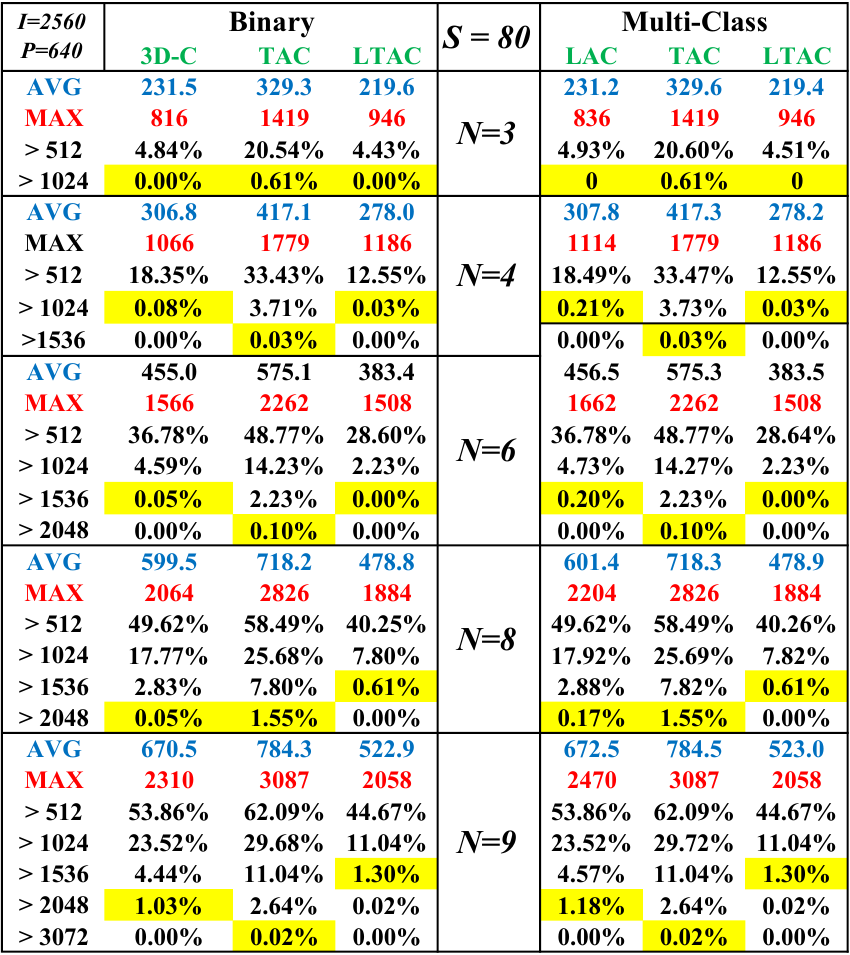}
	\end{tabular}	
	\label{tab:rle_len_vid_len_3_to_9_2560_640}
\end{table}

\bmsection{Vocabulary Sizes}
\label{app_vocab}
This appendix provides tables of vocabulary sizes for $S=40$ (Table \ref{tab:vocab_40}), $S=80$ (Table \ref{tab:vocab_80}), and $S=160$ (Table \ref{tab:vocab_160}) for video segmentation with varying $N$ and all the encoding strategies proposed in Sec. \ref{vid_seg}.
These tables demonstrate the exponential increase in the number of TAC tokens with $N$.
We can also see that, for smaller mask sizes (e.g. $S=40$), straightforward 3D mask flattening remains more practicable for higher values of $N$ due to its linear increase in $V$ with $N$.
Note that these tables only account for limits on $V$ and not those on $L$, which would likely limit $N$ to lower values depending on the complexity of the dataset.
\begin{table}[t]
	\centering
	\caption{Vocabulary sizes ($V$) 
		for video segmentation with $S=40$, varying values of $N$ from $N=2$ to $N=20$ and both binary and multiclass cases.
		The maximum possible $N$ for each case such that $V < 32K$ is highlighted in yellow.
		$V$ is shown in units of thousands.
	}
	\begin{tabular}{c}
		\includegraphics[width=0.47\textwidth]{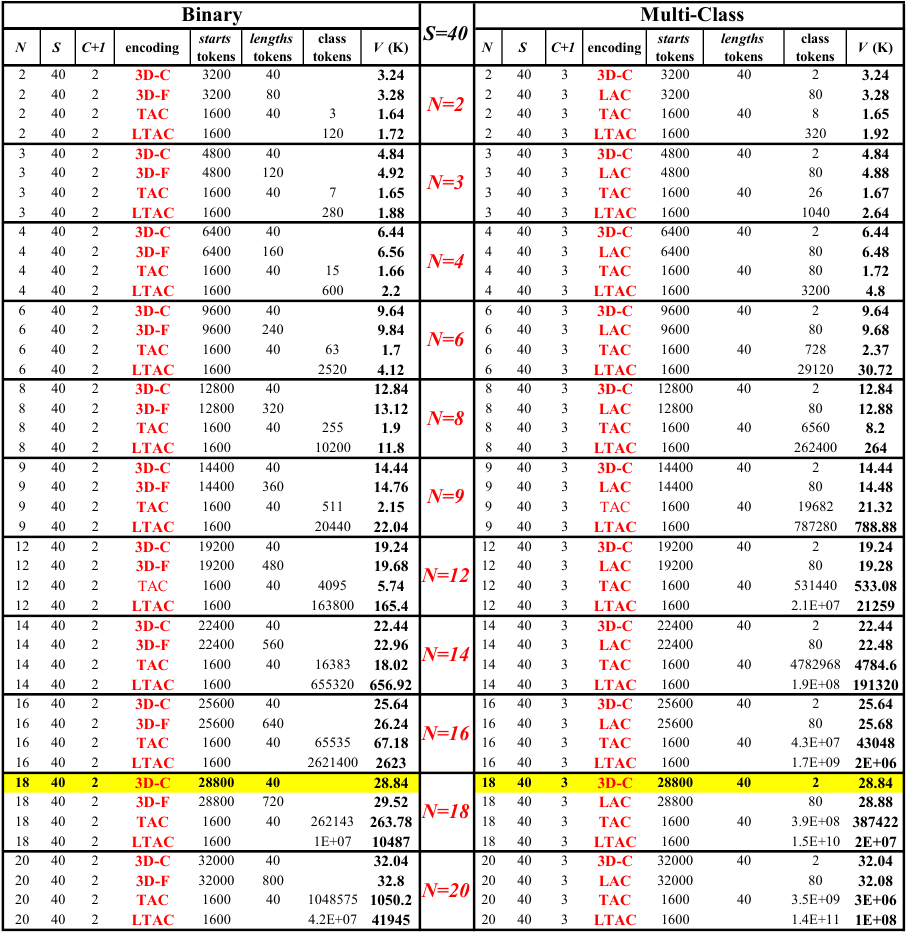}
	\end{tabular}	
	\label{tab:vocab_40}
\end{table}
\begin{table}[t]
	\centering
	\caption{Vocabulary sizes ($V$) 
		for video segmentation with $S=80$, varying values of $N$ from $N=2$ to $N=16$ and both binary and multiclass cases.
		The maximum possible $N$ for each case such that $V < 32K$ is highlighted in yellow.
		$V$ is shown in units of thousands.
	}
	\begin{tabular}{c}
		\includegraphics[width=0.47\textwidth]{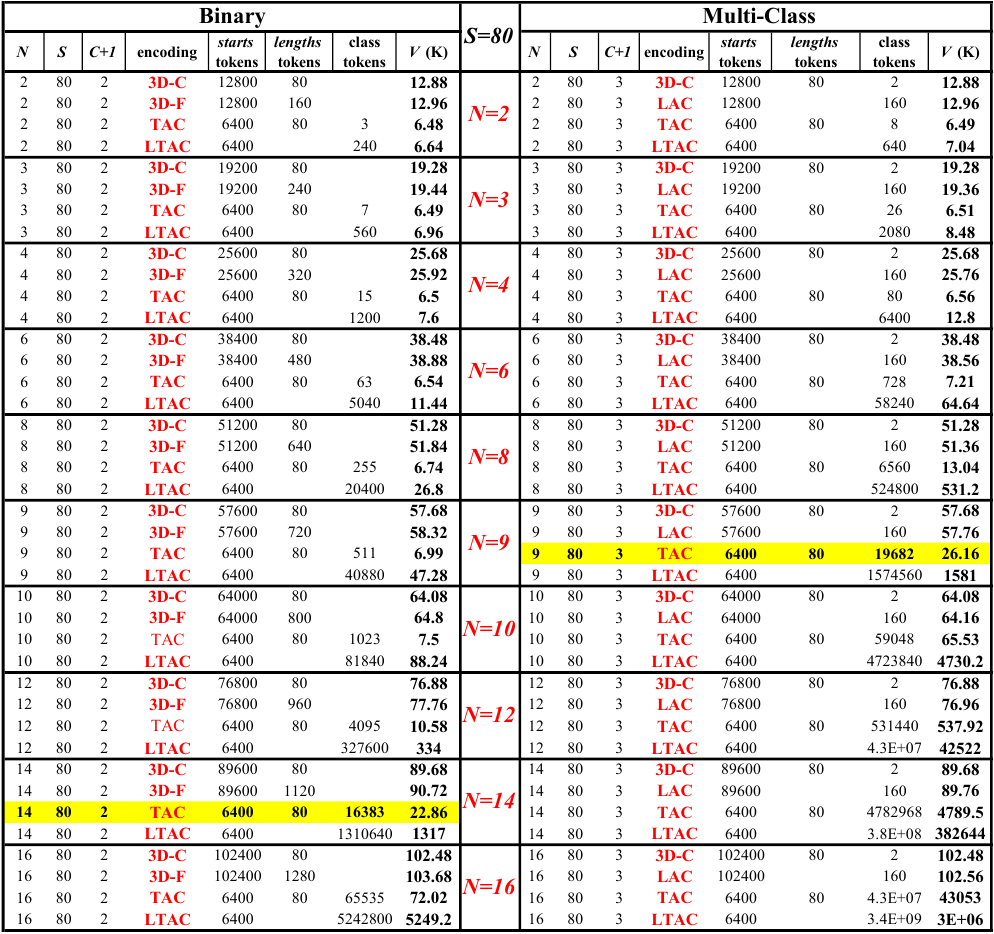}
	\end{tabular}
	
	\label{tab:vocab_80}
\end{table}
\begin{table}[t]
	\centering
	\caption{Vocabulary sizes ($V$) 
		for video segmentation with $S=128$, varying values of $N$ from $N=2$ to $N=16$ and both binary and multiclass cases.
		The maximum possible $N$ for each case such that $V < 32K$ is highlighted in yellow.
		$V$ is shown in units of thousands.
	}
	\begin{tabular}{c}
		\includegraphics[width=0.47\textwidth]{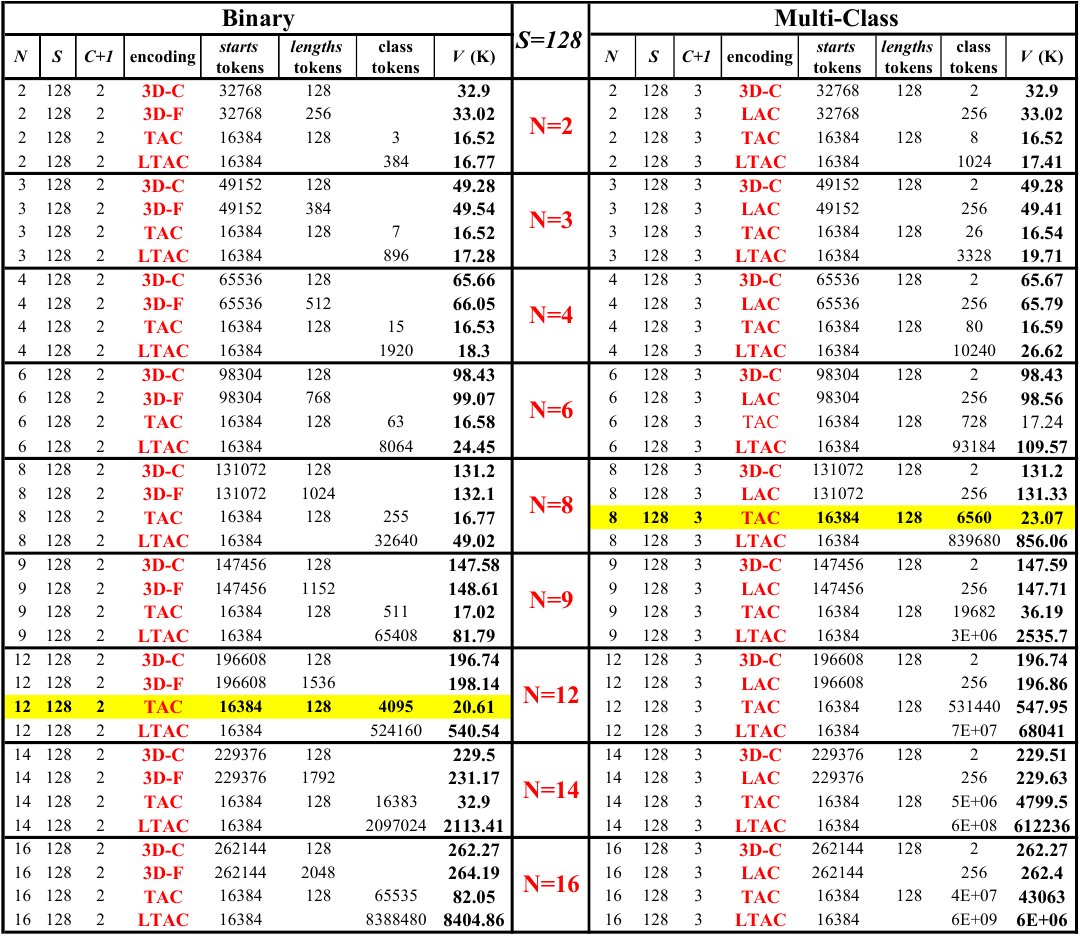}
	\end{tabular}
	
	\label{tab:vocab_128}
\end{table}
\begin{table}[t]
	\centering
	\caption{Vocabulary sizes ($V$) 
		for video segmentation with $S=160$, varying values of $N$ from $N=2$ to $N=16$ and both binary and multiclass cases.
		The maximum possible $N$ for each case such that $V < 32K$ is highlighted in yellow.
		$V$ is shown in units of thousands.
	}
	\begin{tabular}{c}
		\includegraphics[width=0.47\textwidth]{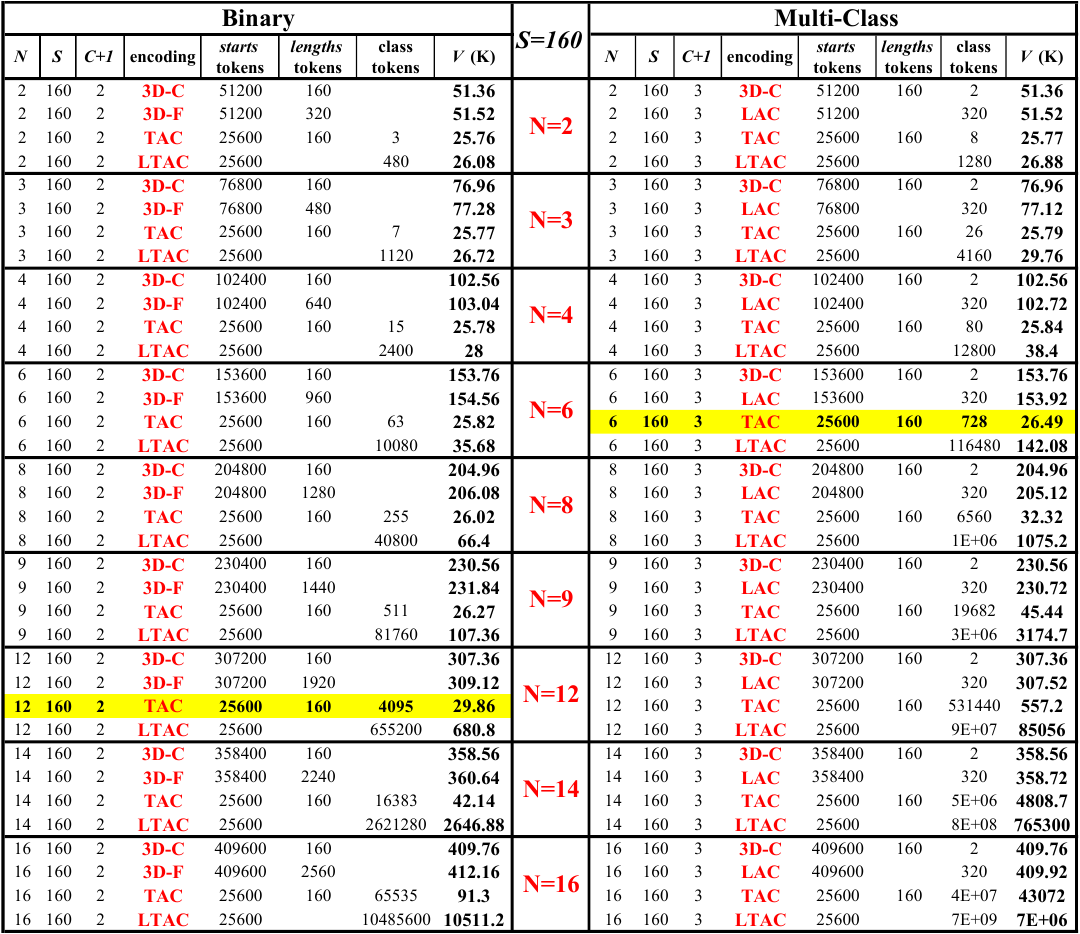}
	\end{tabular}
	
	\label{tab:vocab_160}
\end{table}

\bmsection{Segmentation Metrics for Subsampled Masks}
\label{app_segm_metrics}
This appendix provides statistics about the degradation in mask quality produced by subsampling (Table \ref{tab:segm_metrics_cmb}).
This is quantitatively represented by the three main segmentation metrics used in this paper – dice score, recall and precision – averaged over the entire ARIS and IPSC datasets.
Qualitatively, most subsampled masks are visually almost indistinguishable from the original when these metrics are $ > 90\%$ and sometimes even when they are $ > 80\%$.

\begin{figure*}[!t]
	\centering
	\includegraphics[width=0.49\textwidth]{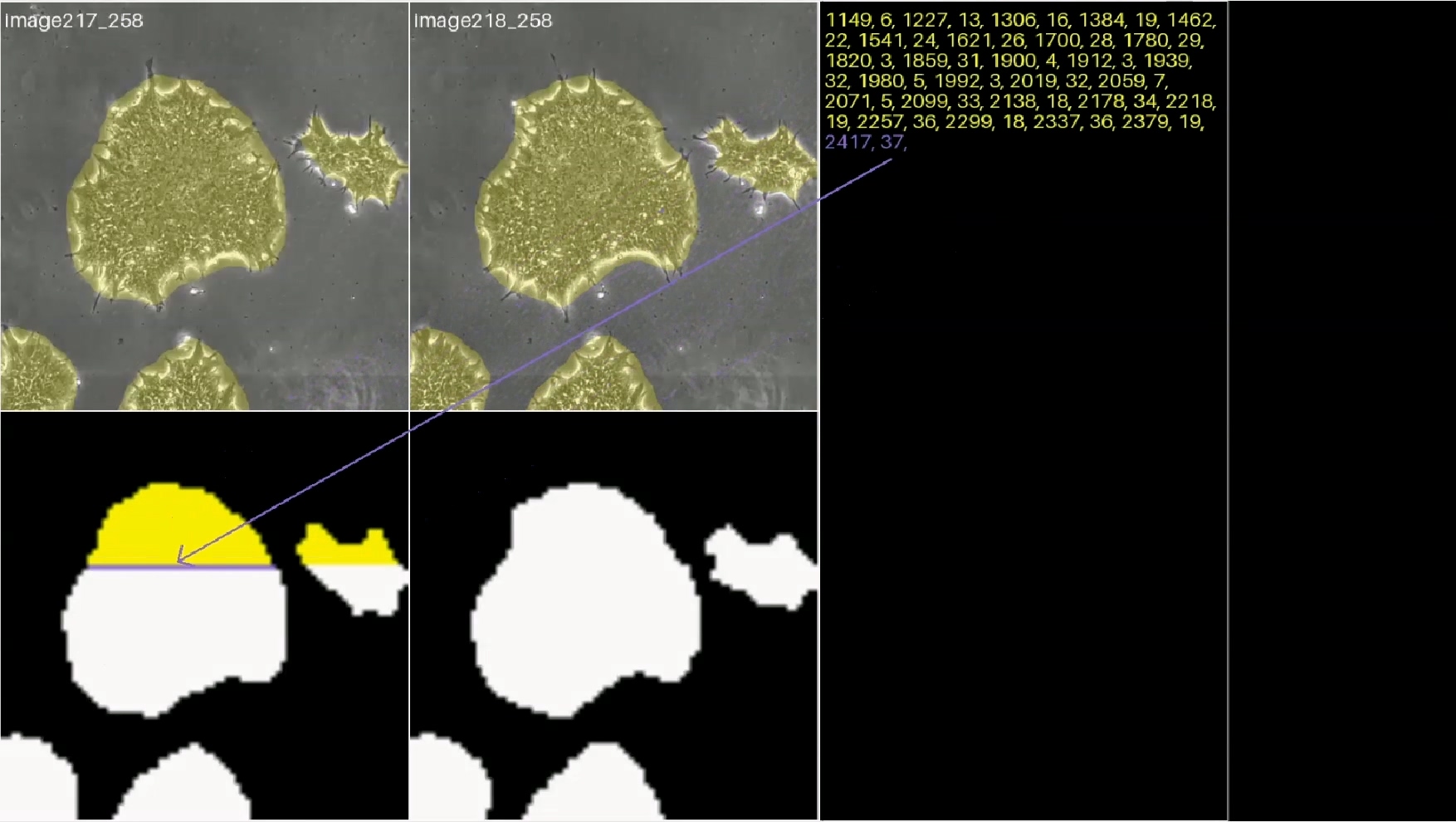}
	\includegraphics[width=0.49\textwidth]{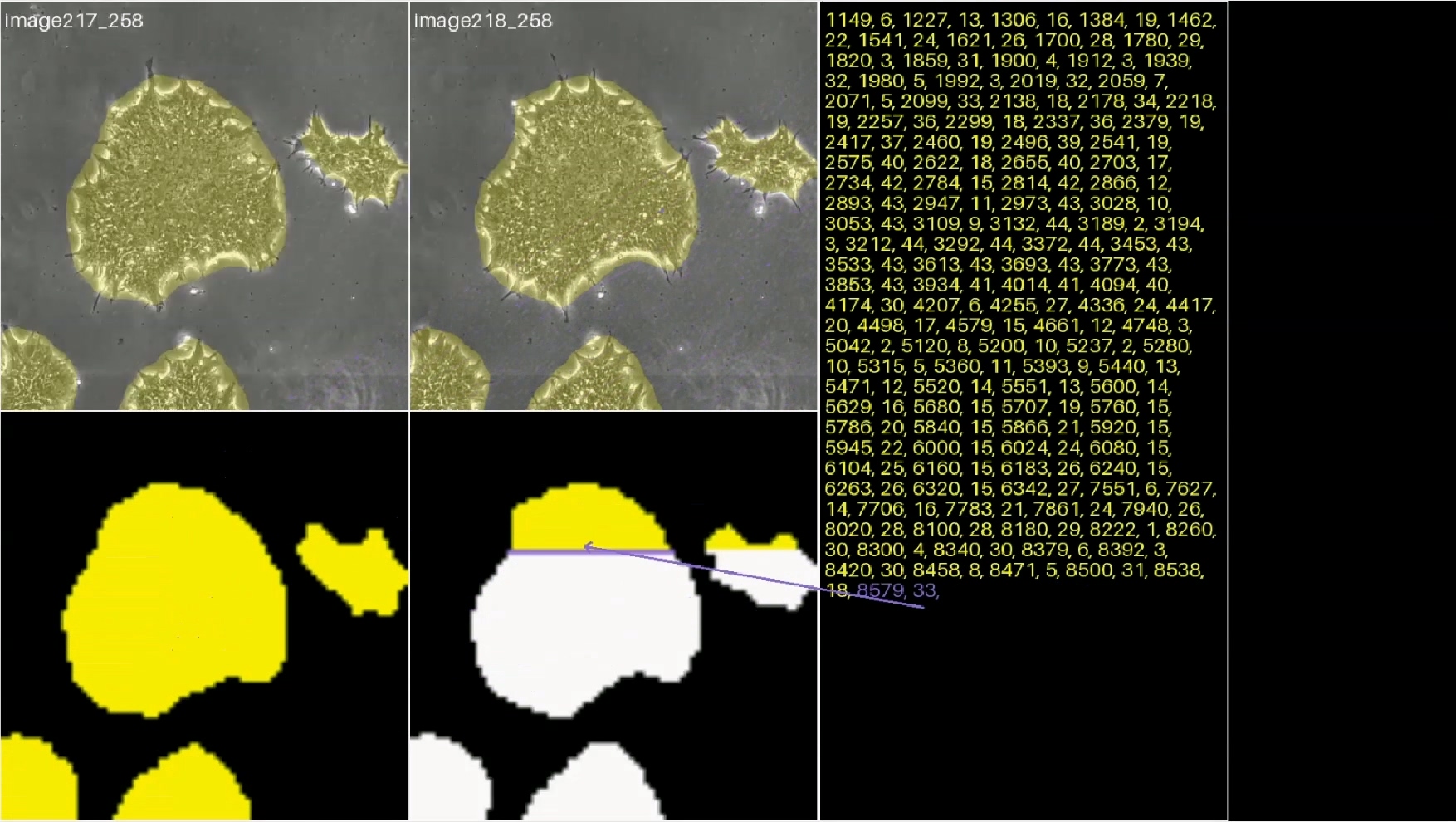}
	\caption{
		Visualization of 3D-C RLE tokenization of binary video segmentation masks with $N=2$, showing runs corresponding to the same part of the same object in $F_1$ (left) and $F_2$ (right).		
		We can see that their tokens are completely unrelated in the sequence.
		Animated version of this figure is available
		\href{https://webdocs.cs.ualberta.ca/~asingh1/p2s\#vid_seg_binary_row_major}{here}.
		Best viewed under high magnification.
	}
	\label{fig:vid_seg_binary_row_major}	
\end{figure*}

\begin{figure*}[t]
	\centering
	\includegraphics[width=0.49\textwidth]{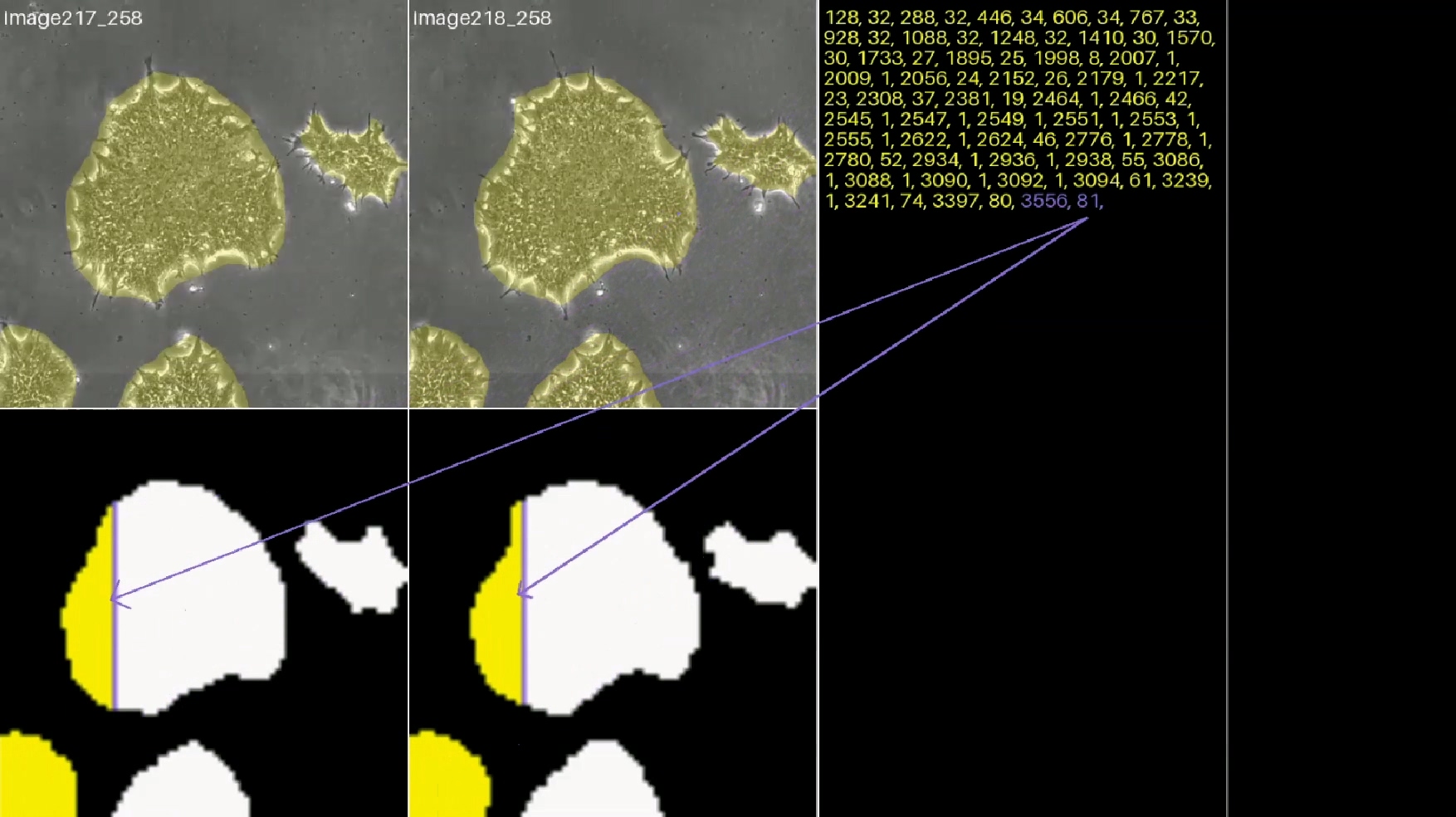}
	\includegraphics[width=0.49\textwidth]{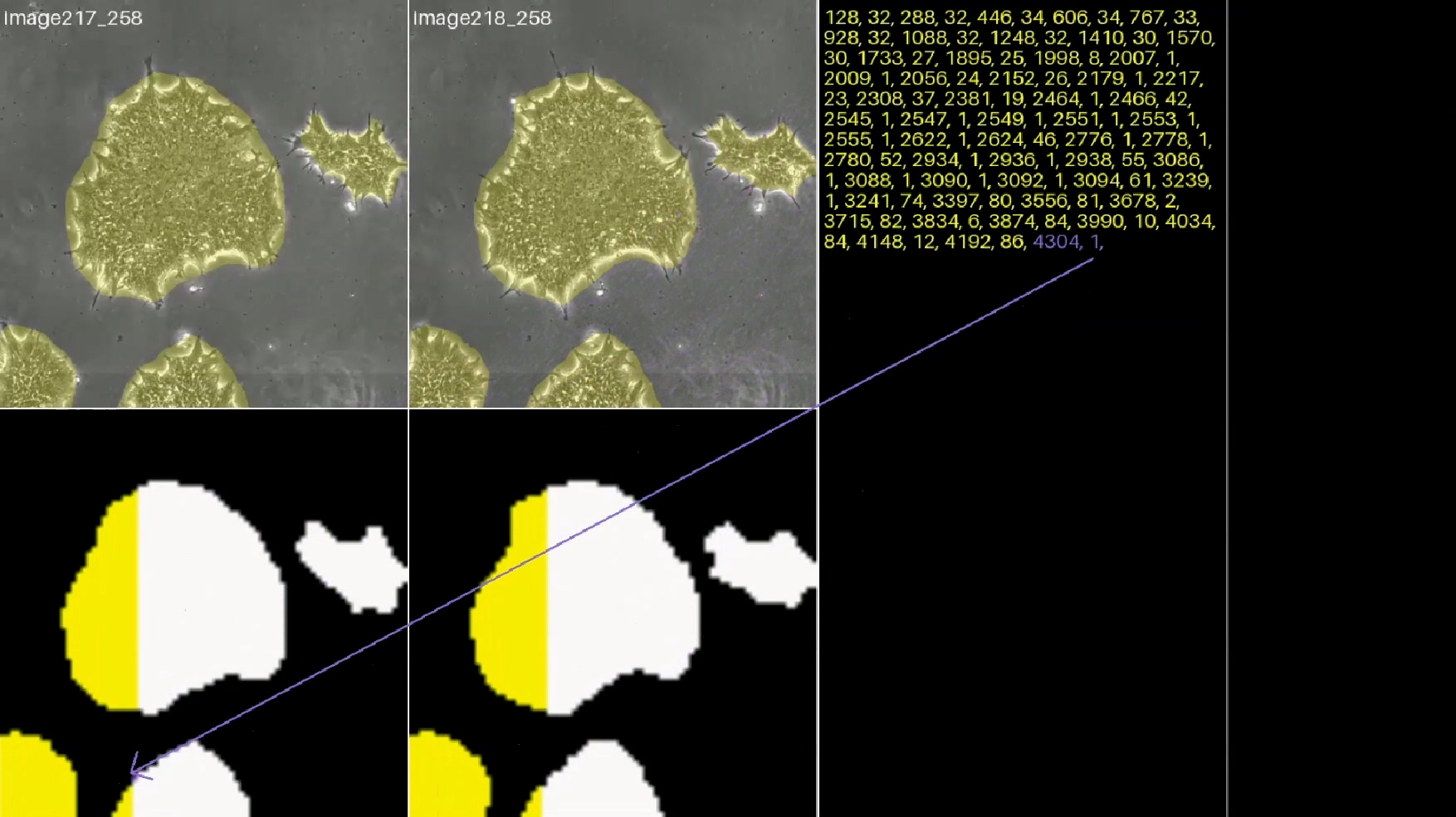}
	\caption{
		Visualization of 3D-F RLE tokenization of binary video segmentation masks with $N=2$.
		The left half shows the same part of the same object in $F_1$ and $F_2$ represented by the same run and the right half shows one of the unit sized runs created by a tiny change in the shape of the cell between the two frames.
		Animated version of this figure is available
		\href{https://webdocs.cs.ualberta.ca/~asingh1/p2s\#vid_seg_binary_column_major}{here}.
				Best viewed under high magnification.
	}
	\label{fig:vid_seg_binary_column_major}	
\end{figure*}

\begin{figure}[t]
	\centering
	\includegraphics[width=0.47\textwidth]{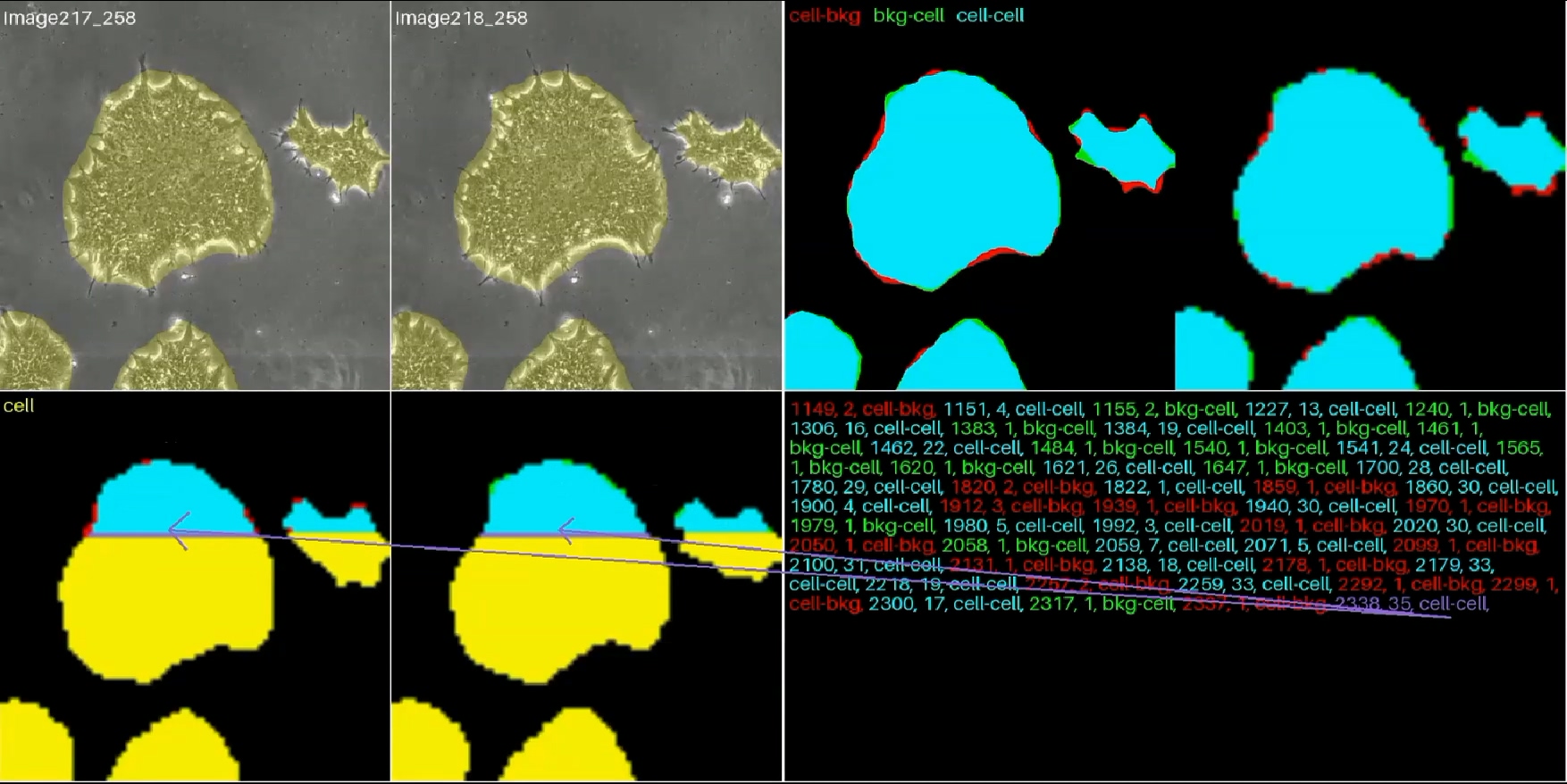}
	\caption{
		Visualization of TAC tokenization for binary segmentation with $N=2$.
		Animated version of this figure is available				\href{https://webdocs.cs.ualberta.ca/~asingh1/p2s\#vid_seg_binary_tac}{here}.
		Best viewed under high magnification.
	}
	\label{fig:vid_seg_binary_tac}	
\end{figure}

\begin{figure}[t]
	\centering
	\includegraphics[width=0.49\textwidth]{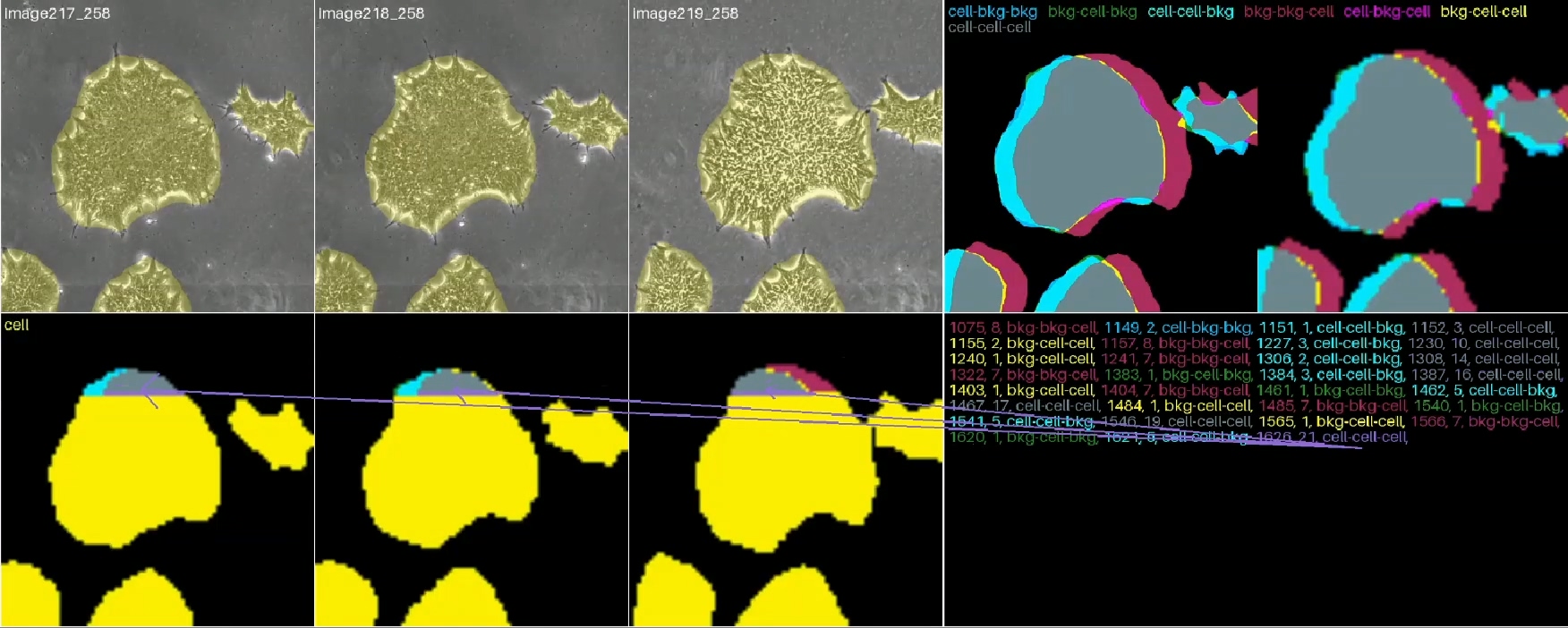}
	\includegraphics[width=0.49\textwidth]{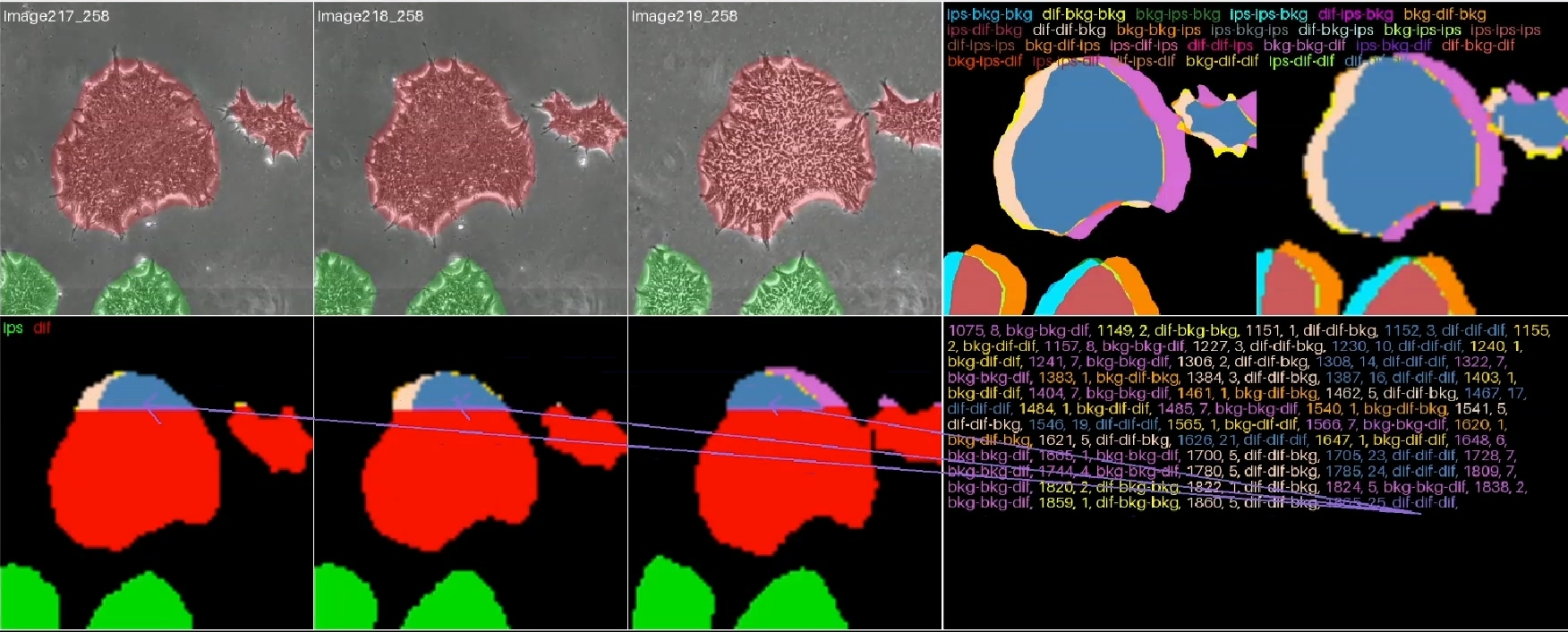}
	\caption{
		Visualization of TAC tokenization with $N=3$ for (top) binary and (bottom) multi-class masks.
		Animated versions of these figures are available
		\href{https://webdocs.cs.ualberta.ca/~asingh1/p2s\#vid_seg_binary_tac_len_3}{here}
		and
		\href{https://webdocs.cs.ualberta.ca/~asingh1/p2s\#vid_seg_multi_class_tac_len_3}{here}
		respectively.
		Best viewed under high magnification.
	}
	\label{fig:vid_seg_tac_len_3}	
\end{figure}

\begin{figure}[t]
	\centering
	\includegraphics[width=0.49\textwidth]{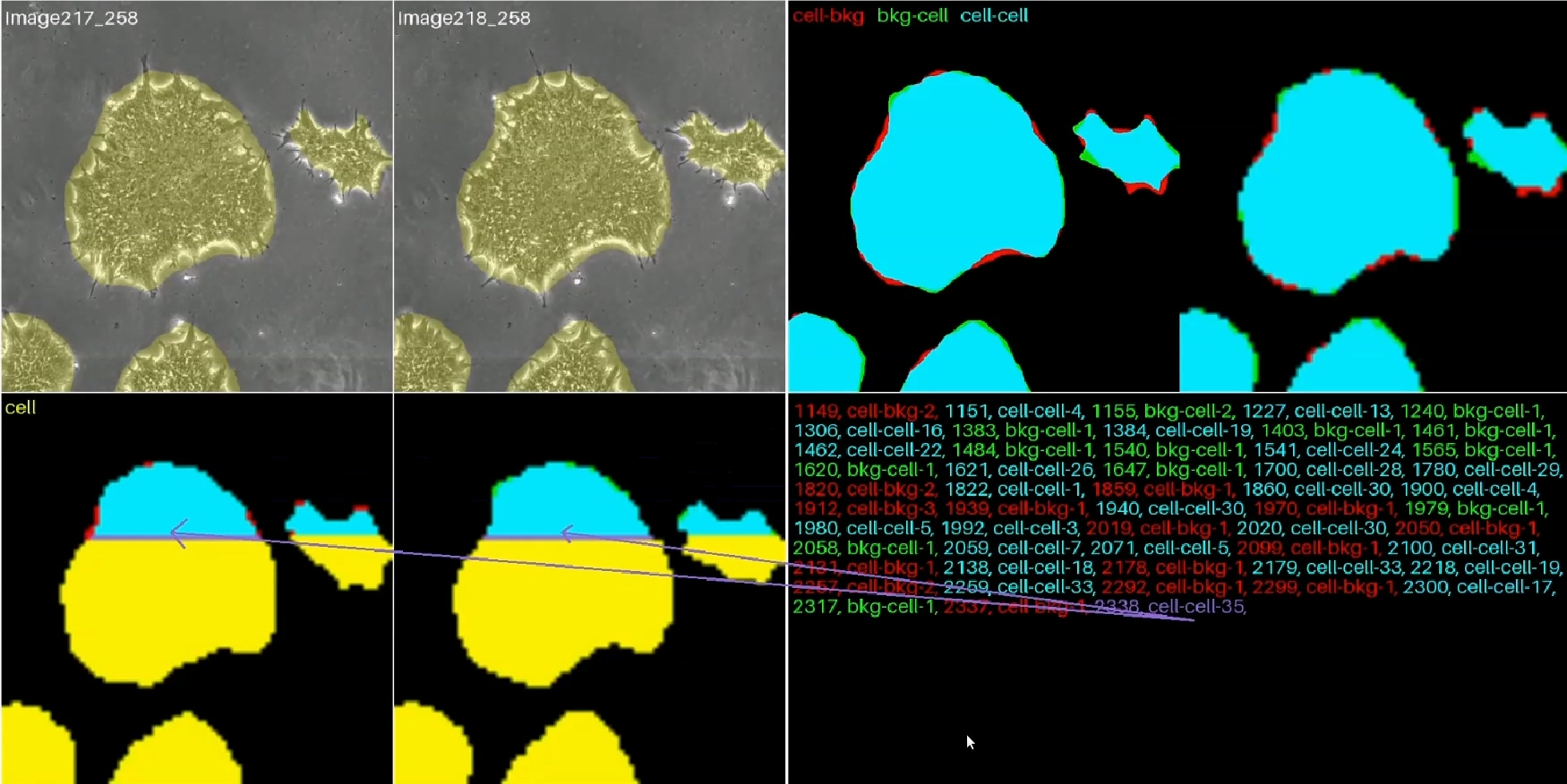}
	\includegraphics[width=0.49\textwidth]{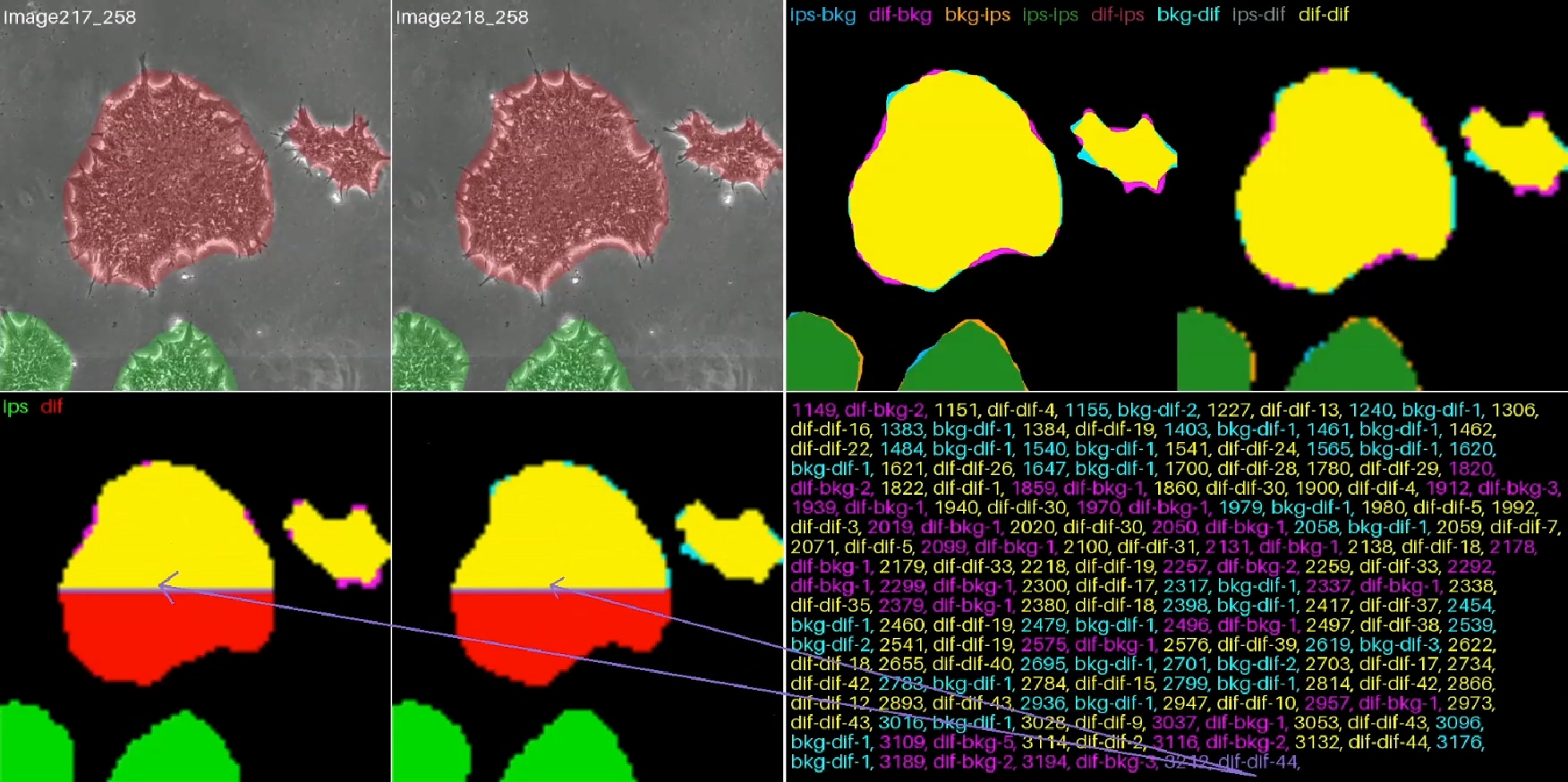}
	\caption{
		Visualization of LTAC tokenization for video segmentation masks with $N=2$ for (top) binary and (bottom) multi-class cases.
		Animated versions of these figures are available 	\href{https://webdocs.cs.ualberta.ca/~asingh1/p2s\#vid_seg_binary_ltac}{here} and 	\href{https://webdocs.cs.ualberta.ca/~asingh1/p2s\#vid_seg_multi_class_ltac}{here} respectively.
		Best viewed under high magnification.
	}
	\label{fig:vid_seg_ltac}	
\end{figure}

\bmsection{Visualization}
\label{app_vis}
This appendix presents visualization images for the various RLE tokenization schemes for video segmentation.
Figures \ref{fig:vid_seg_binary_row_major} and \ref{fig:vid_seg_binary_column_major} respectively show 3D-C and 3D-F tokenization schemes for binary masks.
Figures \ref{fig:vid_seg_binary_tac} and \ref{fig:vid_seg_tac_len_3} show TAC tokenization for $N=2$ and $N=3$ respectively.
Fig. \ref{fig:vid_seg_ltac} shows LTAC tokenization for $N=2$.
Please refer Fig. \ref{fig:vid_seg_multi_class_tac} for details of the layout Figures \ref{fig:vid_seg_binary_tac} - \ref{fig:vid_seg_ltac}.
Each figure has an animated version on the project website \cite{p2sv_web} whose link is provided in its caption.

\begin{table}[t]
	\centering
	\caption{
		Segmentation metrics representing mask quality degradation by subsampling, averaged over the entire ARIS and IPSC datasets.
	}
	\begin{tabular}{c}
		\includegraphics[width=0.47\textwidth]{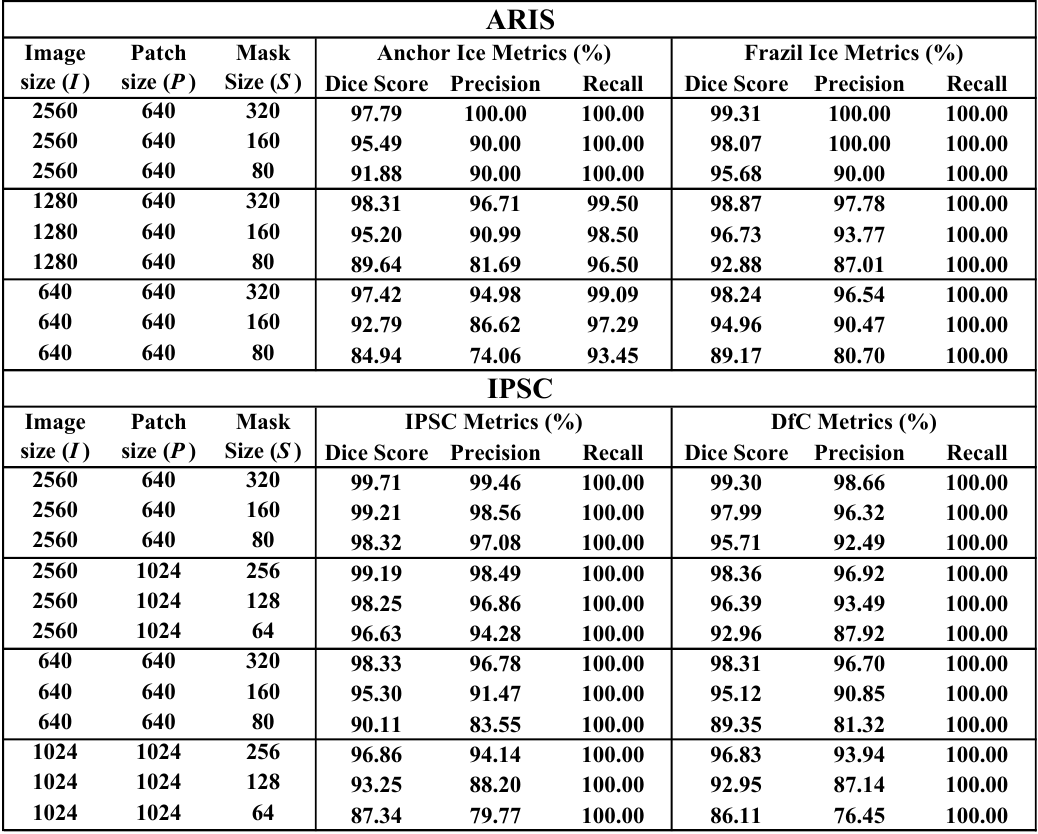}
	\end{tabular}
	
	\label{tab:segm_metrics_cmb}
\end{table}

\begin{figure}[t]
	\centering
	\includegraphics[width=0.49\textwidth]{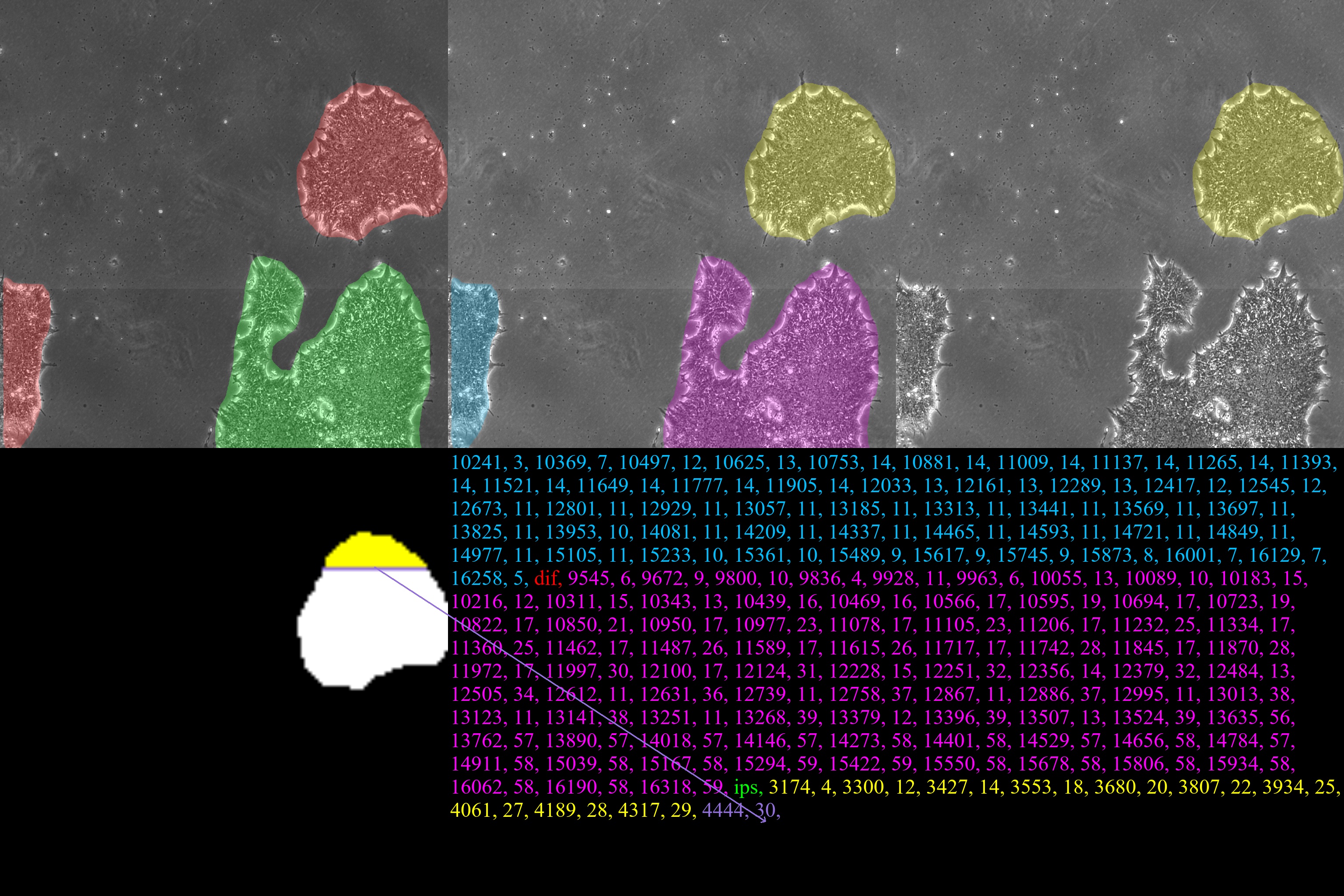}
	\caption{
		Visualization of IW tokenization for multi-class static segmentation mask.
		The first image in the top row shows the class mask where \textit{IPSC} and \textit{DfC} cells are shown in green and red respectively.
		The second image shows the instance mask where each individual cell is shown in a different color.
		The last image shows the cell whose binary RLE tokens are currently being generated. The corresponding subsampled binary mask is shown in the first image in the second row.
		The tokens for the other two cells have already been generated (shown in corresponding colors) and terminated by the respective class tokens.		 
		An animated version of this figure is available 	\href{https://webdocs.cs.ualberta.ca/~asingh1/p2s\#seg_iw}{here}.
		Best viewed under high magnification.
	}
	\label{fig:seg_iw}	
\end{figure}

\section{Results}
\label{app_results}
\begin{figure}[t]
	\centering
	\includegraphics[width=0.47\textwidth]{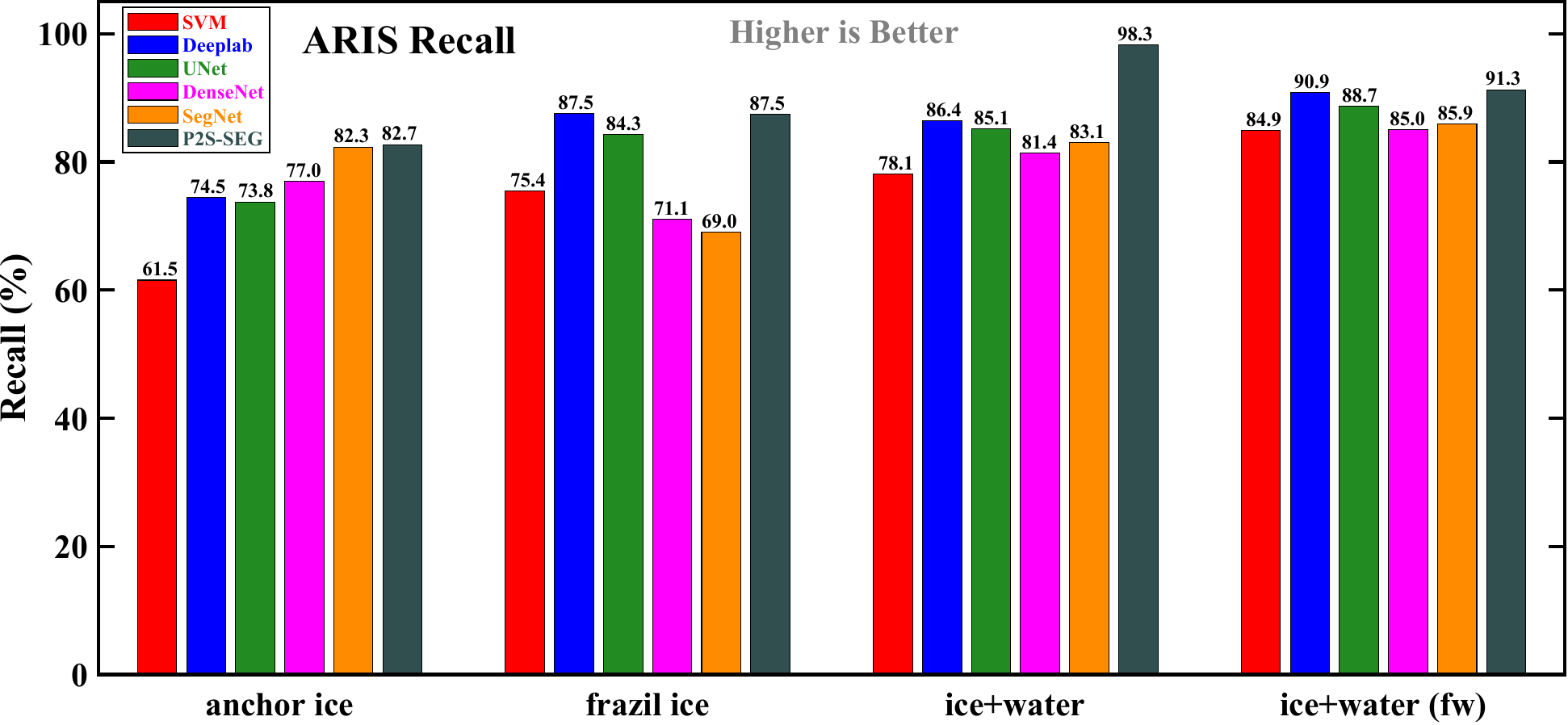}
	\includegraphics[width=0.47\textwidth]{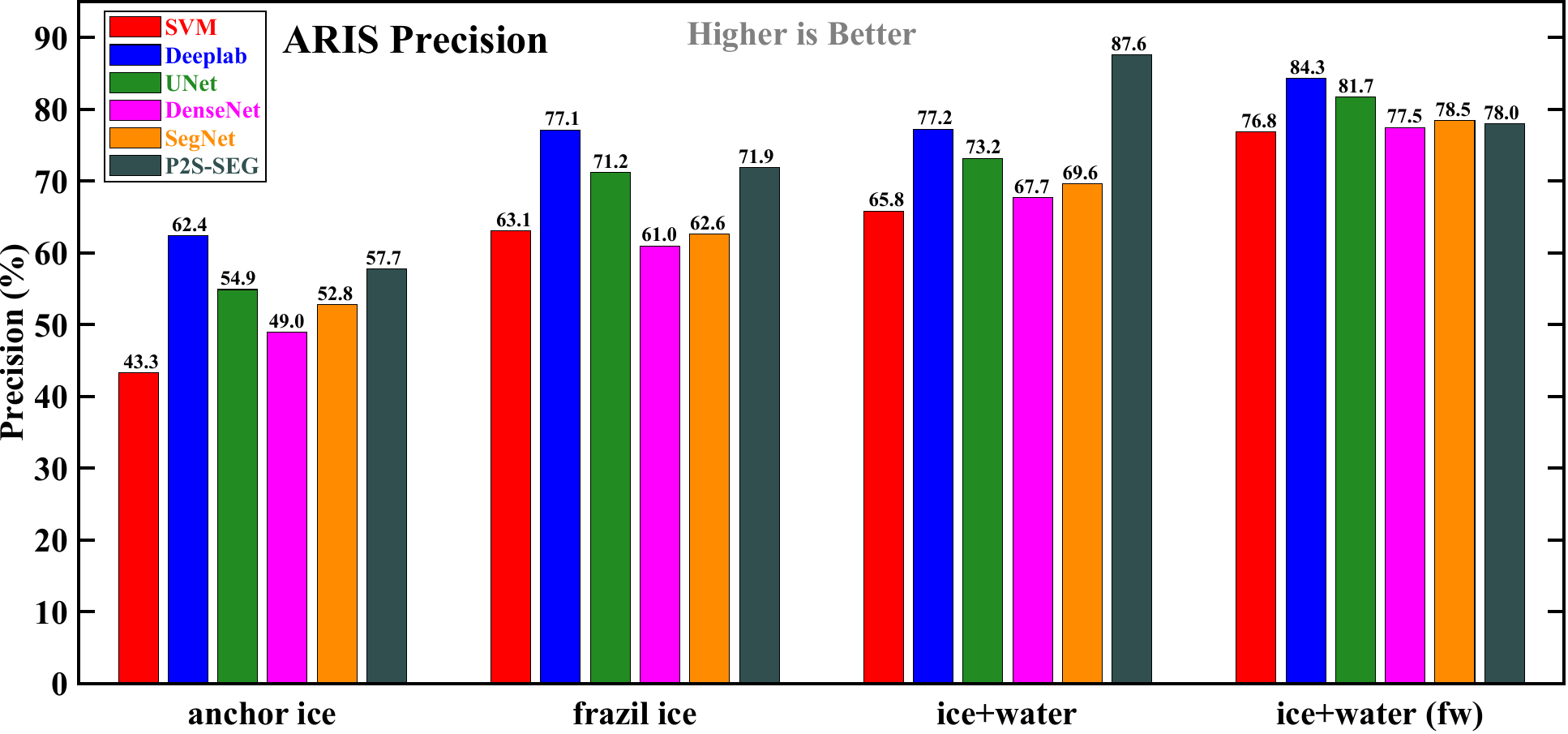}
	\includegraphics[width=0.47\textwidth]{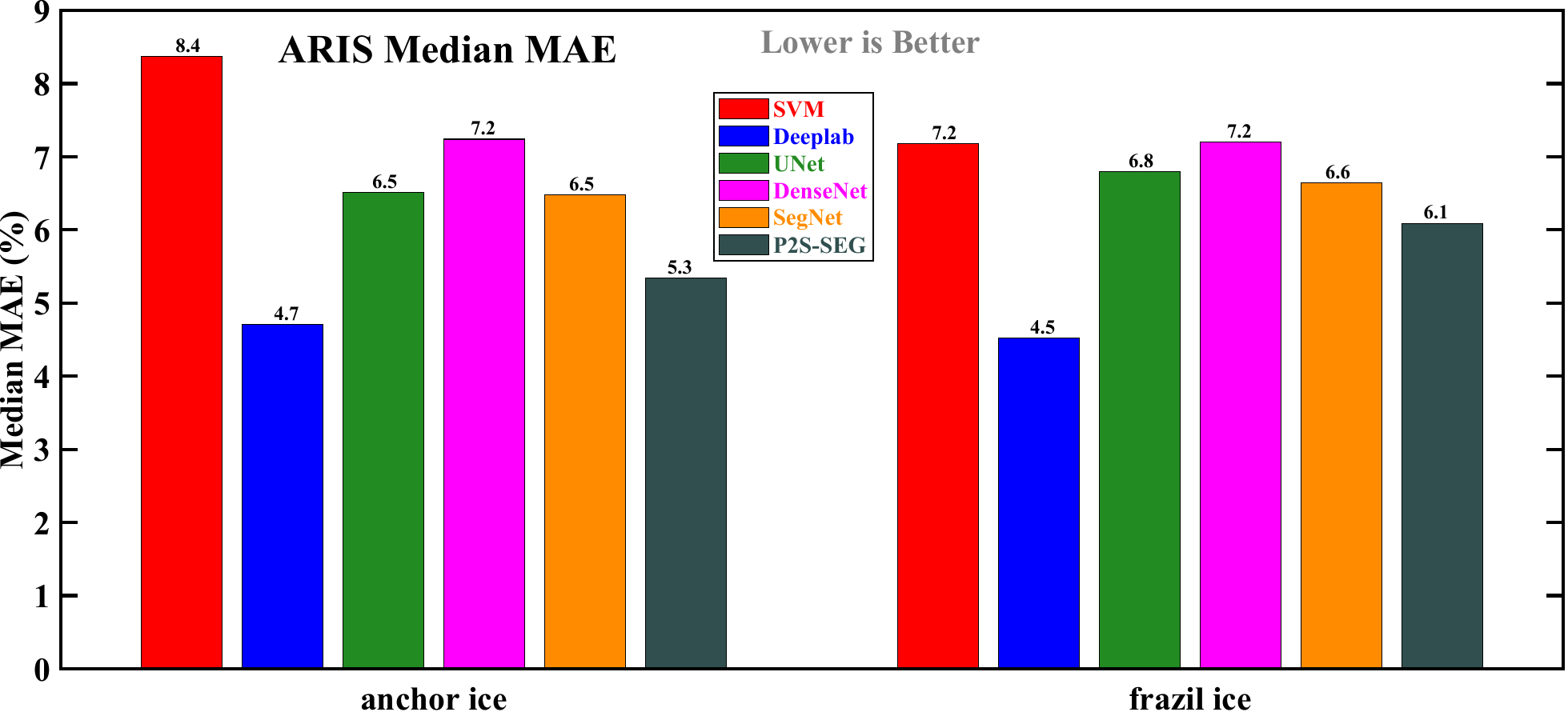}
	\caption{
		Bar plot version of the data in Table \ref{tab:617_summary_single_p2s}.
		Best viewed under high magnification.		
	}
	\label{617_summary_bar}
\end{figure}

\begin{figure*}[t]
	\centering
	\includegraphics[width=0.32\textwidth]{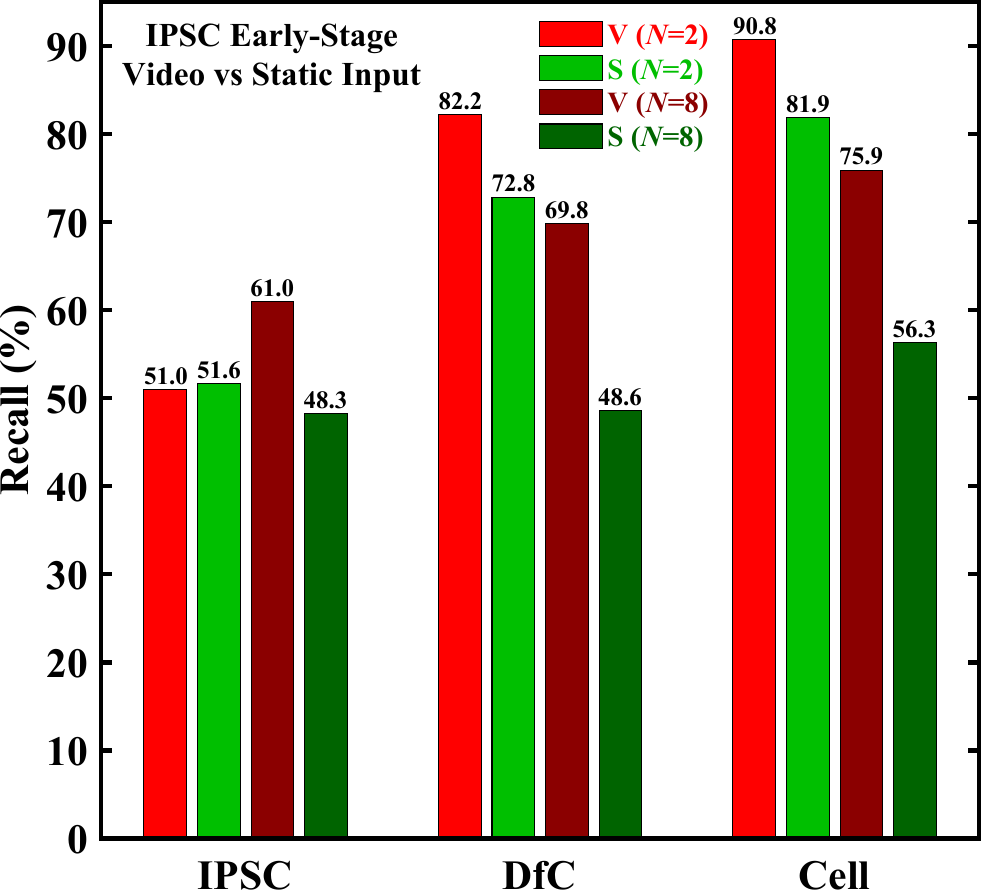}
	\includegraphics[width=0.32\textwidth]{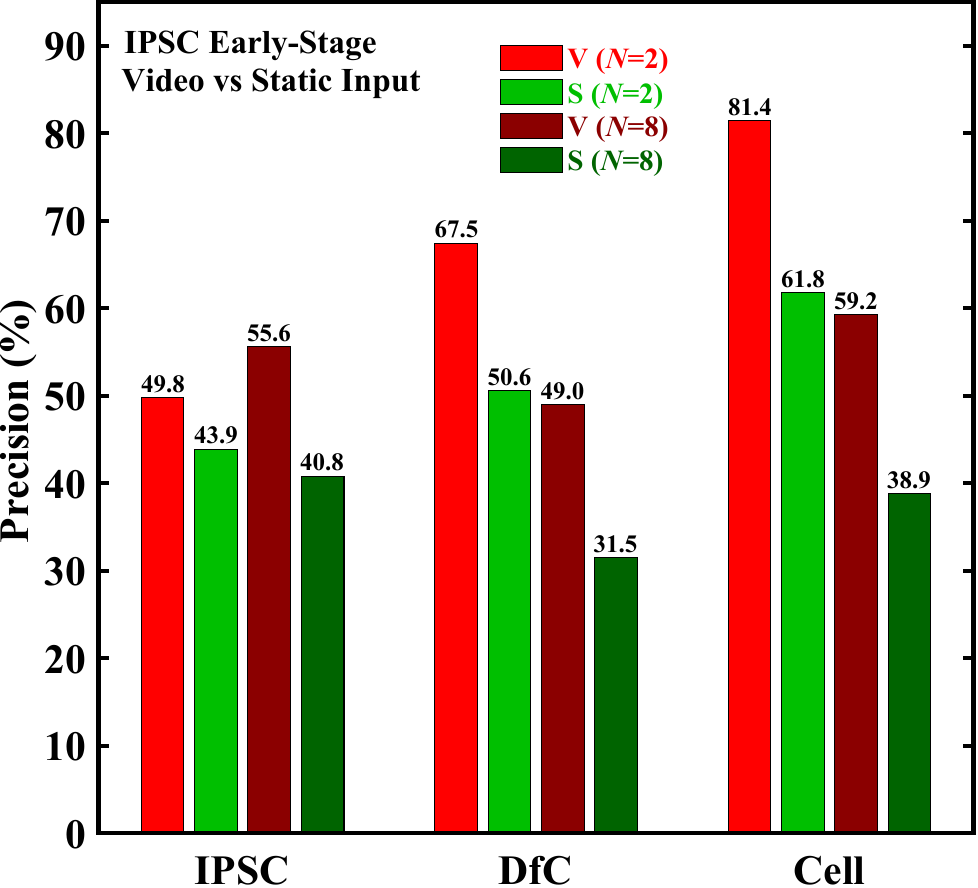}
	\includegraphics[width=0.32\textwidth]{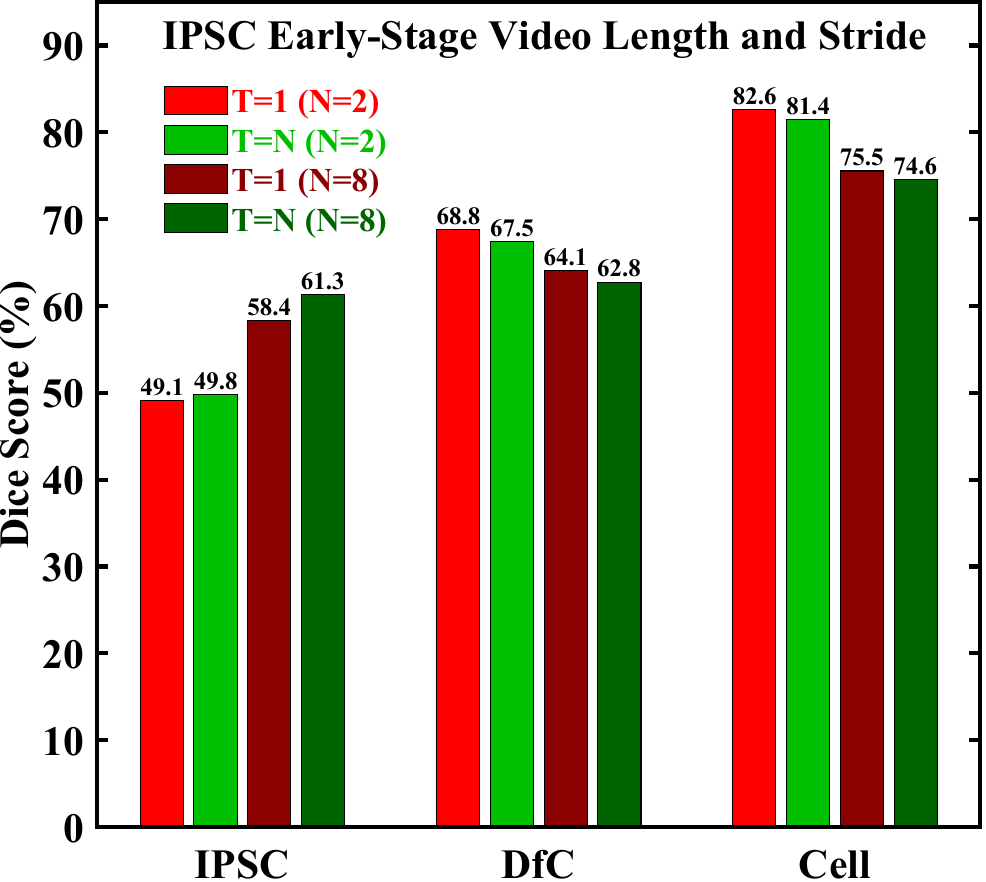}
	\includegraphics[width=0.32\textwidth]{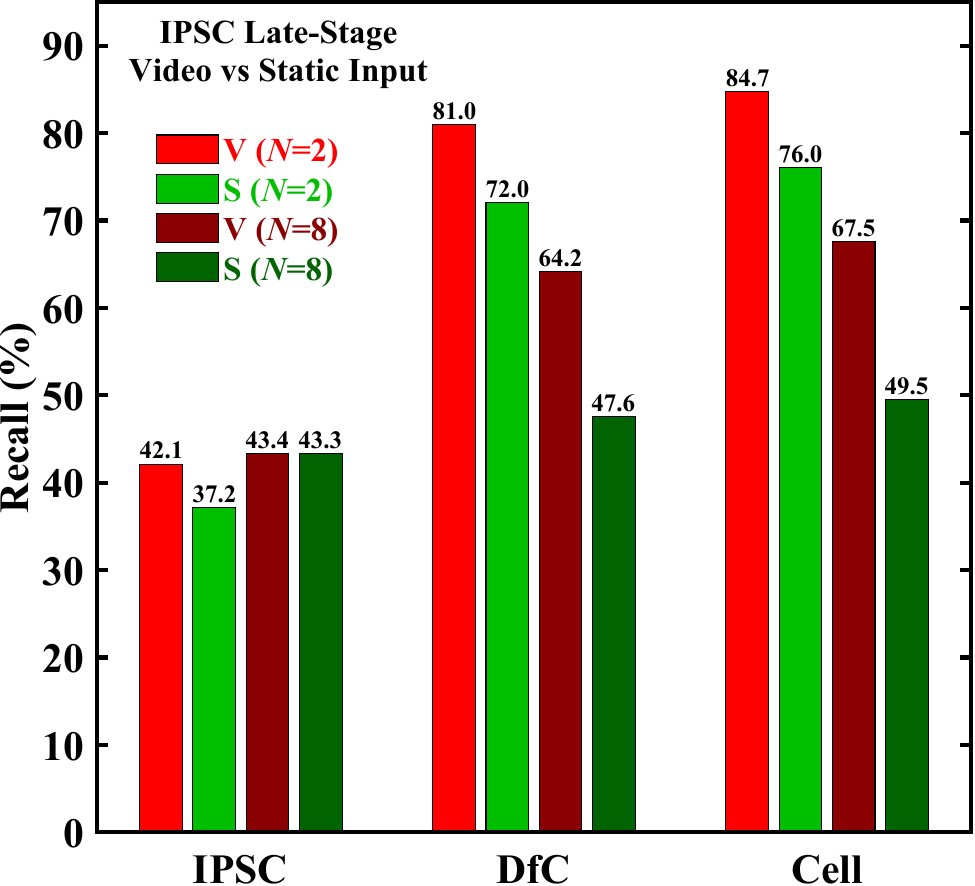}		
	\includegraphics[width=0.32\textwidth]{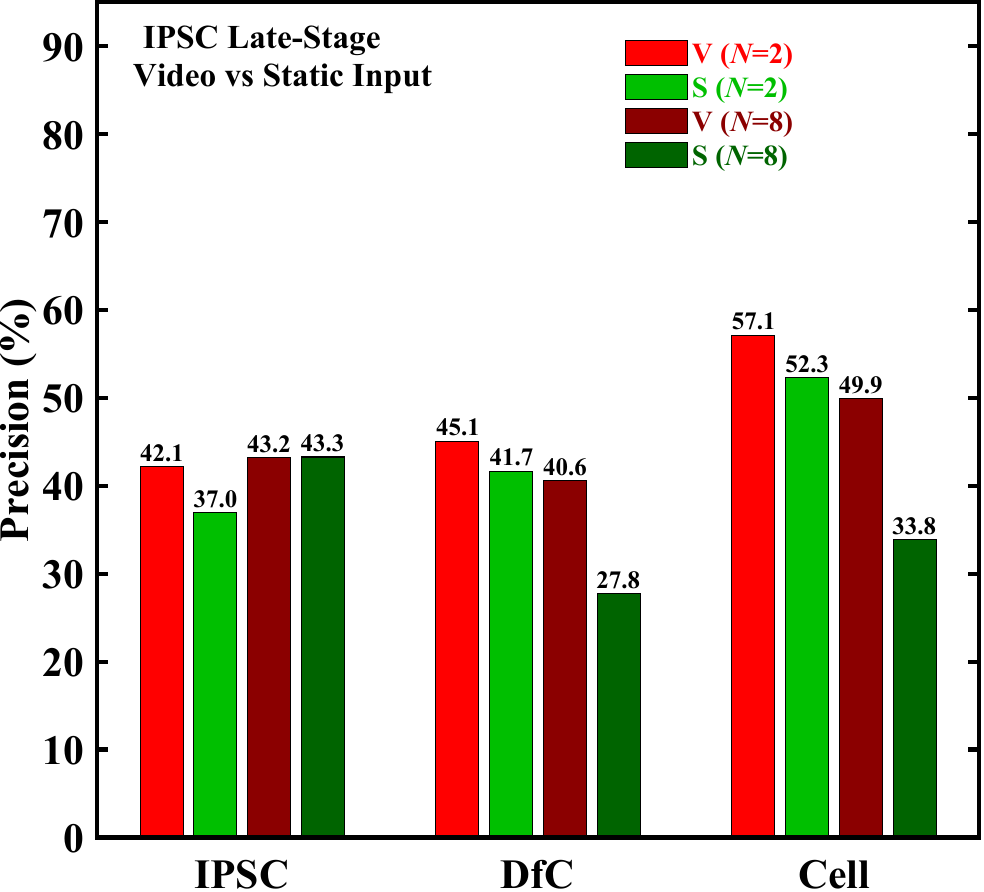}
	\includegraphics[width=0.32\textwidth]{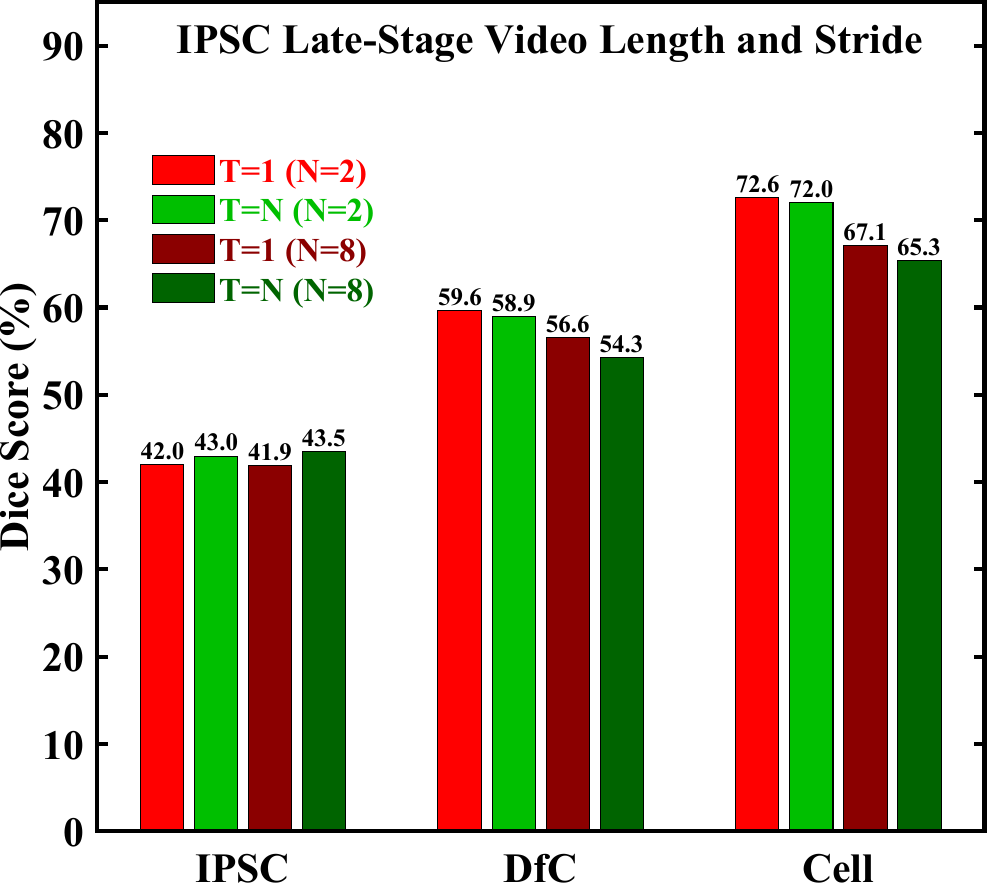}
	\caption{
		Impact of (left and center) replacing $N$ video frames with the first frame and (right) video length $N$ and stride $T$ on P2S-VIDSEG performance over (top) early and (bottom) late-stage configurations of the IPSC dataset.
	}
	\label{app:static_vid_len_seg}
\end{figure*}

This appendix provides some supplementary data (Fig.  \ref{617_summary_bar}, \ref{app:static_vid_len_seg} and Tables \ref{tab:seg_details_ipsc_early_static}, \ref{tab:seg_details_ipsc_late_vid}) and configuration details (Table \ref{tab:model_configs}) that have been used in Sec. \ref{results}.

\begin{table*}[t]
	\centering
	\caption{
		Configuration and training details for models whose results are reported in Sec. \ref{res_overview}
	}
	\begin{tabular}{c}
		\includegraphics[width=0.94\textwidth]{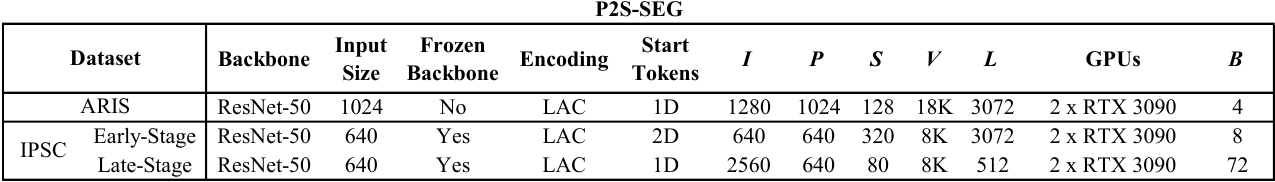}\\
		\includegraphics[width=0.94\textwidth]{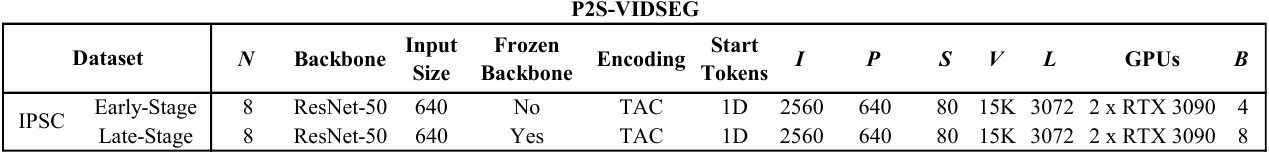}
	\end{tabular}	
	\label{tab:model_configs}
\end{table*}

\begin{table}[t]
	\centering
	\caption{
		Segmentation metrics ($\%$) on validation set while training P2S-SEG on IPSC early-stage dataset.
		Best viewed under high magnification.				
	}
	\begin{tabular}{c}
		\includegraphics[width=0.4\textwidth]{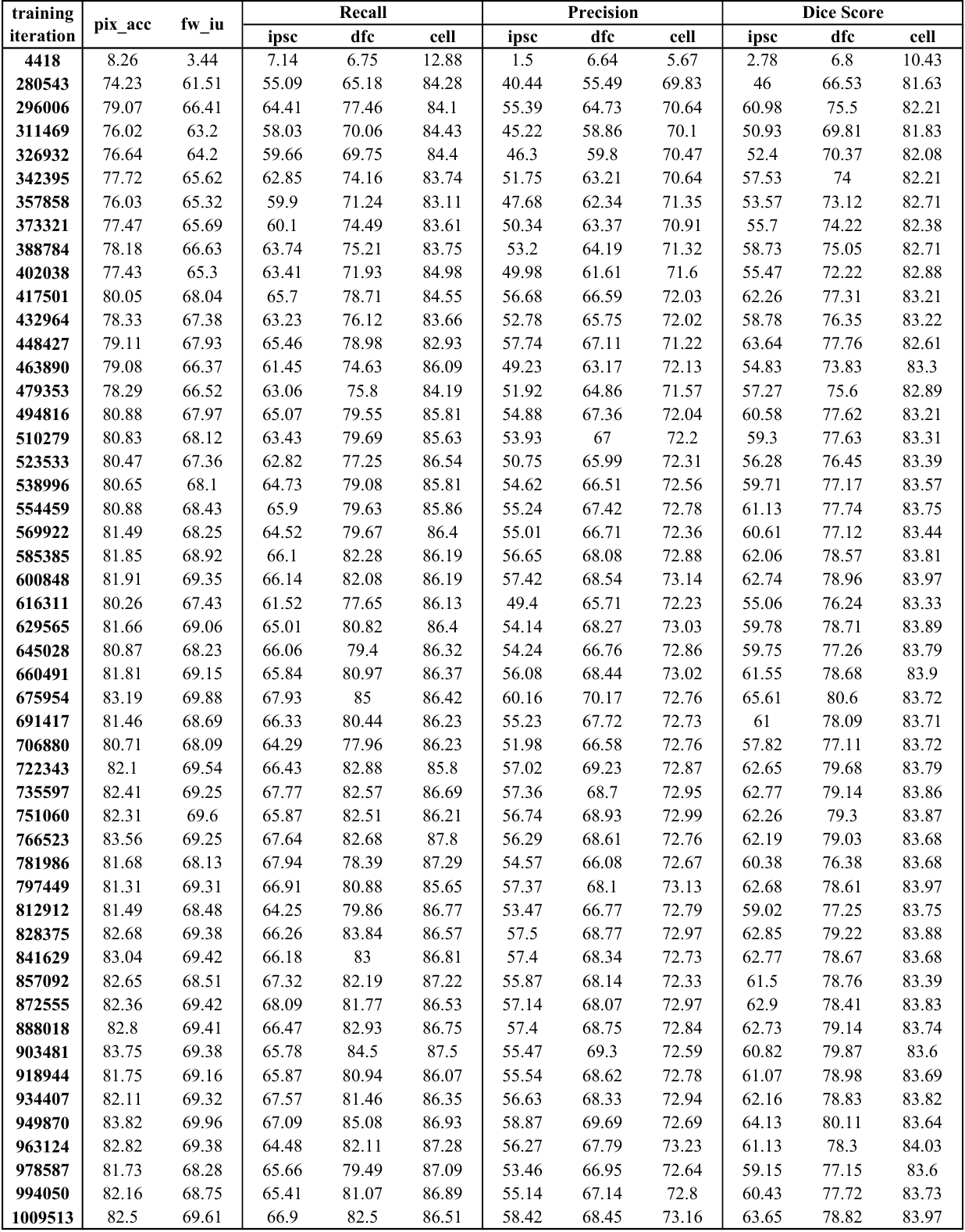}
	\end{tabular}	
	\label{tab:seg_details_ipsc_early_static}
\end{table}

\begin{table}[t]
	\centering
	\caption{
		Segmentation metrics ($\%$) on validation set while training P2S-VIDSEG on IPSC late-stage dataset.
				Best viewed under high magnification.			
	}
	\begin{tabular}{c}
		\includegraphics[width=0.47\textwidth]{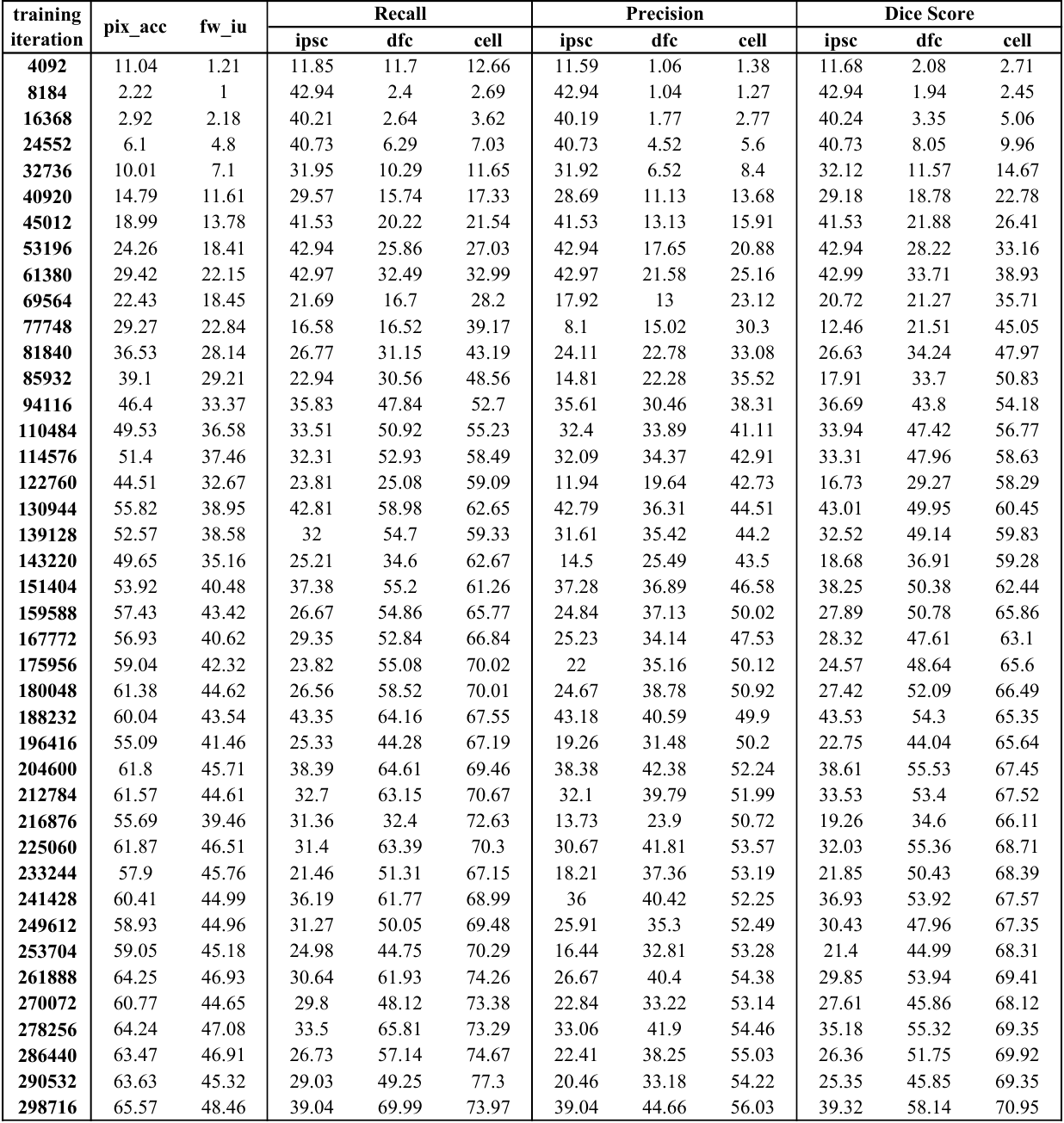}
	\end{tabular}	
	\label{tab:seg_details_ipsc_late_vid}
\end{table}

\begin{table*}[t]
	\centering
	\caption{
	Results of our best performing Cityscapes model on the validation set in terms of image-level, class-level and official metrics. 
	Table \ref{tab:ctscp_cats} provides class and category-wise breakdown of the latter.
	This model was trained with the source images resized from $2048 \times 1024$ to $1280 \times 640$, followed by sliding window patches of size $P=640$.
	We used offline RLE with no data augmentation, LAC encoding, frozen backbone, $S=80$, $L=2K$, $V=8K$, and $B=80$.
	}
	\begin{tabular}{c}
		\includegraphics[width=0.94\textwidth]{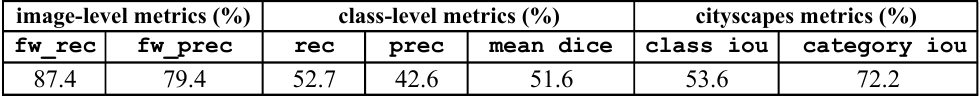}\\
	\end{tabular}	
	\label{tab:ctscp}
\end{table*}

\begin{table*}[t]
	\centering
	\caption{
		Class and category-wise breakdown of the results of our best performing Cityscapes model on the official cityscapes metrics.
		Please refer Table \ref{tab:ctscp} for model details.
		Best viewed under high magnification.				
	}
	\begin{tabular}{c}
		\includegraphics[width=0.98\textwidth]{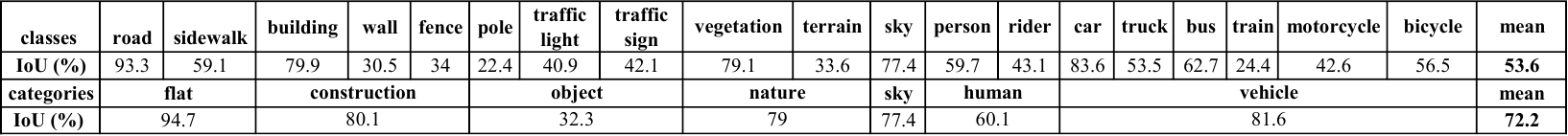}\\
	\end{tabular}	
	\label{tab:ctscp_cats}
\end{table*}

\begin{figure*}[t]
	\centering
	\includegraphics[width=\textwidth]{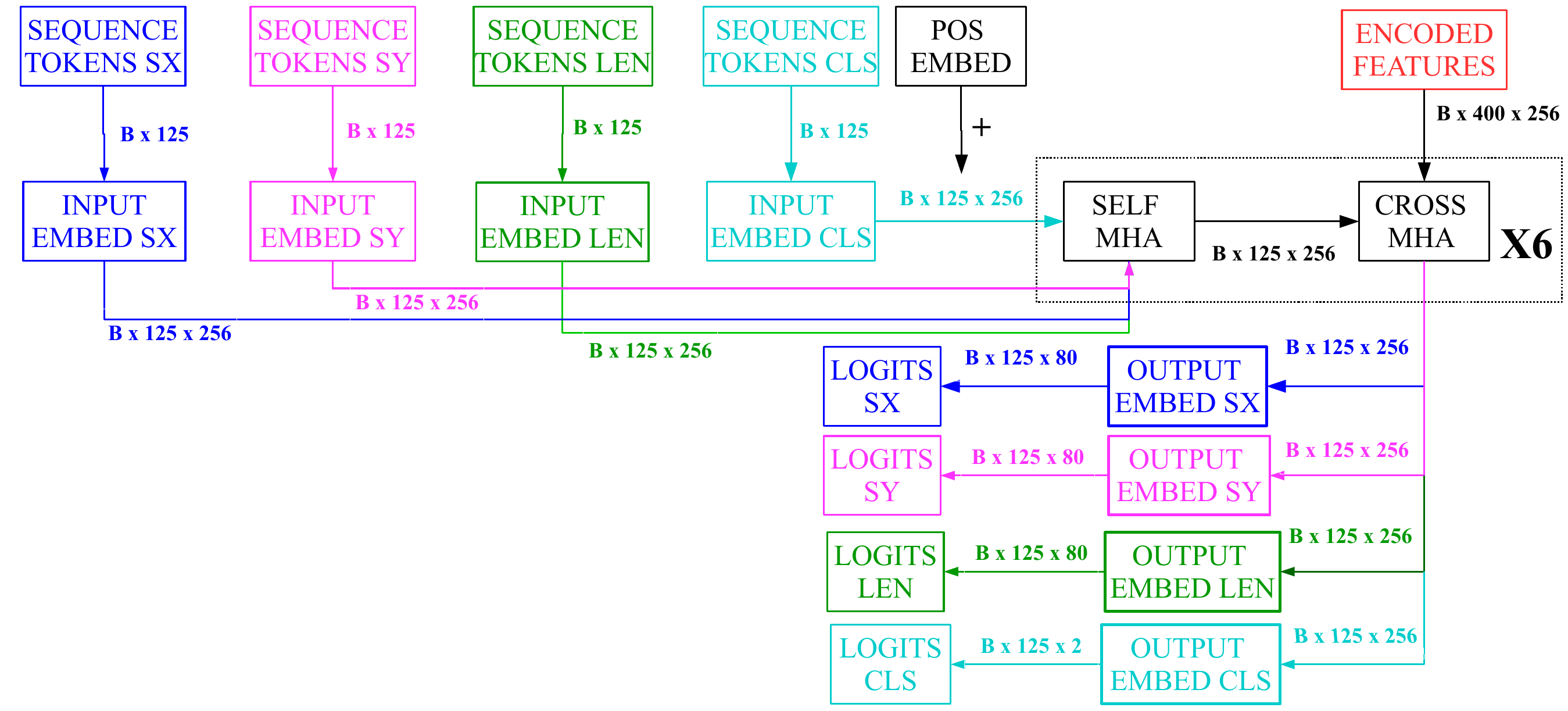}
	\caption{
		An example of a multi-headed decoder architecture to output each RLE component by a separate head, while sharing the memory-expensive MHA module between them.
		Please refer to
		the baseline decoder in \cite[Fig. 3]{singh25_p2s_vid_det_access} that this adapts.
		This uses $L=500$, $S=80$, $C=2$ and naïve RLE encoding with 2D starts and separate length and class tokens.
		There are thus 4 tokens per run, represented by {SX}, {SY}, {LEN} and {CLS}, and shown respectively in blue, magenta, green and cyan.
		Compared to 1D starts and LAC with a single-headed decoder, this not only cuts $L$ by half, but also reduces $V$ by up to 3 orders of magnitude, since each component now has its own vocabulary which can be as small as $S$ for coordinate tokens and $C$ for class tokens.
	}
	\label{fig:image_decoder_multi_embedding}	
\end{figure*}


%
%
%


\end{document}